%% file: main.tex
\documentclass{article}
\usepackage{authblk}
\usepackage[utf8]{inputenc}
\usepackage{main}
\usepackage{microtype}
\usepackage{subcaption}
\usepackage{graphicx}
\usepackage{times}
\usepackage{latexsym}
\usepackage{amsmath}
\usepackage{float}
\usepackage{footnote}
\usepackage{enumitem}
\usepackage{bm}
\usepackage{arydshln}
\usepackage{booktabs}
\usepackage{array}     %
\usepackage{multicol}
\usepackage{multirow}
\usepackage{color}
\usepackage{xcolor}     
\usepackage{colortbl}
\usepackage{bbding}
\usepackage{makecell}
\usepackage{mathtools}
\usepackage{imakeidx}
\usepackage{longtable}
\usepackage{wrapfig}
\usepackage{algorithmic}
\usepackage{rotating}
\makeindex
\usepackage{arydshln}
\usepackage{lipsum}
\usepackage{natbib}
\usepackage[toc]{multitoc}
\usepackage[edges]{forest}
\usepackage[normalem]{ulem}
\definecolor{mydarkblue}{rgb}{0,0.08,0.45}
\usepackage[colorlinks=true,linkcolor=mydarkblue,citecolor=mydarkblue,filecolor=mydarkblue,urlcolor=mydarkblue]{hyperref}
\usepackage{CJKutf8}
\usepackage{awesomebox} %
\usepackage{bbding}
\usepackage[most]{tcolorbox}
\usepackage{booktabs}
\usepackage{geometry}
\definecolor{wkblue}{rgb}{0.2, 0.3, 0.6}
\definecolor{meta-color}{rgb}{0.5, 0.5, 0.5}
\usepackage{amsmath}
\usepackage{enumitem}
\usepackage{lscape} 
\usepackage{booktabs}
\usepackage{algorithm}   
\usepackage{setspace}

\usepackage{algorithmic}
\usepackage{tabularx,booktabs}
\usepackage{makecell}

\usepackage{amssymb}
\usepackage{amsfonts}

\usepackage{booktabs} %
\usepackage{geometry} %

\usepackage{multirow}

\usepackage[tikz]{bclogo}
\usepackage[framemethod=tikz]{mdframed}
\definecolor{bgblue}{RGB}{245,243,253}
\definecolor{ttblue}{RGB}{91,194,224}

\usepackage{pgfplots}
\usepackage{pgfplotstable}

\usepackage{wrapfig}
\usepackage{graphicx}

\mdfdefinestyle{mystyle}{%
  rightline=true,
  innerleftmargin=10,
  innerrightmargin=10,
  outerlinewidth=3pt,
  topline=false,
  rightline=true,
  bottomline=false,
  skipabove=\topsep,
  skipbelow=\topsep
}

\newtcolorbox{myboxi}[1][]{
  breakable,
  title=#1,
  colback=red!5,
  colbacktitle=red!5,
  coltitle=black,
  fonttitle=\bfseries,
  bottomrule=0pt,
  toprule=0pt,
  leftrule=2pt,
  rightrule=2pt,
  titlerule=0pt,
  arc=0pt,
  outer arc=0pt,
  colframe=red,
}

\newtcolorbox{myboxnote}[1][]{
  breakable,
  title=#1,
  colback=orange!0,
  colbacktitle=orange!0,
  coltitle=black,
  fonttitle=\bfseries,
  bottomrule=0pt,
  toprule=0pt,
  leftrule=2pt,
  rightrule=2pt,
  titlerule=0pt,
  arc=0pt,
  outer arc=0pt,
  colframe=orange,
}

\newtcolorbox{myboxii}[1][]{
  breakable,
  freelance,
  title=#1,
  colback=white,
  colbacktitle=white,
  coltitle=black,
  fonttitle=\bfseries,
  bottomrule=0pt,
  boxrule=0pt,
  colframe=white,
  overlay unbroken and first={
  \draw[red!75!black,line width=3pt]
    ([xshift=5pt]frame.north west) -- 
    (frame.north west) -- 
    (frame.south west);
  \draw[red!75!black,line width=3pt]
    ([xshift=-5pt]frame.north east) -- 
    (frame.north east) -- 
    (frame.south east);
  },
  overlay unbroken app={
  \draw[red!75!black,line width=3pt,line cap=rect]
    (frame.south west) -- 
    ([xshift=5pt]frame.south west);
  \draw[red!75!black,line width=3pt,line cap=rect]
    (frame.south east) -- 
    ([xshift=-5pt]frame.south east);
  },
  overlay middle and last={
  \draw[red!75!black,line width=3pt]
    (frame.north west) -- 
    (frame.south west);
  \draw[red!75!black,line width=3pt]
    (frame.north east) -- 
    (frame.south east);
  },
  overlay last app={
  \draw[red!75!black,line width=3pt,line cap=rect]
    (frame.south west) --
    ([xshift=5pt]frame.south west);
  \draw[red!75!black,line width=3pt,line cap=rect]
    (frame.south east) --
    ([xshift=-5pt]frame.south east);
  },
}

\usepackage{fontawesome5}
\usepackage{fancyhdr} %
\usepackage{blindtext} %
\usepackage{makecell}

\DeclareCaptionFont{black}{\color{black}}

\definecolor{myblue}{rgb}{0.9, 0.1, 0.94}
\definecolor{mygreen}{rgb}{0.64, 0.56, 0.88}
\definecolor{myyellow}{rgb}{0.68, 0.6, 0.1}
\definecolor{fancygreen}{rgb}{0.33, 0.68, 0.20}
\definecolor{salmon}{rgb}{0.94, 0.52, 0.49}
\definecolor{tablegreen}{rgb}{0.82, 0.94, 0.75}
\definecolor{tableblue}{rgb}{0.81, 0.90, 0.94}
\definecolor{tablered}{rgb}{0.97, 0.85, 0.85}
\definecolor{tableorange}{rgb}{0.96, 0.85, 0.81}

\newenvironment{itemize*}%
 {\leftmargini=10pt\begin{itemize}%
  \setlength{\itemsep}{0pt}%
  \setlength{\parskip}{0pt}%
  }%
 {\end{itemize}}
\newenvironment{enumerate*}%
 {\begin{enumerate}%
  \setlength{\itemsep}{0pt}%
  \setlength{\parskip}{0pt}}%
 {\end{enumerate}}

\usepackage{xcolor}
\usepackage{listings}  

\newcommand\JSONnumbervaluestyle{\color{blue}}
\newcommand\JSONstringvaluestyle{\color{red}}

\newif\ifcolonfoundonthisline

\makeatletter

\lstdefinestyle{json}
{
  showstringspaces    = false,
  keywords            = {false,true},
  alsoletter          = 0123456789.,
  morestring          = [s]{"}{"},
  stringstyle         = \ifcolonfoundonthisline\JSONstringvaluestyle\fi,
  MoreSelectCharTable =%
    \lst@DefSaveDef{`:}\colon@json{\processColon@json},
  basicstyle          = \ttfamily,
  keywordstyle        = \ttfamily\bfseries,
}

\newcommand\processColon@json{%
  \colon@json%
  \ifnum\lst@mode=\lst@Pmode%
    \global\colonfoundonthislinetrue%
  \fi
}

\lst@AddToHook{Output}{%
  \ifcolonfoundonthisline%
    \ifnum\lst@mode=\lst@Pmode%
      \def\lst@thestyle{\JSONnumbervaluestyle}%
    \fi
  \fi
  \lsthk@DetectKeywords%
}

\lst@AddToHook{EOL}%
  {\global\colonfoundonthislinefalse}

\makeatother

\usepackage{etoolbox}
\usepackage{natbib}
\usepackage{url}
\newcounter{bibcount}
\makeatletter
\patchcmd{\@lbibitem}{\item[}{\item[\hfil\stepcounter{bibcount}{[\thebibcount]}}{}{}
\renewcommand\NAT@bibsetup%
  [1]{\setlength{\leftmargin}{\bibhang}\setlength{\itemindent}{-\parindent}%
      \setlength{\itemsep}{\bibsep}\setlength{\parsep}{\z@}}
\makeatother

\definecolor{mybrown}{RGB}{128,64,0}

\definecolor{titlecolor}{HTML}{4c9cff}
\definecolor{coolblue3}{rgb}{0.91, 0.94, 0.98}

\usetikzlibrary{arrows.meta,calc,positioning}
\pgfplotsset{compat=1.18}
\definecolor{scheduleTeal}{HTML}{187D8D}
\definecolor{schedulePurple}{HTML}{975579}
\definecolor{scheduleGray}{HTML}{627080}
\definecolor{scheduleInk}{HTML}{203040}
\newcommand{\SchedulingSpeedup}{2.26}

\providecommand{\PaperMaiStyle}{1}

\usepackage{main_revise}

\begin{document}

\providecommand{\PLACEHOLDER}[1]{{\color{red}\textbf{[TODO: #1]}}}
\providecommand{\TODOcite}[1]{{\color{red}\textbf{[CITE?: #1]}}}
\providecommand{\modelname}{801-30B-A3B\xspace}

\title{No Second Run: Does RSI Stop at Frontier Pre-training?}
\author{\textbf{SII} \& \textbf{Xinxu AI}  }

\input{preamble}

\maketitle
\ifdefined\PaperRecyclerStyle
\pagestyle{plain}
\else
\ifdefined\PaperMaiStyle
\pagestyle{plain}
\else
\pagestyle{fancy}
\thispagestyle{fancy}
\fancyhead{}
\lhead{
  \raisebox{-0.3cm}{\includegraphics[height=0.95cm]{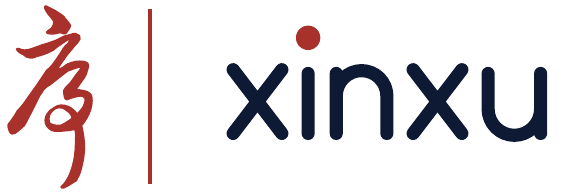}}
}
\rhead{%
  \raisebox{-0.2cm}{\includegraphics[height=0.7cm]{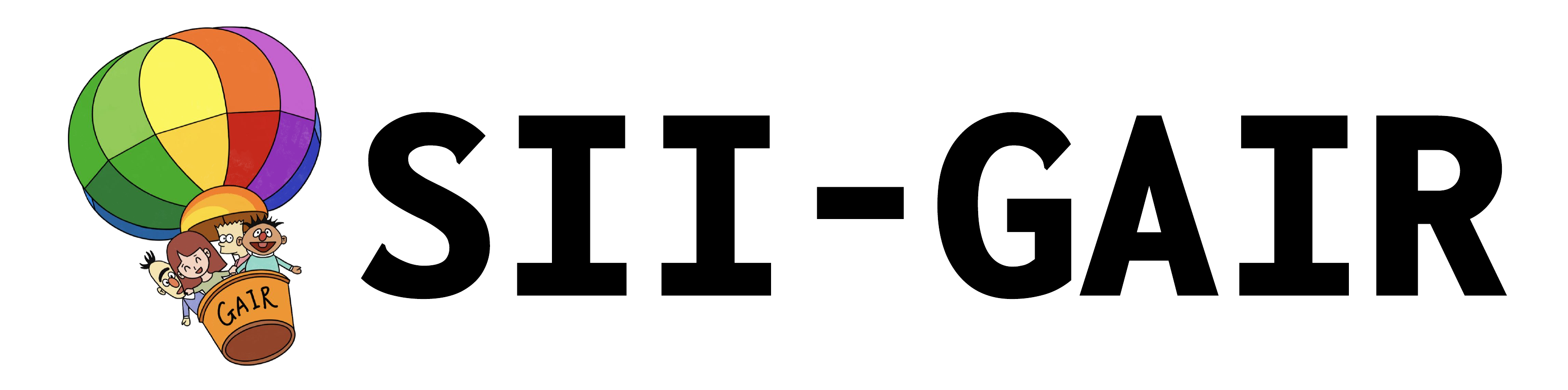}}%
}
\renewcommand{\headrulewidth}{0pt}
\setlength{\headsep}{2mm} %
\fi
\fi

\renewcommand{\thefootnote}{}
\ifdefined\PaperRecyclerStyle\vspace{2pt}\else\vspace{-50pt}\fi

\vspace{10pt}

\begin{abstract}
Pre-training is where every foundation model begins; bringing recursive
self-improvement (RSI) there would let AI improve the process that
creates AI, and could change how the entire industry builds foundation models.
Yet public RSI work has so far targeted only post-training, which starts from
an existing base model, runs trials costing GPU-hours, and returns verifiable
scores. RSI for pre-training breaks all three conditions at orders-of-magnitude higher
cost: a run consumes hundreds of thousands of GPU-hours over months and cannot
be repeated; its objective, post-training headroom, cannot be read off any
single score; and what must improve is not a model but the whole production \& organization
system, from data pipelines and training infrastructure to the teams that run
them. To our knowledge, \textbf{we are the first in the world to apply RSI to
an entire frontier pre-training run from scratch}, and to report it in detail.
RSI does not stop at frontier pre-training, but it changes form: since no
end-to-end loop can close over a single unrepeatable run, we decompose the
\emph{pre-training loop} into smaller loops that can, each repairing a step
the full run breaks.

\smallskip
The result is \modelname-Base,\footnote{We present 801 (pronounced eight-oh-one), named after the classroom where it was built.}
a 30B-A3B MoE model trained from scratch on 11.8T tokens in three months by a
newly formed team. We release it as a \textbf{\emph{base} model}: full
post-training is left to a subsequent release, and the lightweight SFT
reported here (6B tokens, no RL) only probes post-training headroom. As a
base model, \modelname{} outperforms Qwen3-30B-A3B and
Nemotron-3.5-Lightning-30B-A3B, trailing only the larger Qwen3.5-35B-A3B, and
keeps this position under identical SFT; with lightweight SFT alone, it
achieves the highest average on four agentic benchmarks against fully
post-trained models, including Qwen3.6-35B-A3B. Post-training RSI starts where
this work ends: it presupposes a strong base model, exactly what pre-training
must produce. Bringing RSI into pre-training thus acts on the foundation every
later stage builds on. We will open-source \modelname-Base and the lightweight
SFT data, a ready-made resource for evaluating the post-training headroom of
base models, at \url{https://github.com/GAIR-NLP/801-30B}.

  \begin{figure}[h]
      \centering
      \includegraphics[width=0.93\linewidth]{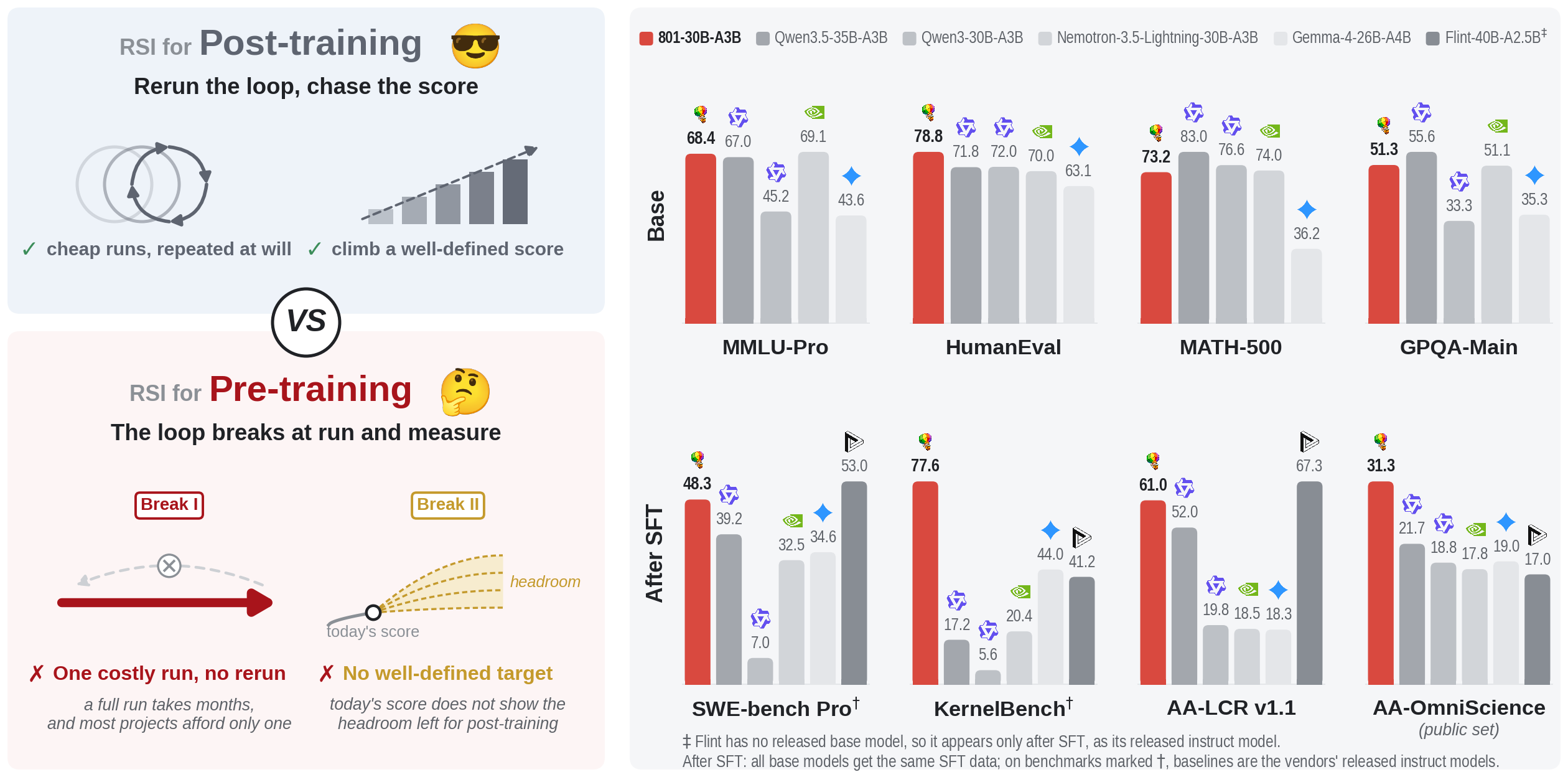}
      \caption{\emph{Left:} recursive self-improvement (RSI) takes a different form in pre-training than in post-training. \emph{Right:} \modelname (red) against open baselines on base-model benchmarks (top) and after supervised fine-tuning (bottom).}
      \label{fig:teaser}
  \end{figure}

\end{abstract}

\newpage

\pagestyle{fancy}
\lhead{\rightmark}
\renewcommand{\headrulewidth}{0.7pt}
\setlength{\headsep}{5mm}

\renewcommand{\thefootnote}{\arabic{footnote}}
\setcounter{footnote}{0}  %

\ApplyPaperBodyStyle
\input{1_intro}

\input{1_5_preliminary}

\input{2_context_infra}
\input{3_measure_pretrain}
\input{4_agentic_pretrain}

\input{5_scaling_agentic_data}

\input{6_training_recipe}
\input{7_evaluation}
\input{8_conclusion}

\vspace{2em}

\input{9_contributors}

\newpage
\bibliographystyle{acl_natbib}
\bibliography{main}
\clearpage
\input{10_appendix}

\end{document}

%% file: preamble.tex
\newcommand{\MethodFull}{Chain of Pull Requests\xspace}
\newcommand{\methodShort}{daVinci-Agency\xspace}
\newcommand{\ourDataset}{daVinci-Agency\xspace}
\definecolor{medalGold}{HTML}{D4A017}
\definecolor{medalSilver}{HTML}{9EA3A8}
\definecolor{medalBronze}{HTML}{B0703C}
\providecommand{\rankmedal}[2]{\tikz[baseline=(m.base)]{%
  \fill[#1!65!black] (-0.11,0.06)--(-0.17,0.30)--(-0.06,0.30)--(0.0,0.10)--cycle;%
  \fill[#1!65!black] (0.11,0.06)--(0.17,0.30)--(0.06,0.30)--(0.0,0.10)--cycle;%
  \fill[#1] (0,0) circle (0.155);%
  \draw[white,line width=0.35pt] (0,0) circle (0.12);%
  \node[inner sep=0pt,font=\sffamily\bfseries\scriptsize,text=white] (m) at (0,-0.005) {#2};}\,}
\providecommand{\goldmedal}{\rankmedal{medalGold}{1}}
\providecommand{\silvermedal}{\rankmedal{medalSilver}{2}}
\providecommand{\bronzemedal}{\rankmedal{medalBronze}{3}}

%% file: 1_intro.tex
\section{Introduction}
\label{sec:intro}

Recursive self-improvement (RSI), in which a system turns its own experience
and feedback into persistent changes that improve its capability, has moved
quickly from idea to practice. Agents now run the loop
$\textit{propose}\rightarrow\textit{run}\rightarrow\textit{measure}\rightarrow\textit{update}$
on neural architectures~\citep{xu2026asiarch}, GPU kernels~\citep{kernelsurvey2026,fu2026davinci},
agent harnesses~\citep{xia2026rrsi}, and, closest to model building, on
post-training itself~\citep{cao2026neohorse,posttrainbench2026,rsiarena2026}.
Where these loops touch pre-training, they do so through small-scale proxies:
data-curation and data-selection strategies validated on small models~\citep{mi2026dataevolve,xu2026asievolve,autodata2026},
or single-GPU training runs measured in minutes~\citep{karpathy2026autoresearch,autollmresearch2026}.
These settings share three conditions: they start from an existing base
model or a toy-scale one, each trial costs GPU-hours or less, and each
returns a verifiable score. Pre-training is where every foundation model
begins, and it satisfies none of the three. Bringing RSI there would let AI
improve the very process that creates AI; yet, to our knowledge, no public
work has attempted it on a full frontier pre-training run.

RSI for pre-training is not a harder instance of RSI for post-training but
a different regime (Table~\ref{tab:post-vs-pre}). \textbf{First}, it breaks
\textit{run}. A single run consumes hundreds of thousands of GPU-hours over
months, and most projects can afford only one at full length. An ablation
that isolates a single cause costs GPU-weeks, so most hypotheses are never
tested end to end. \textbf{Second}, it breaks \textit{measure}. The goal of
pre-training is not the highest score today but headroom for post-training.
A run that chases the metric can quietly borrow against post-training, for
example by annealing in supervised fine-tuning (SFT) data too early, or by
collapsing output diversity at the expense of the headroom that
reinforcement learning (RL) later relies on~\citep{yue2025rlpassk}. Such an
objective cannot be read off any single score, and if even a crisply
defined benchmark gets overfit by self-improvement loops~\citep{xia2026rrsi},
a target this diffuse offers far more ways to be hacked than to be met.
\textbf{Third}, what must improve is not a model but a production system
and the organization that runs it. Before a single token is trained, the
data must be built: in our run, six pipelines with more than 30 processing
stages turned raw web pages, documents, and code into 1,053 data subsets.
The training stack evolved through more than 130 infrastructure code
versions and 106 hyperparameter fields. Behind these artifacts stands an
organization: data, infrastructure, training, and evaluation teams, each
holding a different part of the evidence, with the hardest problems falling
between them. A load-balance anomaly, for instance, surfaced in training
metrics but traced to an infrastructure change interacting with optimizer
settings (Section~\ref{sec:context_infra}); no single team could have closed
that loop alone. Improving pre-training therefore means improving how the
organization divides work, shares evidence, and remembers what it has
learned. Post-training RSI starts where this
system ends: it presupposes a strong base model, which is exactly what the
system must produce.

\begin{table}[!b]
\centering
\caption{RSI for Post-training vs. RSI for frontier pre-training operate in different regimes.}
\label{tab:post-vs-pre}
\small
\setlength{\tabcolsep}{4pt}
\renewcommand{\arraystretch}{1.2}
\begin{tabular}{p{0.2\linewidth}p{0.36\linewidth}p{0.36\linewidth}}
\toprule
& \textbf{RSI for post-training} & \textbf{RSI for frontier pre-training (ours)} \\
\midrule
Starting point & An existing base model & Random initialization \\
Cost per trial & GPU-hours (e.g., 10 H100-hours~\citep{posttrainbench2026}) & Hundreds of thousands of GPU-hours for one run \\
Full-length trials & Tens to hundreds & One \\
Signal at \textit{measure} & Verifiable task score & Post-training headroom, not readable from any single score \\
Data & A fixed dataset or environment & Built from raw sources: 6 pipelines, $>$30 stages, 1,053 subsets, 10.5T unique tokens \\
Infrastructure & A training script & $>$130 code versions, 106 hyperparameter fields \\
What improves & A model or a harness & A production system and the organization that runs it \\
Who is in the loop & One agent or researcher & Data, infrastructure, training, and evaluation teams with their agents \\
\bottomrule
\end{tabular}
\end{table}

Small-scale proxies and scaling laws seem to offer a fast loop, but a cheap
loop cannot answer every expensive question. Seemingly minor fitting choices
shift the fitted exponent~\citep{besiroglu2024chinchilla,weng2026scaling},
small differences at fitting scale grow large at trillion-token scale, and a
ladder of proxy runs assumes that scale is the only thing
changing~\citep{weng2026scaling}. Once a question couples architecture,
recipe, infrastructure, and multi-source data mixtures, the proxy study
needed to settle it becomes as expensive as the run it was meant to de-risk.
We therefore use our two pilot runs to surface and eliminate problems, not
to extrapolate scores.

If the end-to-end loop cannot be rerun, where do agents earn their place?
The run leaves behind a vast context: internal metrics, hyperparameter
changes, data-mixture edits, infrastructure diffs, and dozens of
evaluations on every intermediate checkpoint, far more than any team can
read, let alone remember. The characteristic waste in this regime is not a
bad experiment but re-investigating a question settled weeks earlier,
because no one linked the symptom to its class of cause. The bottleneck
thus shifts from compute to institutional memory: what was tried, what
failed and why, and which signals co-occurred. Agents earn their place by
recalling and attributing across evidence no human will finish reading,
and, wherever a fast signal can be built, by doing the work itself:
building data-pipeline components, crawling the web, and generating
agentic training data at a breadth no team could manage by hand.

\begin{figure}[!htbp] 
    \centering
    \includegraphics[width=\linewidth]{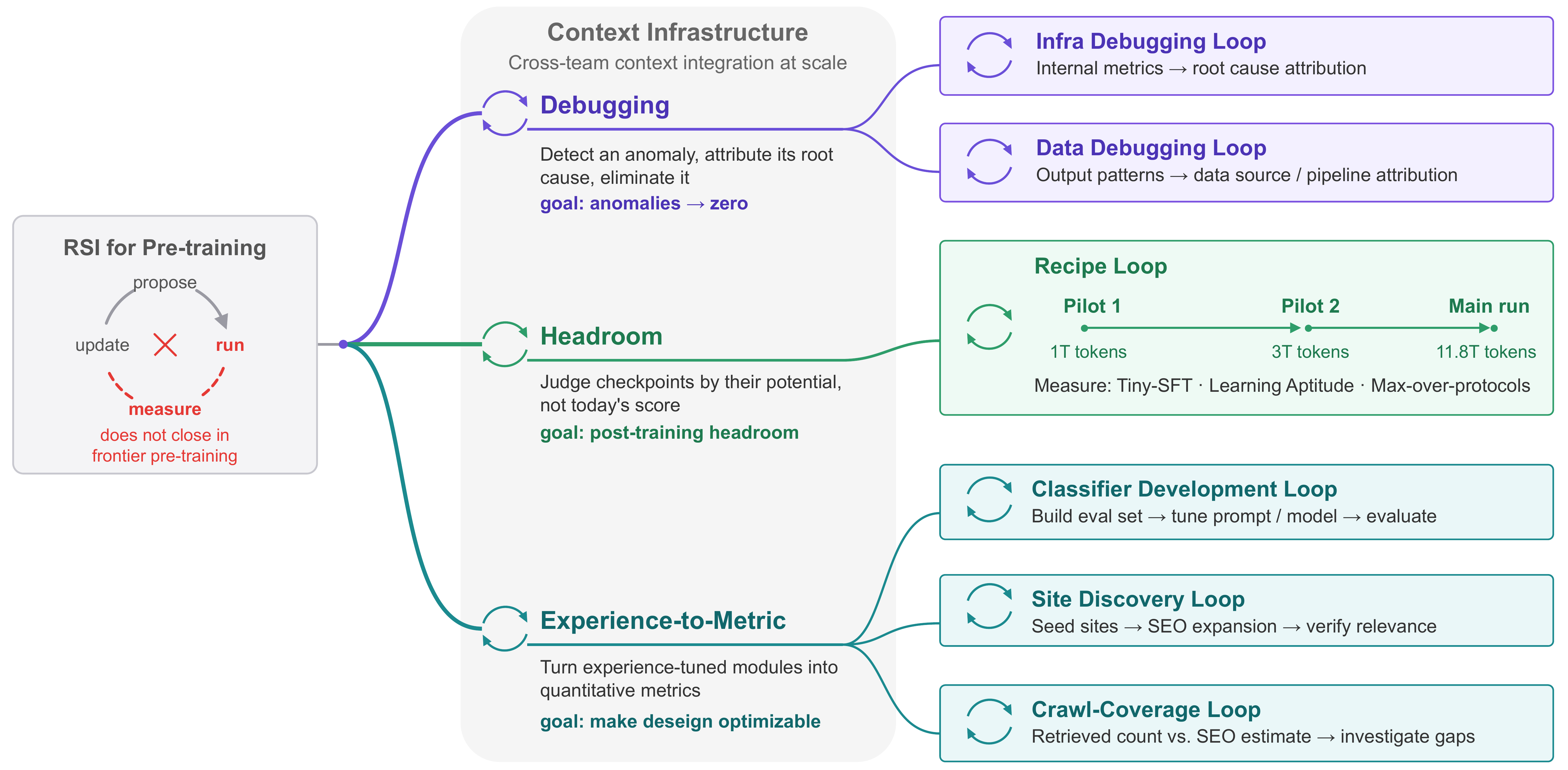}
    \caption{The standard RSI loop ($\textit{propose}\rightarrow\textit{run}\rightarrow\textit{measure}\rightarrow\textit{update}$)
    does not close in frontier pre-training: a run takes months and cannot be
    repeated, and today's score does not reflect the headroom a checkpoint
    leaves for post-training. We propose three types of loops, instantiated as six
    practical loops. Debugging loops comprise the infra debugging loop and
    the data debugging loop, which trace internal-metric and output anomalies
    to their causes. The headroom loop is instantiated as the
    recipe loop, using headroom measurements to guide the whole pre-training process.
    Experience-to-metric loops comprise the classifier development loop,
    the site discovery loop, and the crawl-coverage loop, turning
    experience-based judgment into measurable feedback. All six loops share
    a context infrastructure that integrates evidence across teams at scale.%
    }
    \label{fig:three_loops}
\end{figure}

Our answer to the question in the title is that RSI does not stop at
frontier pre-training, but it changes form. Since no end-to-end loop can
close over a single unrepeatable run, we decompose pre-training into smaller
loops that can close, each repairing a step the full run breaks.
\emph{Debugging loops} relocate \textit{run} onto the infrastructure and
data stacks, which can be rerun cheaply; the \emph{headroom loop} repairs
\textit{measure} for the run as a whole; and \emph{experience-to-metric
loops} build a \textit{measure} where none existed. A shared \emph{context
infrastructure} serves as the organization's memory: it integrates evidence
across teams and preserves negative results and their reasons, which is where the \textit{update} step persists
and the recursion lives (Figure~\ref{fig:three_loops}).

We built these loops while training \modelname, a 30B-A3B
Mixture-of-Experts (MoE)~\citep{shazeer2017moe} model trained from scratch
on 11.8T tokens in three months by a newly formed team. To our knowledge,
this is the first public account of RSI applied to an entire frontier
pre-training run. Our contributions are:
\begin{itemize}
\item \emph{(a) Debugging loops (Section~\ref{sec:context_infra}).}
An \emph{infra debugging loop} and a \emph{data debugging loop}, driven by
internal-metric and output anomalies, respectively. Across two pilots
(stopped at 1T and 3T tokens) and the main run, agents localized causes
spanning infrastructure, data, and recipe, from a load-balance penalty
interacting with optimizer state to low-diversity synthetic data, and
resolved five cross-stack diagnoses in about one month. Together with data
and recipe revisions, these fixes raised the mean score over 22 benchmarks
(Appendix~\ref{app:pretraining-benchmarks}) at a matched 1T-token budget
from 47.78 (initial pilot) to 51.96 (revised pilot) to 56.83 (main run).
\item \emph{(b) A headroom loop (Section~\ref{sec:measure}).} Because base
scores do not preserve post-training rankings, we judge checkpoints by what
they are \emph{worth} after adaptation, using tiny-SFT, a learning aptitude
curve, and a max-over-protocols estimate of the capability ceiling. These
measurements separate apparent stagnation in base evaluation from
continued gains after adaptation.
\item \emph{(c) Experience-to-metric loops (Section~\ref{sec:agentic_pretrain}).}
Classifier development, site discovery, and crawl coverage, in which agents
turn judgment once made by feel into recorded, quantitatively evaluated
workflows. We built six data pipelines from scratch this way.
\item \emph{(d) Agentic data at pre-training scale (Section~\ref{sec:agentic_data}).}
Agents cross from consuming training to producing it: 220K executable pull
requests rolled out across nine harnesses yielded more than 100B tokens of
verified agentic trajectories (134B agentic tokens in total), used in the
long-context stage of pre-training.
\end{itemize}

\paragraph{What we release and how to compare it.}
We release \modelname-Base as a \emph{base} model. Full post-training is
left to a subsequent release; the lightweight SFT reported here (6B tokens,
no RL) serves only to probe post-training headroom. We therefore compare at
three levels, from the fairest to the least favorable to us
(Section~\ref{sec:evaluation}). (1) \emph{Base against base}: \modelname{}
outperforms Qwen3-30B-A3B~\citep{yang2025qwen3} and
Nemotron-3.5-Lightning-30B-A3B~\citep{nvidia2026nemotron35lightning},
trailing only the larger Qwen3.5-35B-A3B~\citep{qwen2026qwen35}.
(2) \emph{Identical SFT}: when every base model is fine-tuned on the same
SFT data, it keeps this position. (3) \emph{Lightweight SFT against full
post-training}: even so, it achieves the highest average across four
agentic benchmarks among the fully post-trained models we compare,
including Qwen3.6-35B-A3B~\citep{qwen2026qwen36}. The third comparison pits
a few billion SFT tokens against complete vendor post-training pipelines;
we report it as a reference point for headroom, not as a like-for-like
comparison. We will open-source \modelname-Base and the lightweight SFT
data, a ready-made resource for evaluating the post-training headroom of
base models, at \url{https://github.com/GAIR-NLP/801-30B}.

%% file: 1_5_preliminary.tex
\section{Adapting RSI to Frontier Pre-Training}
\label{sec:preliminary}

Recursive self-improvement (RSI) is the idea that a system turns its own
experience and feedback into persistent changes that improve its capability
and, in the strongest form, the process of improvement itself. 
The engine behind it is a closed loop, exemplified by
AlphaEvolve~\citep{novikov2025alphaevolve} and
ASI-Evolve~\citep{xu2026asievolve}. We write that loop as $\textit{propose}\rightarrow\textit{run}\rightarrow\textit{measure}
\rightarrow\textit{update}$ and use it as the operational definition of RSI throughout this report. In this report, the improving system is the pre-training project itself, comprising its agents, researchers, and infrastructure, rather than the model being trained.

The four steps are worth defining precisely. \emph{Propose}
forms a candidate change or hypothesis, the role played by the generator in
generate-critique-revise schemes~\citep{madaan2023selfrefine} and by the proposer in proposer-solver
self-play~\citep{zhao2025absolutezero}. \emph{Run} realizes that candidate to produce evidence, by training,
rolling out, or executing it. \emph{Measure} scores the evidence against a
signal that stands in for human judgment or the desired target. \citet{chen2026rsisurvey} arrange
candidate signals into a verification hierarchy, from formal verifiers at the
strong end to a model's intrinsic self-assessment at the weak end.
\emph{Update} folds the outcome back into a persistent change, keeping what
scored higher and discarding the rest, and in the recursive case it also
revises the loop machinery that produced the change. This persistence is what makes the loop recursive rather than a one-off search.

Two preconditions decide whether this loop improves anything. The first is
cheap, repeatable iteration: gains compound only when a trial is cheap enough
that \textit{run} and \textit{update} can cycle many times. The second is a
reliable signal at \textit{measure}, aligned with the true goal. Without an
external signal there is no reliable improvement
;
with a weak or misaligned one, the loop degenerates through self-confirming
errors, model and diversity collapse~\citep{shumailov2024collapse}, or frame
lock-in on a well-posed but wrong objective~\citep{chen2026rsisurvey}.

As argued in Section~\ref{sec:intro}, frontier pre-training breaks both
preconditions: a full-length run cannot be repeated, and its true objective,
post-training headroom, is too diffuse to serve as a trustworthy
\textit{measure}. The system that must improve is, moreover, an organization
rather than a model. The standard end-to-end loop therefore does not transfer
to this regime. We keep the loop, but change where it closes.

\begin{table}[!t]
\centering
\small
\caption{How each loop type instantiates the four RSI steps. Debugging and
experience-to-metric loops close quickly on cheap components; the headroom loop
closes only a few times per project, because its \textit{run} is the
pre-training run itself.}
\renewcommand{\arraystretch}{1.2}
\begin{tabularx}{\linewidth}{
  >{\raggedright\arraybackslash\hsize=0.8\hsize}X
  >{\raggedright\arraybackslash\hsize=1.05\hsize}X
  >{\raggedright\arraybackslash\hsize=1.05\hsize}X
  >{\raggedright\arraybackslash\hsize=1.05\hsize}X
  >{\raggedright\arraybackslash\hsize=1.05\hsize}X}
\toprule
\textbf{Loop type} & \textbf{Propose} & \textbf{Run} & \textbf{Measure} & \textbf{Update} \\
\midrule
Debugging & A suspected cause: code version, configuration, or data subset & Targeted diagnosis on the infra or data stack; cheap and repeatable & Does the anomaly disappear? & Apply the fix; record ruled-out causes \\
\addlinespace
Headroom & A recipe change, or whether to continue training & Pilots and main-run checkpoints; costly and not repeatable & Tiny-SFT, learning aptitude, max-over-protocols & Revise the recipe; judge training progress \\
\addlinespace
Experience-to-metric & A new prompt, model, or crawler version & Component-level evaluation; cheap & Task-specific evaluation set; SEO page-count estimate & Deploy, or iterate again \\
\bottomrule
\end{tabularx}
\label{tab:loop-steps}
\end{table}

\begin{figure}
    \centering
    \includegraphics[width=0.7\linewidth]{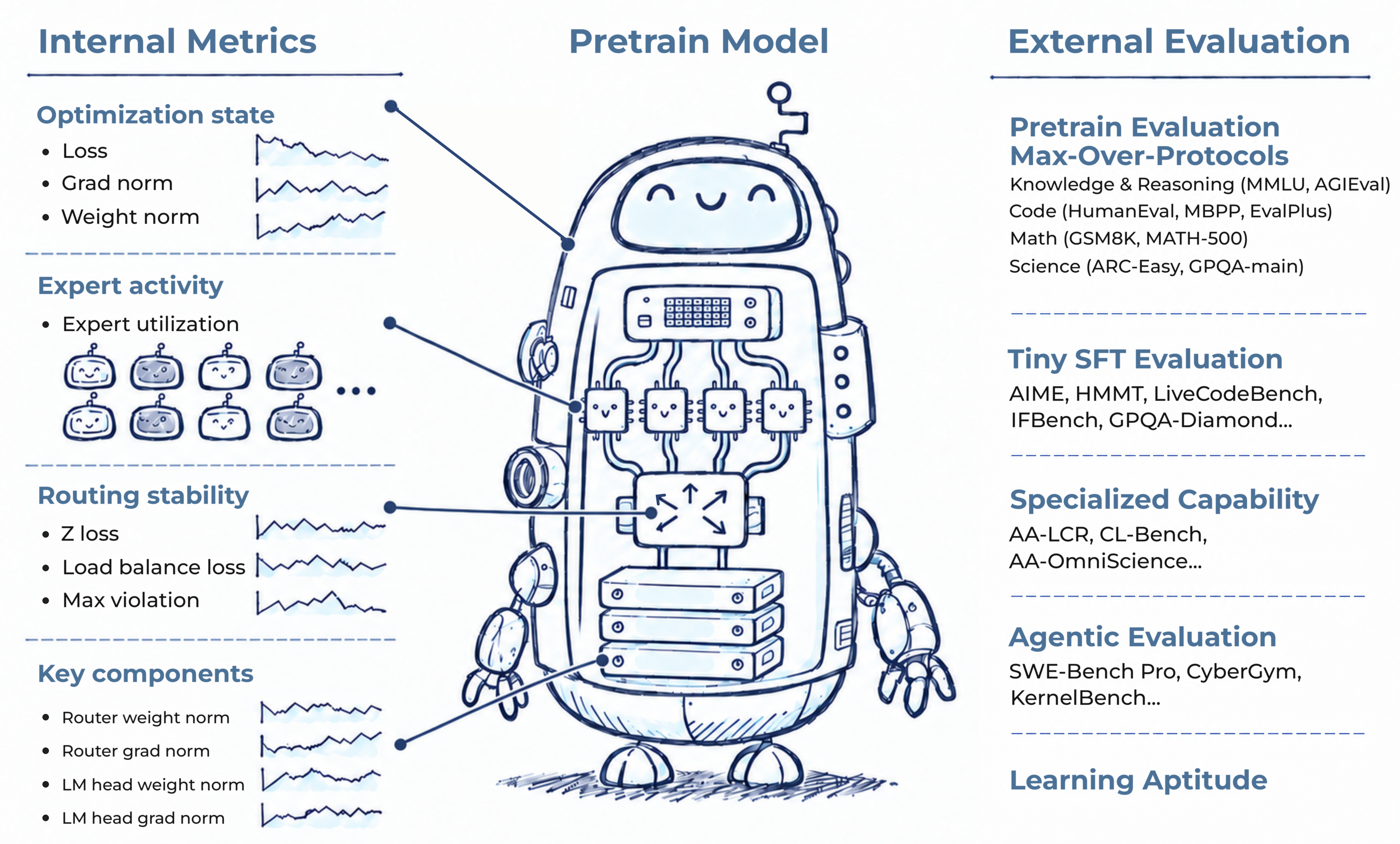}
    \caption{Overview of internal training metrics and external evaluation protocols for pre-training monitoring.}
    \label{fig:model-metric}
\end{figure}

\paragraph{Three types of loops.}
Rather than casting pre-training as one score-driven loop, we close three
types of smaller loops around it, with six concrete instances
(Figure~\ref{fig:three_loops}). Table~\ref{tab:loop-steps} shows how each
type instantiates the four steps.
\begin{itemize}
\item \emph{Debugging loops} relocate \textit{run} onto the infrastructure and data-pipeline stacks, where a diagnosis can be rerun cheaply. They are driven by anomalies rather than by a score to maximize: \textit{measure} asks whether the anomaly disappears, and each ruled-out cause is recorded as a negative result. The \emph{infra debugging loop} traces unusual internal metrics to infrastructure code or training-recipe settings, and the \emph{data debugging loop} traces unusual output patterns to a specific data source or processing stage (Section~\ref{sec:context_infra}).
\item The \emph{headroom loop} repairs \textit{measure} for the run as a whole. Its signal is post-training headroom, estimated through tiny-SFT, learning aptitude, and max-over-protocols evaluation rather than read off a base-model score. It is instantiated as the \emph{recipe loop}, which uses these measurements to revise the recipe from the two pilots, stopped at 1T and 3T tokens, to main pre-training, and to monitor progress within the main run (Section~\ref{sec:measure}). Because its \textit{run} is the costly run itself, it closes only a few times per project.
\item \emph{Experience-to-metric loops} build a \textit{measure} where none existed. Steps such as building data-pipeline components and acquiring web data were historically tuned by hand; we give each an explicit evaluation, such as a task-specific evaluation set or an SEO page-count estimate, so that revisions can be compared. This type is instantiated as the \emph{classifier development loop}, the \emph{site discovery loop}, and the \emph{crawl-coverage loop} (Section~\ref{sec:agentic_pretrain}).
\end{itemize}

\paragraph{Context infrastructure.}
These loops operate on different parts of pre-training, but all require
evidence that spans teams: an anomaly may connect a model's output to a
data-processing decision made weeks earlier, headroom measurements must be
read alongside recipes and checkpoint histories, and data workflows must
retain the evaluations behind each revision. We built a \emph{context
infrastructure} to integrate this evidence (internal metrics such as per-layer
loss, per-module gradient norms, and expert load-balance statistics, together
with hyperparameter changes, data-mixture edits, infrastructure diffs, and
checkpoint evaluations), produced by agents and by researchers who separately
own the data, infrastructure, training, and evaluation stacks, at a scale no
team can read unaided (Table~\ref{tab:post-vs-pre}). It records not only what
was tried but what failed and why, so that negative results become persistent
knowledge. This is where the \textit{update} step persists and where the
recursion in our loops lives: each new proposal starts from what earlier
iterations have ruled out, and agents direct their effort to the next
unresolved bottleneck instead of rediscovering old ones.

\begin{figure}[t]
\centering
\includegraphics[width=\linewidth]{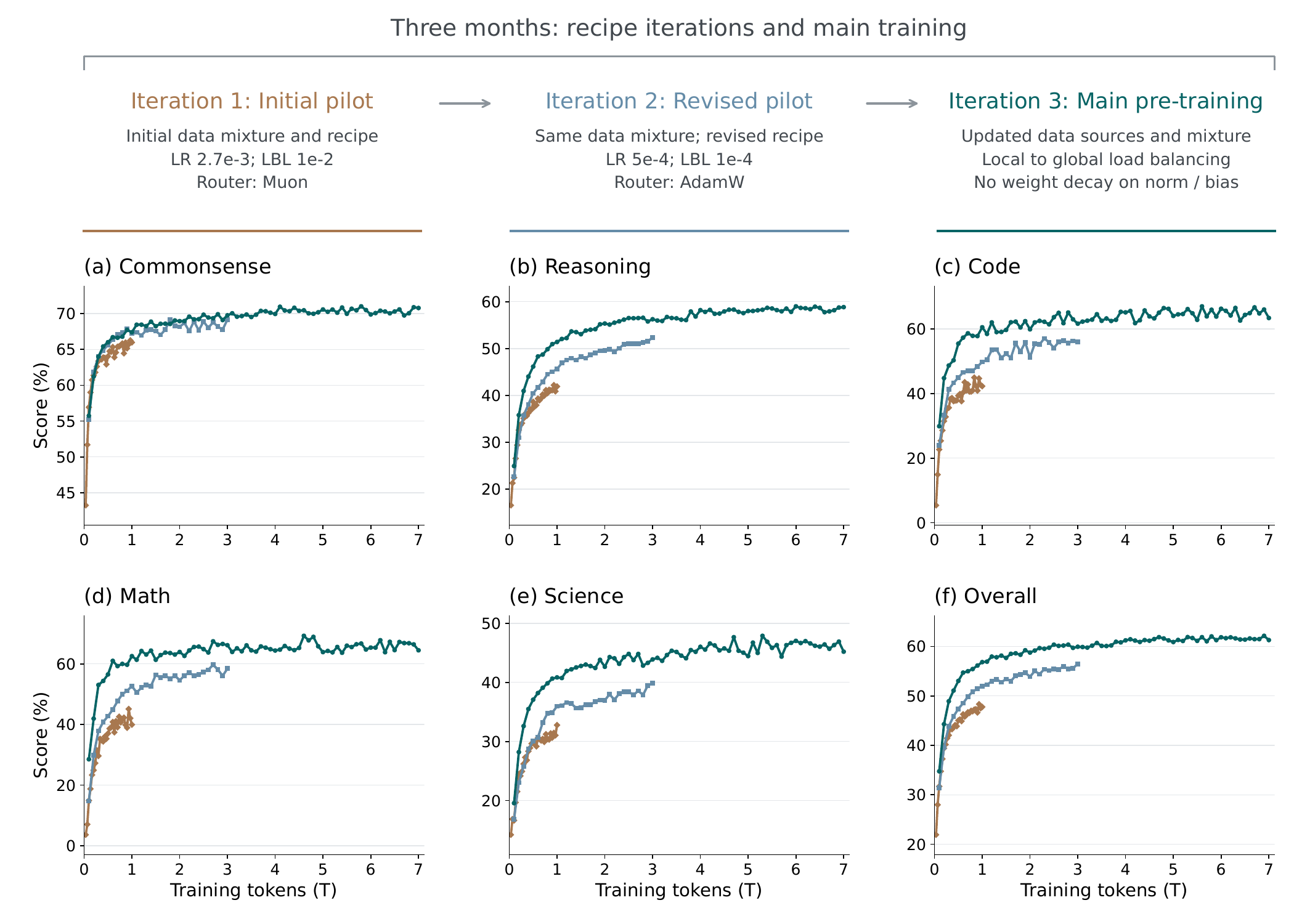}
\caption{Two pilots and the main pre-training in a three-month period. \emph{Top}: main configuration changes; matching colors link
each iteration to its evaluation curves below. \emph{Bottom}: initial pilot (warm brown) and revised pilot (gray-blue) stopped at 1T and 3T tokens, respectively; main pre-training (dark teal) is shown through Stage1-1 at 7T tokens. The five subset scores are means over their constituent benchmarks, and Overall is the mean over all shared benchmarks.}
\label{fig:pilot-vs-formal}
\end{figure}

\paragraph{From pilots to main pre-training.}
Together, these loops supported revisions of the data and training recipe across two pilots and main pre-training (Figure~\ref{fig:pilot-vs-formal}). The initial pilot stopped at 1T tokens, the revised pilot at 3T, and the main run used approximately 11.8T tokens; the figure shows main pre-training through Stage1-1 at 7T. These revisions brought together debugging, headroom assessment, and quantitatively evaluated data workflows.

At 1T tokens the mean score over the 22 common benchmarks rises from
47.78 to 51.96 to 56.83, and at 3T tokens main pre-training reaches 59.94
against the revised pilot's 56.48. Because data and recipe changed together, these gains
reflect the combined iterations supported by the three types of loops rather
than any single fix.
The following sections describe how each type contributed to this process.

%% file: 2_context_infra.tex
\section{Debugging Loops}
\label{sec:context_infra}

\begin{figure}
    \centering
    \includegraphics[width=\linewidth]{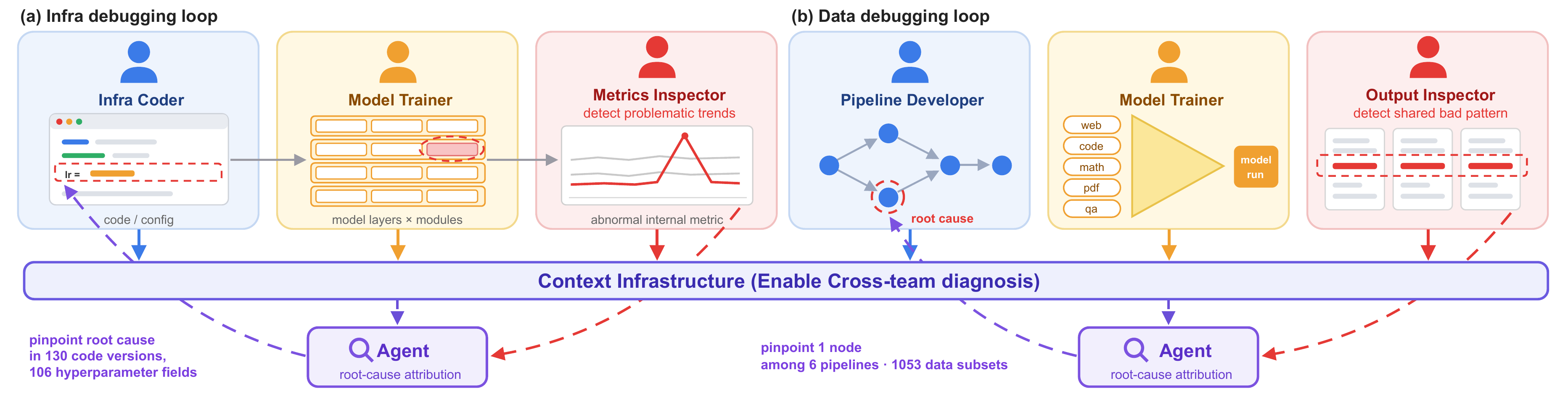}
    \caption{The debugging loops are instantiated as \textbf{(a)}~the infra debugging loop and \textbf{(b)}~the data debugging loop.}
    \label{fig:two_loops}
\end{figure}

This section shows the two debugging loops operating on the context
infrastructure as we introduced in Section~\ref{sec:preliminary}.
(1) The \emph{infra debugging loop} traces unusual internal metrics to
infrastructure code or training-recipe settings.
(2) The \emph{data debugging loop} traces unusual output patterns to a
specific data source or processing stage
(Figure~\ref{fig:two_loops}; Table~\ref{tab:loop-steps} maps both loops
to the four RSI steps).
Unlike conventional RSI approaches, which improve performance by
repeatedly training and comparing candidate models, these loops operate
within a single pre-training run. Rather than restarting training to
evaluate new candidate models, they continuously identify, attribute,
and eliminate problems as training progresses, improving the ongoing
run itself.
We conclude with the five cross-stack diagnoses produced by these
loops (Table~\ref{tab:diagnoses}).

\subsection{Infra Debugging Loop}
\label{sec:infra-debug}

The infra debugging loop uses internal training metrics as the
starting point for identifying and resolving anomalies during
pre-training. By linking these signals to infrastructure code
and training-recipe configurations through the context
infrastructure, agents can trace unexpected model behavior
to its underlying causes, propose targeted fixes, and verify
their effects within the ongoing training run. We illustrate
this process through the anomalies observed in the two pilot
runs, following four steps: detect, attribute, fix, and verify.

\begin{figure}[!htb]
\centering
\includegraphics[width=0.9\linewidth]{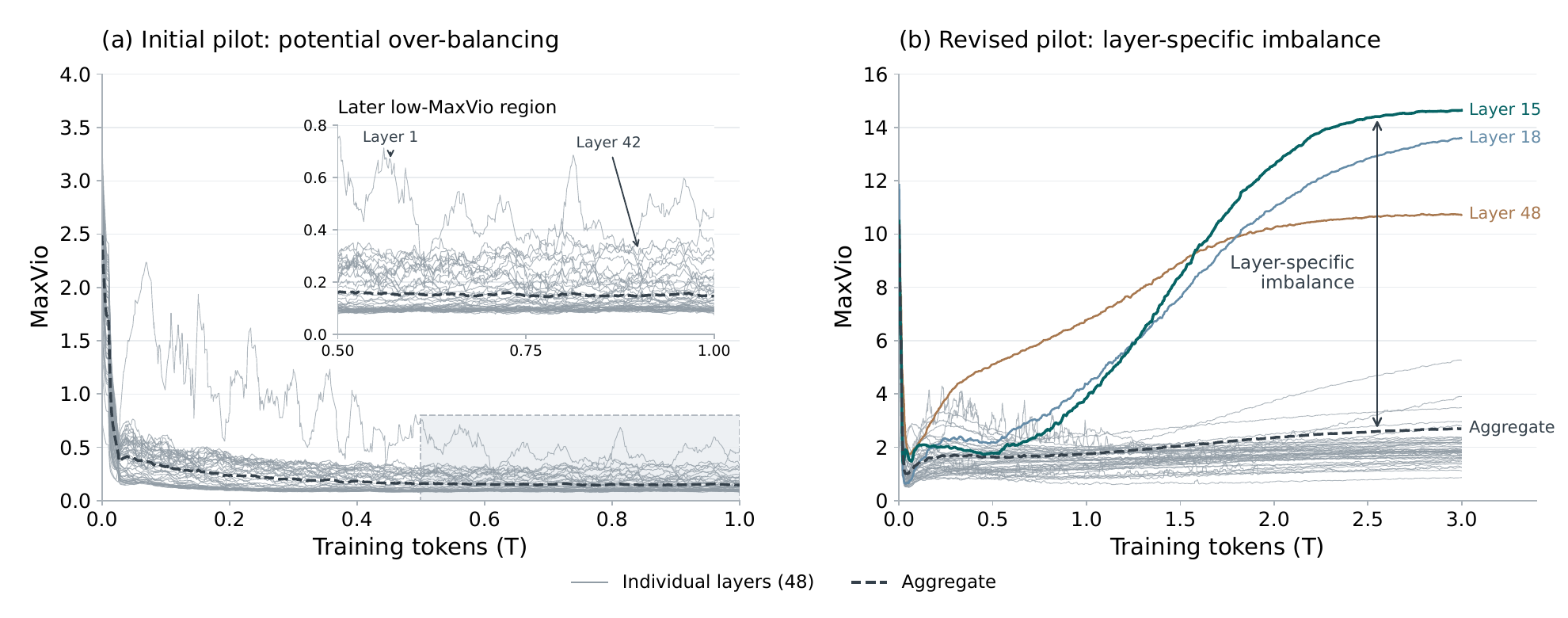}
\caption{Two contrasting load-balancing anomalies in pilot runs. Gray curves show layer-wise MaxVio across 48 layers; the dashed curve shows aggregate MaxVio. (a) In the initial pilot, both aggregate and layer-wise MaxVio decline under strong balancing, suggesting potential over-balancing. The inset enlarges the low-MaxVio region. (b) In the revised pilot, layers 15, 18, and 48 exhibit elevated MaxVio despite a moderate aggregate, revealing layer-specific imbalance obscured by aggregate monitoring.}
\label{fig:maxvio-two-pilots}
\end{figure}

\textbf{Detect.}
During the two pilot runs, agents scanning internal metrics through the context infrastructure identified two contrasting patterns in MoE load-balancing indicators (Figure~\ref{fig:maxvio-two-pilots}).
In the initial pilot, both aggregate and layer-wise MaxVio~\citep{wang2024auxfree} declined to low levels under a strong balancing penalty, raising concerns about potential over-balancing. The revised pilot revealed a different pattern: layers 15 and 18 developed large, persistent MaxVio violations, while the aggregate remained within a typical range. This discrepancy showed that aggregate monitoring alone could obscure severe layer-specific imbalance. Layer 48, the final layer, was also elevated, but we do not treat this alone as evidence of failure, since its function is closely tied to the output space. Together, these two patterns illustrate why monitoring MoE load balancing requires examining both aggregate trends and layer-specific deviations.

\textbf{Attribute.}
To identify the source of these anomalies, agents used the context infrastructure to correlate metric trajectories with infrastructure code commits and optimizer configurations. By searching across more than 130 infrastructure code versions and 106 hyperparameter fields, they narrowed the cause to a change in the load-balance penalty implementation and its interaction with optimizer state updates for certain parameter groups.
This process established an attribution chain from training stagnation to internal-metric anomalies, and from those anomalies to specific infrastructure and configuration changes. Rather than treating unusual metrics as isolated symptoms, agents traced them across the training stack to identify their underlying causes.

\textbf{Fix.}
Based on this diagnosis, we first revised the pilot configuration by lowering the load-balancing~\citep{fedus2022switch} coefficient from $10^{-2}$ to $10^{-4}$, reducing the learning rate from $2.7\times10^{-3}$ to $5\times10^{-4}$, and switching the optimizer for router parameters from Muon~\citep{jordan2024muon,liu2025muon} to AdamW~\citep{loshchilov2019adamw}. These settings were carried into main pre-training, with two further adjustments: computing the load-balance loss over the global batch rather than per micro-batch~\citep{qiu2025globalbatch}, and excluding normalization and bias parameters from weight decay.
Because several factors changed together, the comparison does not isolate the contribution of any single adjustment. Together, these changes constituted the revised training configuration used in main pre-training.

\begin{figure}[!htb]
\centering
\subfloat[Internal signal: layer-wise load imbalance.\label{fig:maxvio-internal}]{%
\includegraphics[width=0.9\linewidth]{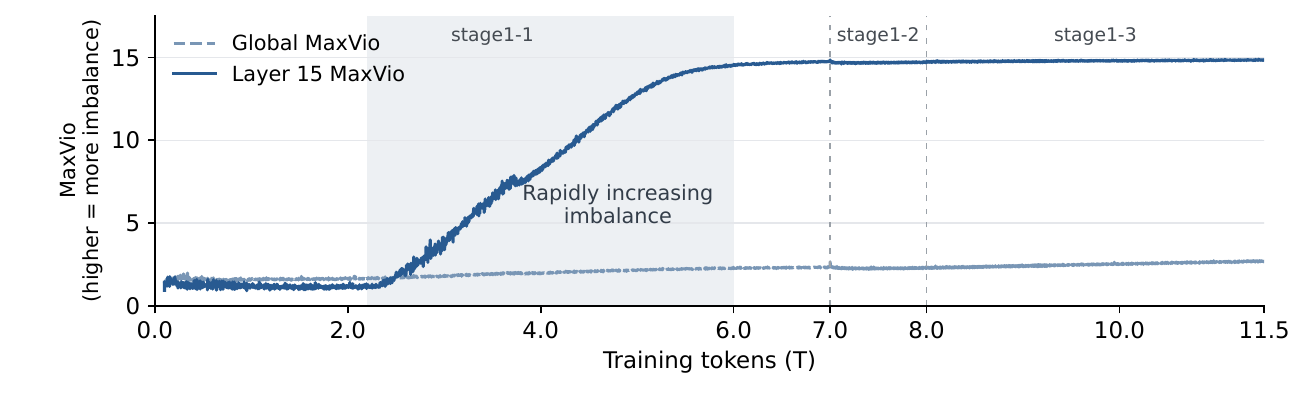}}
\par\medskip
\subfloat[External signal: downstream evaluation.\label{fig:maxvio-evaluation}]{%
\includegraphics[width=0.9\linewidth]{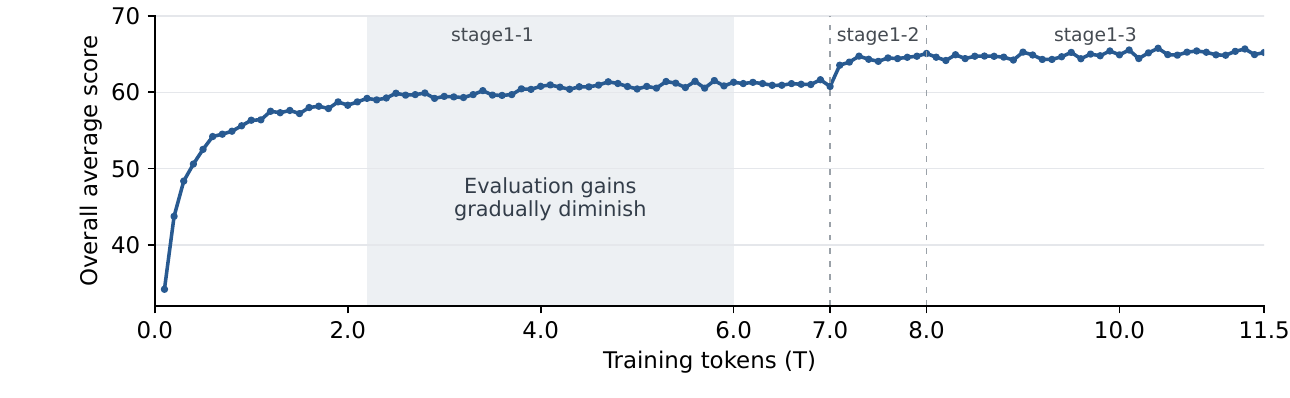}}
\caption{Layer-wise load imbalance rises as downstream evaluation
gains diminish during training. The gray shaded region marks the
same interval (approximately 2.2--6T tokens) in both panels.
(a) Layer-15 MaxVio increases rapidly during this interval, while
global MaxVio remains relatively stable. (b) Over the same interval,
downstream evaluation gains gradually diminish. Higher MaxVio
indicates greater load imbalance. This contrast illustrates how
layer-wise internal metrics provide a more granular signal for
detecting potential training issues than downstream evaluation
alone.}
\label{fig:maxvio-saturation}
\end{figure}

\textbf{Verify.}
Subsequent training shows that the revised configuration mitigates layer-wise load imbalance and suggests that internal metrics may reveal potential training problems earlier than downstream evaluation alone. (1) First, the revised configuration substantially mitigated the layer-specific imbalance observed in the pilot. At a matched training budget of 3T tokens, layer-15 MaxVio had already exceeded 14 in the revised pilot, whereas it was around 5 in main pre-training, with pronounced escalation occurring only at a later stage (Figure~\ref{fig:maxvio-saturation}). This indicates that the revised configuration substantially alleviated the earlier anomaly, although it did not eliminate it entirely.
(2) Second, the internal metrics provided a more granular view of training dynamics than downstream evaluation alone. During main pre-training, layer-15 MaxVio rose sharply over approximately 2.2--6T tokens (the shaded interval in Figure~\ref{fig:maxvio-saturation}). Over the same interval, downstream evaluation gains gradually diminished. After approximately 6T tokens, layer-15 MaxVio remained at a high plateau, while downstream evaluation gains within Stage1-1 had largely flattened.
This temporal correspondence suggests a possible association between layer-wise load imbalance and training saturation. Layer-wise MaxVio rises well before downstream performance plateaus, indicating that internal metrics can flag potential training problems while evaluation scores are still improving. Monitoring these metrics alongside downstream evaluations may therefore enable earlier investigation and intervention.

Finally, the context infrastructure allows these diagnostic experiences to persist across iterations. Because the previous diagnosis and its supporting evidence were preserved, the recurrence of similar anomalies could be anticipated rather than rediscovered from scratch. This allowed us to use the early MaxVio signal as a basis for further investigation, rather than waiting until downstream evaluation gains had flattened.

\subsection{Data Debugging Loop}
\label{sec:data-debug}

\begin{figure}[!htb]
    \centering
    \includegraphics[width=\linewidth]{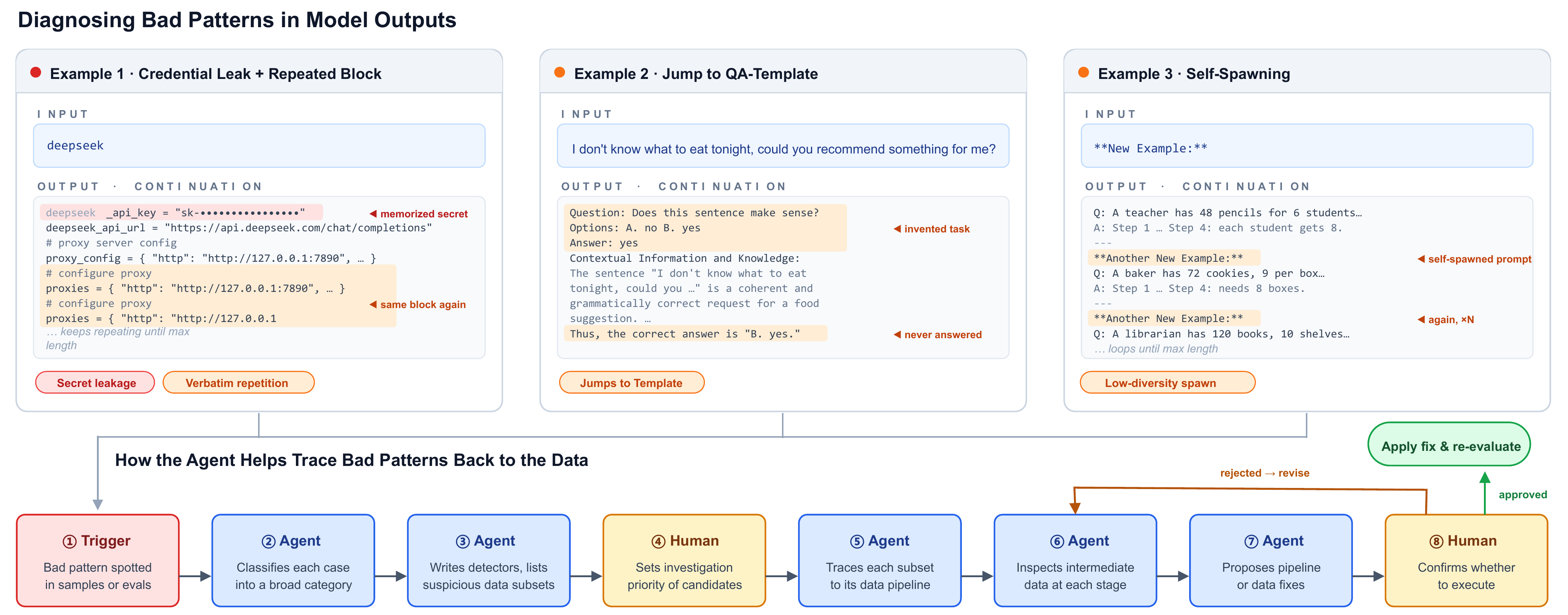}
    \caption{Diagnosing bad patterns in model outputs. \emph{Top}: three representative output patterns, a credential leak with a verbatim-repeated config block, a jump into a fixed ``Does this sentence make sense?'' QA template, and a self-spawning chain of near-identical word problems. \emph{Bottom}: the eight-step
    loop that traces each pattern back to a data-pipeline defect. Agents handle detection, tracing, and inspection at scale, humans set triage priority and approve the fix.}
    \label{fig:badpattern}
\end{figure}

During the two pilots, we continuously monitored the output behavior
of the trained model. Agents identified recurring patterns in the
model's responses to greetings and common questions, which are
indicative of repetitive, low-diversity, or insufficiently cleaned
training data. They extracted these anomalous patterns and searched
the data pipeline for matching signatures at scale.

\textbf{Anomalous patterns.}
Figure~\ref{fig:badpattern} illustrates three representative
patterns observed in model outputs.
(1) First, \emph{credential leakage and repetition}: given the prompt
\texttt{deepseek}, the model emits a secret resembling an API key,
followed by repeated copies of an identical proxy-configuration
block, continuing until the generation length limit is reached.
(2) Second, \emph{jumping to a fixed QA template}: when given a natural
request for a dinner recommendation, the model fails to answer the
question and instead switches to a fixed ``Does this sentence make
sense?'' multiple-choice template, proceeding to solve this
invented task.
(3) Third, \emph{self-spawning repetitions}: given only the marker
\texttt{**New Example:**}, the model generates an unbounded sequence
of nearly identical division word problems, each following the same
four-step solution scaffold.

\textbf{Automated diagnosis process.}
The bottom of Figure~\ref{fig:badpattern} illustrates how agents
trace these anomalous patterns back to specific data-pipeline
defects. The eight-step process can be organized into three stages.
(1) The first stage is \emph{detection and pattern extraction}.
Agents identify anomalies in model-generated samples or evaluation
results and classify the corresponding cases. Based on the
identified patterns, agents write detectors and perform large-scale
matching across all 1,053 data subsets to identify those potentially
containing the corresponding anomalies.
(2) The second stage is \emph{attribution and localization}.
After candidate data subsets have been identified, humans determine
the priority of investigation. Agents then trace these subsets to
their data pipelines (six in total) and inspect the
intermediate data produced at more than 30 processing stages across
these pipelines, ultimately locating the specific data sources and
processing stages responsible for the anomalies.
(3) The final stage is \emph{resolution and verification}.
Once the source of an anomaly has been identified, agents propose fixes to the data or pipeline. Humans decide whether
to approve their execution. Approved fixes are applied to the
relevant processing stages, followed by model re-evaluation to
verify their effectiveness; rejected proposals are returned for
revision.

Throughout this process, agents handle large-scale anomaly
detection, data tracing, and stage-by-stage inspection, while
humans retain control over two key decision points: investigation
prioritization and fix approval. This division of labor makes
large-scale investigations systematic and practical, beyond what
could reasonably be accomplished through manual inspection alone.

\textbf{Root cause attribution and fixes.}
Through this process, the three anomalous patterns were traced to
three distinct data-pipeline defects.
(1) First, the credential-leakage and repetition case was traced to
insufficiently cleaned configuration files in the code corpus.
These files contained sensitive information and repetitive
configuration content that had entered the training data and
subsequently surfaced in model generations. We addressed this
issue by fixing the corresponding data-processing stage to remove
such inadequately cleaned content.
(2) Second, the fixed-QA-template case was traced to insufficient
filtering of synthetic data in a subset of an open-source QA corpus.
This subset contained a large volume of
``Does this sentence make sense?'' questions. We strengthened
synthetic-data filtering at the corresponding stage to reduce the
presence of such fixed-template examples in the training data.
(3) Finally, the self-spawning case was traced to a synthetic-data
pipeline whose prompt design lacked sufficient diversity, resulting
in highly similar generated examples. We revised the corresponding
prompt design to improve the diversity of the synthetic data.
For all three cases, we modified the corresponding data-pipeline
stages and re-evaluated the model after applying the fixes to
verify their effectiveness.

\subsection{Accumulated Diagnoses}
\label{sec:diagnoses}

The two debugging loops described above illustrate how agents
identify anomalies in model behavior and trace them back to
underlying issues in the infrastructure, training recipe, and
data pipeline. However, these are only representative examples
of a broader effort to diagnose and resolve problems throughout
the training process.

Across the two pilots and main pre-training, we applied the
same approach to systematically investigate issues across
different parts of the training system. Many of these issues
could not be attributed to a single component. Diagnosing them
required connecting evidence from internal metrics,
infrastructure code, training configurations, and data
pipelines, and tracing the interactions among them.
Table~\ref{tab:diagnoses} summarizes five cross-stack issues
identified and resolved through this process, covering
load balancing, optimizer configurations, data quality,
weight decay, and data loading. These findings
demonstrate how agents, supported by the context
infrastructure, can identify and resolve problems that
span multiple parts of the training system.

\begin{table}[!htb]
\centering
\small
\caption{The five cross-stack diagnoses resolved with the context infrastructure across the two pilots and main pre-training.}
\renewcommand{\arraystretch}{1.15}
\begin{tabularx}{\linewidth}{@{}l>{\raggedright\arraybackslash}X>{\raggedright\arraybackslash}X@{}}
\toprule
\textbf{Area} & \textbf{Finding} & \textbf{Fix} \\
\midrule
Load balancing
 & Expert load was either over-balanced across all layers or severely imbalanced in a few individual layers (Figure~\ref{fig:maxvio-two-pilots}).
 & Adopt global-batch load balancing~\citep{qiu2025globalbatch}, which relaxes the per-micro-batch constraint and allows experts to specialize. \\
\addlinespace
Optimizer
 & Weight-norm and gradient-norm trends diverged across modules.
 & Reassign which modules use Muon and which stay on AdamW (Table~\ref{tab:optimizer_allocation}). \\
\addlinespace
Data quality
 & Pilot models generated advertisements, which were traced to advertisements and low-quality web pages remaining in the training corpus.
 & Filter them further with quality classifiers. \\
\addlinespace
Weight decay
 & Router logits showed abnormal trends during training.
 & Exclude normalization and bias parameters from weight decay. \\
\addlinespace
Data loader
 & When resuming a training task with the same global batch size but a different DP size, the sampler did not preserve the per-step data order of the previous task.
 & Restore the per-step sample order of the previous task when resuming with a different DP size. \\
\bottomrule
\end{tabularx}

\label{tab:diagnoses}
\end{table}

Beyond the fixes themselves, the time required to complete these
diagnoses demonstrates the value of the context
infrastructure. In our experience, diagnosing issues of this
kind can take months without such infrastructure. With the
support of agents and the ability to connect evidence across
the training stack, we completed all five diagnoses in roughly
one month, substantially reducing the manual effort required
to identify and trace the problems.

More importantly, the value of these diagnoses extends beyond
the immediate fixes. The context infrastructure preserves the
observations, attempted fixes, and negative results from each
investigation, allowing subsequent iterations to build on
previous findings rather than start from scratch. Each
investigation thus contributes not only to improving the
ongoing training run, but also to a growing body of knowledge
that can be reused in future iterations. Together, the two
debugging loops turn individual problem-solving efforts into
a cumulative process of continuous improvement, with concrete
data and training-recipe improvements carried forward into
the successive iterations shown in Figure~\ref{fig:pilot-vs-formal}.

%% file: 3_measure_pretrain.tex
\section{Headroom Loop}
\label{sec:measure}

The goal of pre-training is not to maximize today's score but to leave the
greatest headroom for post-training. Measuring this headroom can supply the signal
for the headroom loop. We first describe three complementary measurements
(Section~\ref{sec:measure-methods}), then show what they revealed that base
evaluation did not (Section~\ref{sec:measure-methods}), and finally how the
recipe loop used them (Section~\ref{sec:recipe-loop}).

\subsection{Measuring Post-Training Headroom}
\label{sec:measure-methods}

We use three measures to assess post-training
headroom: Tiny-SFT, learning aptitude, and max-over-protocols
evaluation. Together, they capture three aspects of a model's
potential: its performance after limited post-training, its
efficiency in learning from additional data, and the capabilities
it can demonstrate under evaluation protocols.

\begin{figure}[!htb]
\centering
\begin{minipage}[c]{0.51\linewidth}
\centering
\includegraphics[width=\linewidth]{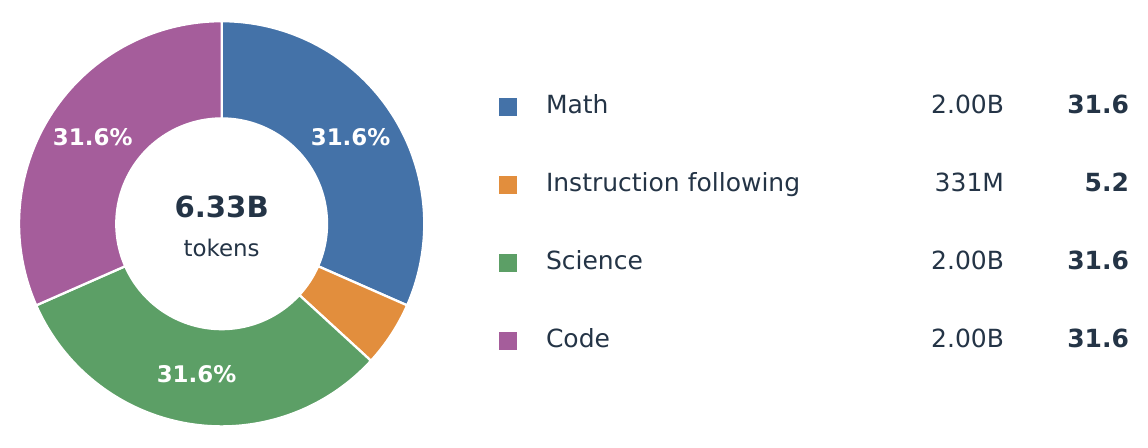}
\end{minipage}\hfill
\begin{minipage}[c]{0.47\linewidth}
\centering
\includegraphics[width=\linewidth]{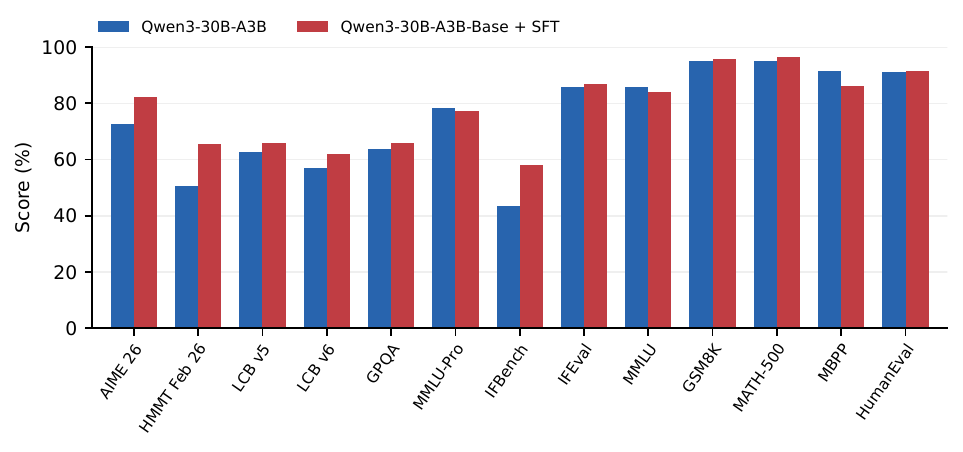}
\end{minipage}
\caption{Tiny-SFT data composition and validation.
\emph{Left:} Token composition of the 6.33B-token mix. Math,
Science~\citep{xu2026logicsstemempoweringllmreasoning}, and Code~\citep{ahmad2025opencodereasoningiisimpletesttime} each contribute approximately 2.00B tokens
(31.6\%), while instruction following~\citep{pyatkin2025ifbench} contributes 0.331B tokens
(5.2\%). 
\emph{Right:} Qwen3-30B-A3B versus its base model
(Qwen3-30B-A3B-Base) after one epoch of Tiny-SFT. Scores are
percentages across 13 benchmarks; the fine-tuned model scores
higher on 10 of them. This validation supports the effectiveness
of Tiny-SFT as a proxy for monitoring model performance after
limited post-training.}
\label{fig:tiny-sft-composition-baseline}
\end{figure}

\begin{figure}[!htb]
\centering
\includegraphics[width=\linewidth]{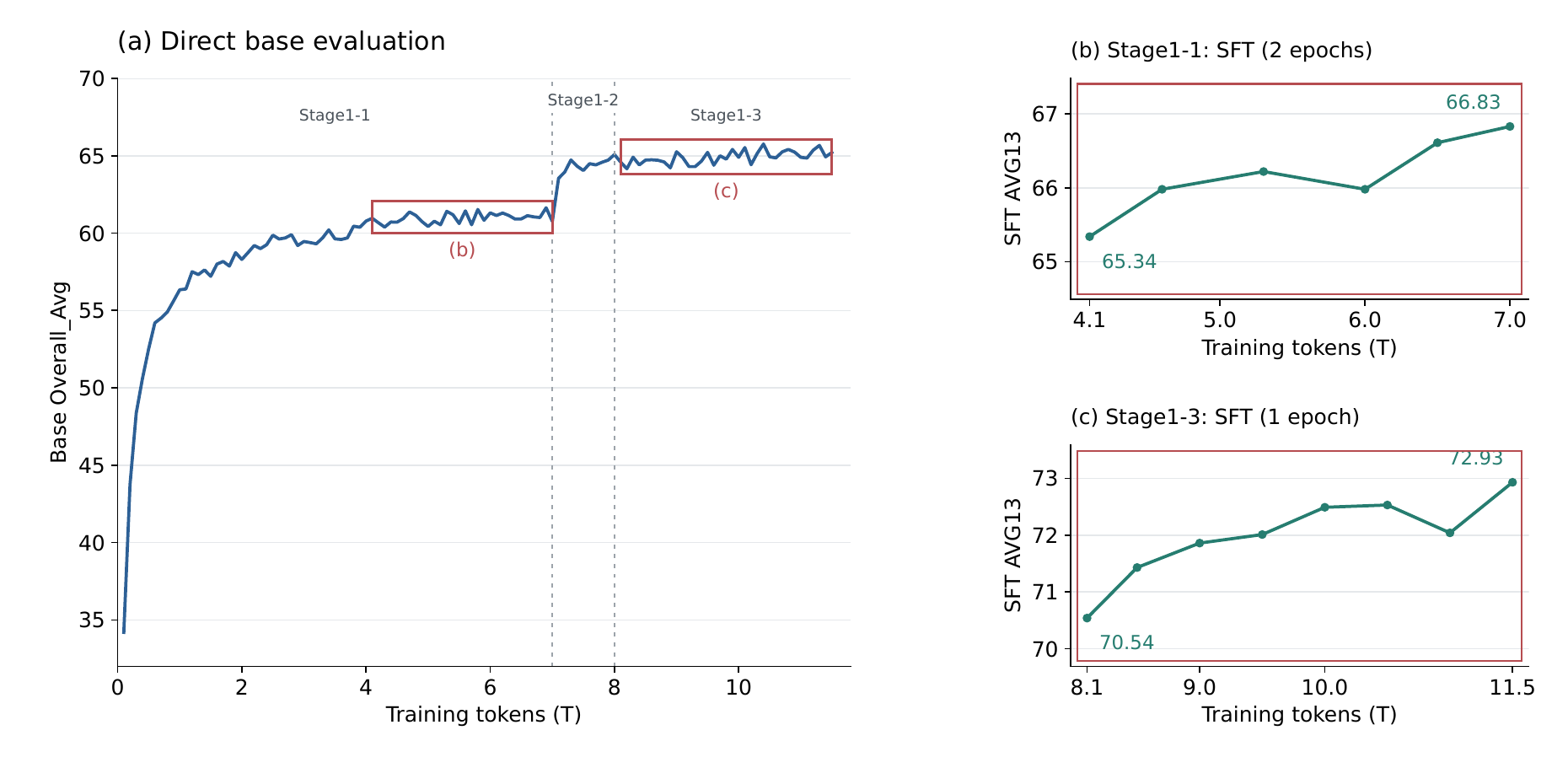}
\caption{Apparent stagnation in base evaluation can coexist with improving
post-SFT performance. (a) Direct base-model Overall\_Avg over pre-training;
red boxes mark the checkpoint intervals examined in (b) and (c).  Each point in (b) and (c) reports SFT AVG13 after applying the same SFT procedure independently to a checkpoint from (a).
(b) Stage1-1, steps 123,000--210,000 (4.1T-7T tokens): post-SFT scores improve by
1.49 points despite the near-flat base trend.
(c) Stage1-3, steps 243,000--345,000 (8.1T-11.5T tokens): post-SFT scores improve by
2.39 points while base scores fluctuate around a plateau.}
\label{fig:false-stagnation}
\end{figure}

\textbf{Tiny-SFT.}
Base-model evaluation alone does not fully capture a model's
potential after post-training. We therefore apply a small SFT
budget to each evaluated pre-training checkpoint and directly
measure its performance on benchmarks designed for post-trained
models. The resulting scores approximate how the checkpoint
would perform after a representative post-training phase,
providing a more realistic measure of its downstream potential.
The SFT corpus spans multiple task categories to approximate
the diversity of actual post-training workflows
(Figure~\ref{fig:tiny-sft-composition-baseline}).
We note that SFT was not repeated with multiple random seeds,
so small differences between adjacent checkpoints should
be interpreted with caution.
In our experiments, Tiny-SFT helped us identify two
interesting patterns that are not readily apparent from
base-model evaluation alone:

\begin{figure}[t]
\centering
\includegraphics[width=\linewidth]{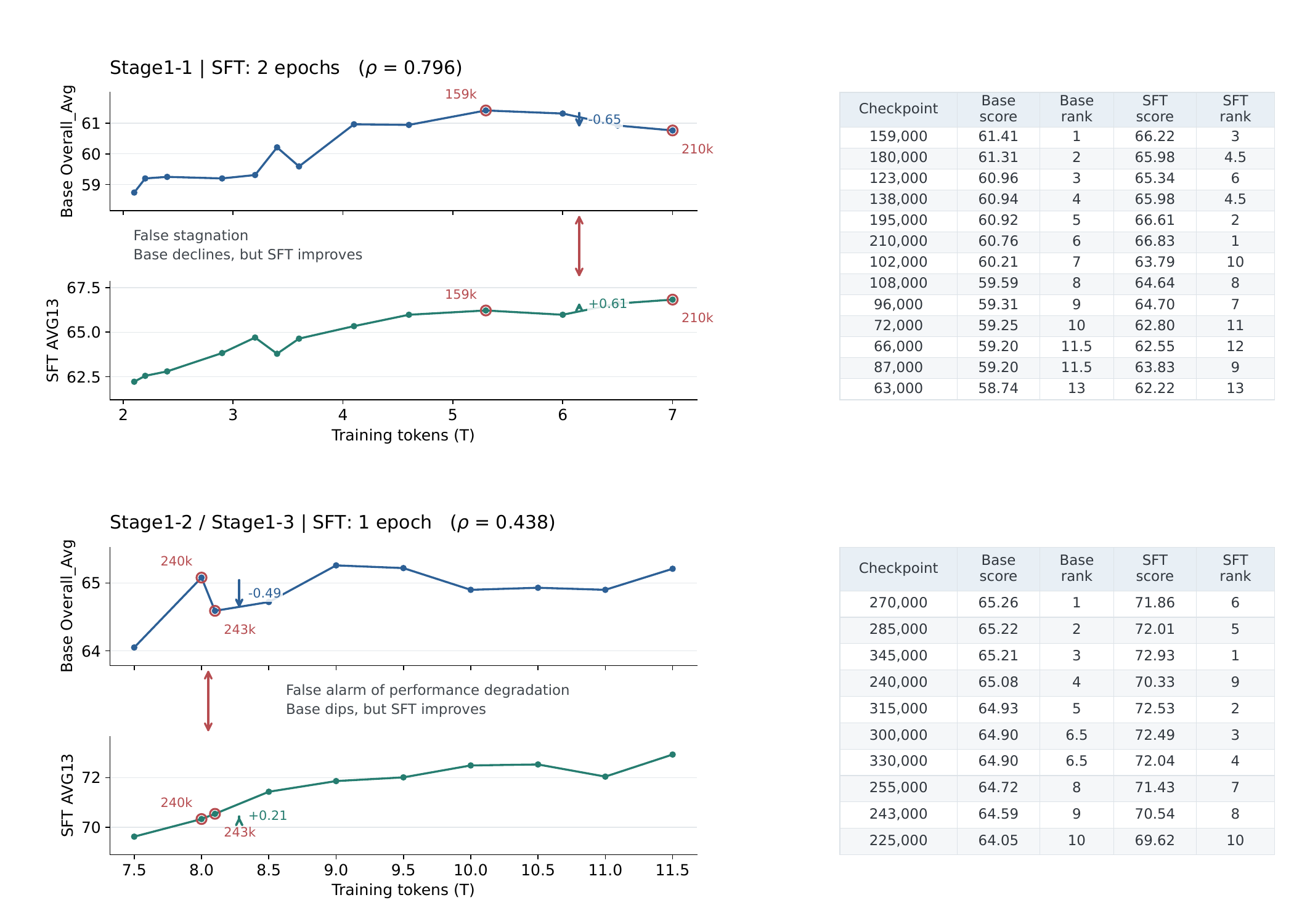}
\caption{Base and post-SFT checkpoint rankings are not interchangeable. \emph{Left}: base Overall\_Avg and SFT AVG13 for the same checkpoints. \emph{Right}: all paired checkpoints, sorted by base rank within each group, with their corresponding SFT ranks. Rank 1 is best, ties receive average ranks, and $\rho$ denotes Spearman correlation computed within each group from the reported scores. Base and SFT scores use different benchmark suites.}
\label{fig:base-sft-ranks}
\end{figure}

\begin{quote}
\noindent\textbf{(1) \textit{False stagnation.}}
We observed that base-model scores can appear to plateau,
even as the model continues to improve after SFT.
Figure~\ref{fig:false-stagnation} shows two intervals in
which average base-model scores fluctuate around a
near-plateau, while average scores after SFT continue
to improve.
In Stage1-1, from step 123,000 to 210,000 (4.1T--7T tokens),
the average base-model score remains near 61, while the
average score after two epochs of SFT increases from 65.34
to 66.83. In Stage1-3, from step 243,000 to 345,000
(8.1T--11.5T tokens), the average base-model score stays
around 64--66, while the average score after one epoch
of SFT rises from 70.54 to 72.93. It is worth noting that the base-model and post-SFT scores
are computed over different benchmark collections. We
therefore compare their trends within each evaluation
setting rather than their absolute values. These results show that relying only
on base-model evaluation could lead to premature
decisions to stop training or unnecessary changes
to the training recipe.

\noindent\textbf{(2) \textit{Ranking disagreement.}}
We also found that the relative ranking of checkpoints
can change after SFT, compared with their ranking under
base-model evaluation. To examine this difference,
we paired 23 pre-training checkpoints with their available
SFT results and compared their average base-model scores
across the 22 public and internal benchmarks listed in
Appendix~\ref{app:pretraining-benchmarks} with their average
scores after SFT (Figure~\ref{fig:base-sft-ranks}).
Since checkpoints early in training have weaker
instruction-following ability, we used the same Tiny-SFT
data but assigned different SFT budgets across training
stages. Specifically, we computed rankings separately for
13 Stage1-1 checkpoints evaluated after two SFT epochs
and 10 Stage1-2/Stage1-3 checkpoints evaluated after one
epoch. The corresponding Spearman rank correlations are
0.796 and 0.438. The rankings are positively associated,
particularly in Stage1-1, but they do not provide the
same assessment of checkpoint headroom.
For example, within the Stage1-1 group, step 159,000
ranks first under base-model evaluation but third after
SFT, whereas step 210,000 moves from sixth to first.
Between these checkpoints, the average base-model score
decreases from 61.41 to 60.76, while the average score
after SFT increases from 66.22 to 66.83.
These results show that base-model scores alone cannot
fully capture differences in post-training potential
across checkpoints.

\end{quote}

\begin{figure}[t]
\centering
\includegraphics[width=\linewidth]{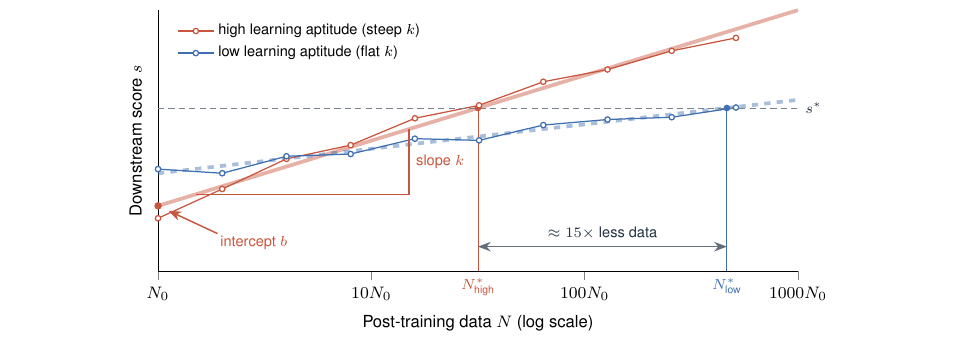}
\caption{A toy illustration of learning aptitude. Thin lines with hollow markers are raw scores at SFT checkpoints, and each straight line is their least-squares fit of Eq.~\eqref{eq:learnability}: the intercept $b$ is the score at the first checkpoint $N_0$, and the slope $k$ is the gain per tenfold increase in post-training data. The model with high learning aptitude starts lowest, yet reaches the target score $s^{*}$ with about $15\times$ less data than the baseline (Eq.~\eqref{eq:data-budget}). The high-start, flat model only reaches it at the end of the plotted budget.}
\label{fig:learnability}
\end{figure}

\textbf{Learning aptitude.}
A model's initial performance does not necessarily reflect how
effectively it can learn from additional post-training data.
We therefore measure its \emph{learning aptitude}, which captures
how efficiently a model improves as the post-training data
budget increases.

To measure learning aptitude, we select a specific domain or
task family and fine-tune different base models using the same
domain-specific SFT data. We evaluate a sequence of intermediate
checkpoints throughout training to track how each model's
performance changes as it sees more data. This allows us to fit
a learning curve and measure how efficiently the model learns
from additional data. For most models, scores grow approximately linearly with the
logarithm of the post-training data budget. We therefore fit
a log-linear learning curve for each base model $m$
(Figure~\ref{fig:learnability}):

\begin{equation}
  s_m(N) = b_m + k_m \log_{10}\frac{N}{N_0},
  \label{eq:learnability}
\end{equation}

where $s_m(N)$ is the downstream score after $N$ units of
post-training data (in practice, SFT steps at a fixed batch
size), and $N_0$ is the data budget of the first evaluated
SFT checkpoint. The intercept $b_m$ is the score at this
first checkpoint, representing the model's initial performance
in the domain. The slope $k_m$ measures the score gain for
every tenfold increase in post-training data, capturing how
efficiently the model learns from additional data. We estimate
both parameters by ordinary least squares over all SFT
checkpoints, excluding the untuned base model.

Neither parameter alone captures a model's learning
aptitude. A model with a high initial score but a shallow
learning curve starts strong but gains little from
additional data. In contrast, a model with a lower initial
score but a steeper learning curve may eventually surpass it
as the data budget increases. We therefore consider both
parameters together and estimate the amount of post-training
data required for model $m$ to reach a target score $s^*$:

\begin{equation}
  N_m^{*}(s^{*}) = N_0 \cdot 10^{(s^{*}-b_m)/k_m},
  \label{eq:data-budget}
\end{equation}

We then compare this data requirement with that of a baseline model, using the ratio $N^{*}_{\text{base}}/N^{*}_m$ to measure data savings. A larger ratio means that the model requires less post-training data to reach the same target performance.

In this work, we use learning aptitude to compare different
models rather than different checkpoints within a single
training run. As shown in Section~\ref{sec:evaluation},
a model with a lower initial score but higher learning
aptitude can eventually surpass a model with a higher initial
score after sufficient fine-tuning. Applying this measure
to checkpoint selection within a single training run is
a natural direction for future work.

\textbf{Max-over-protocols evaluation.}
A model's evaluation score does not always fully reflect its
acquired capabilities, as the extent to which these capabilities
are demonstrated can depend on the evaluation protocol. The same
benchmark may yield very different scores under perplexity
scoring, few-shot prompting, and zero-shot prompting
(Figure~\ref{fig:base-protocol-scores}; protocol definitions in
Appendix~\ref{app:base-eval-protocols}). Moreover, the relative
performance of these protocols can change across pre-training
checkpoints, as illustrated by HumanEval in
Figure~\ref{fig:base-humaneval-training}.

A single evaluation protocol may therefore underestimate model's capability. To obtain a more complete picture, we
evaluate each checkpoint under all available protocols and
take the highest score for each benchmark. Our goal is to
measure the capabilities a model can demonstrate, rather
than how well it adapts to a particular evaluation format.
Taking the maximum across protocols provides a more complete
estimate of the model's capability ceiling. For cross-model
comparisons in Section~\ref{sec:evaluation}, we follow the
evaluation protocols specified in that section.

\begin{figure}[t]
\centering
\captionsetup{skip=3pt}
\captionsetup[subfigure]{skip=2pt}
\begin{subfigure}{0.92\linewidth}
\centering
\includegraphics[width=\linewidth]{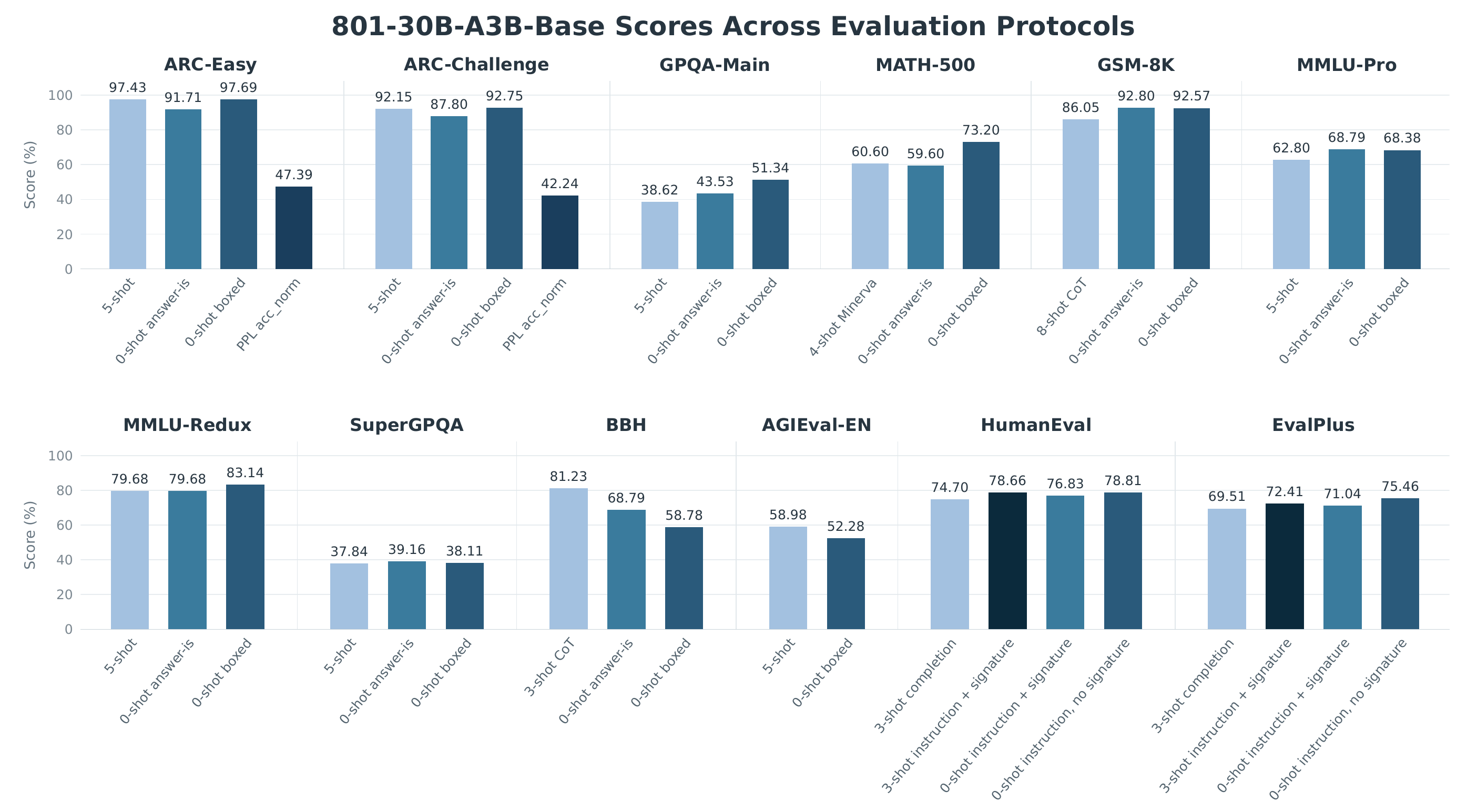}
\caption{\modelname Base scores across benchmarks under different
evaluation protocols. The same benchmark can yield different scores depending on the protocol, showing that a single
evaluation setting may not capture the capabilities a model
has acquired.}
\label{fig:base-protocol-scores}
\end{subfigure}
\par\vspace{2pt}

\begin{subfigure}{0.92\linewidth}
\centering
\includegraphics[width=\linewidth]{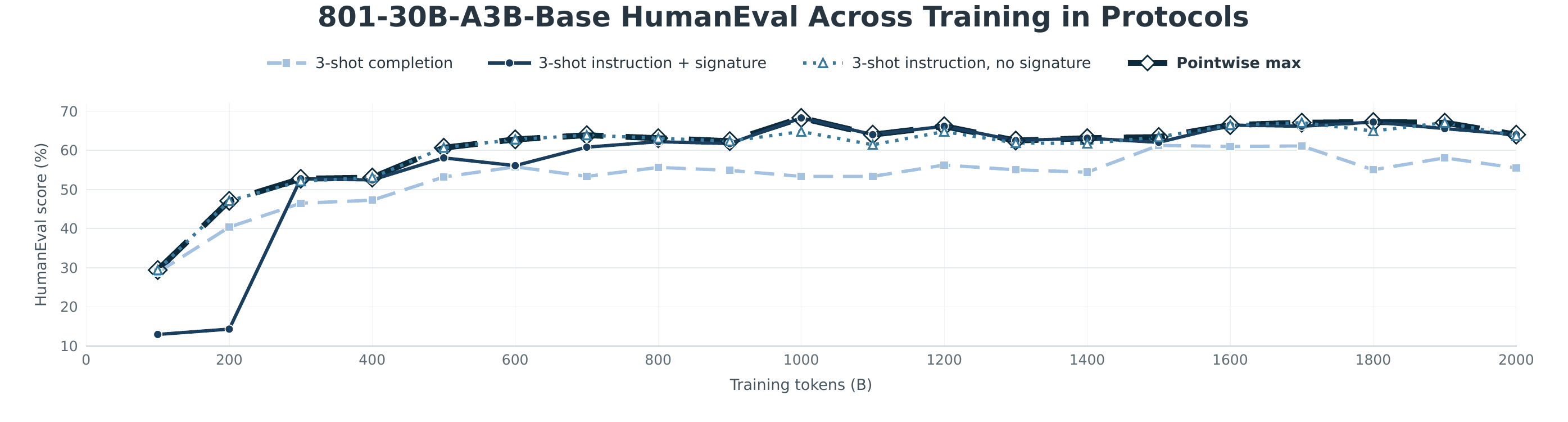}
\caption{HumanEval scores throughout pre-training under different
evaluation protocols. Both the scores and the relative ranking of
protocols change across checkpoints. We therefore take the pointwise
maximum across available protocols at each checkpoint to obtain a
more complete measure of the model's demonstrated capability.}
\label{fig:base-humaneval-training}
\end{subfigure}
\caption{Assessing base-model capabilities with max-over-protocols
evaluation.}
\label{fig:base-evaluation-protocols}
\end{figure}

\subsection{The Recipe Loop}
\label{sec:recipe-loop}

The \emph{recipe loop} brings post-training headroom
measurements into the pre-training process, using them as
feedback to assess training progress, guide recipe
adjustments, and inform subsequent iterations. The two
pilots, stopped at 1T and 3T tokens, and the subsequent
main pre-training run form the sequence of iterations
in this loop (Figure~\ref{fig:pilot-vs-formal}).

Throughout this process, we use headroom measurements to
track how the model's post-training potential evolves
across training stages and feed the results back into
the pre-training recipe. Rather than focusing solely
on current base-model scores, we consider post-training
performance, learning efficiency, and the capabilities
the model can demonstrate under different evaluation
protocols. This helps us judge whether pre-training is
still accumulating useful capabilities and whether
the current recipe needs further adjustment.

Based on the issues identified through earlier debugging,
we continue to refine the infrastructure and training
configurations while developing the training data.
In particular, main pre-training introduced new QA data
and revised the proportions and sampling weights of web,
code, mathematics, books, and papers.

The central idea is that pre-training progress should
not be judged solely by current base-model scores, but
also by whether the model continues to accumulate
capabilities that can be further developed through
post-training. Even when base-model evaluation appears
to plateau, pre-training may still be making meaningful
progress if post-training headroom continues to grow.

%% file: 4_agentic_pretrain.tex
\section{Experience-to-Metric Loops}
\label{sec:agentic_pretrain}

Many data-processing decisions in pre-training have
traditionally relied on expert experience, such as how
to clean and filter data, and how to judge whether a
processing strategy works well. We turn these
experience-driven tasks into standardized workflows
with explicit evaluation metrics. This allows agents
to analyze errors, revise their approaches, compare
alternatives, and improve their performance through
repeated iterations.

We call this the \emph{Experience-to-Metric Loop}, and
apply it to three tasks: classifier development,
site discovery, and crawl-coverage optimization.
All three follow the same principle: turn human
experience into measurable tasks, let agents carry
out and improve the work, and use evaluation results
to guide the next iteration.

\subsection{Classifier Development Loop}
\label{sec:datapipeline}
\begin{figure}[!htb]
    \centering
    \includegraphics[width=\linewidth]{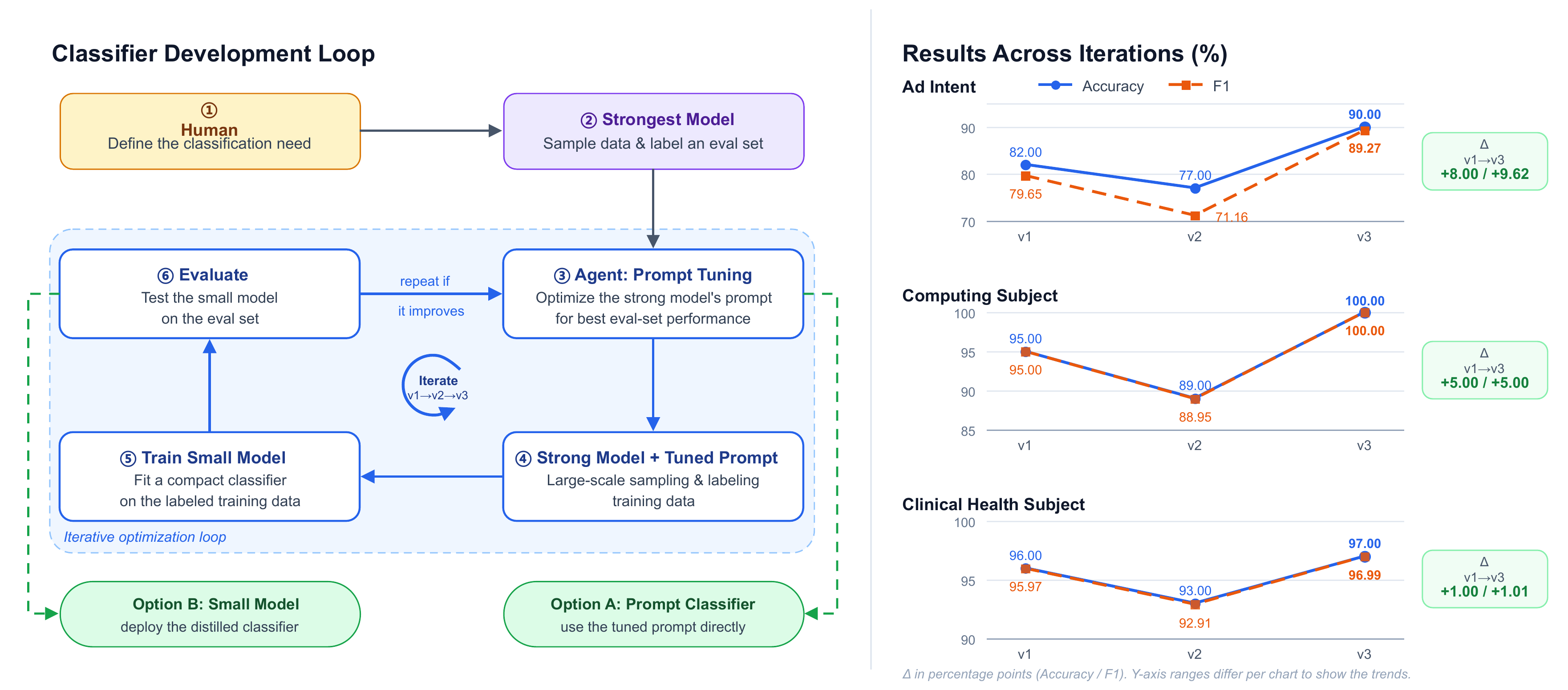}
    \caption{Iterative development of a data classifier. \emph{Left}: a human defines the classification task (1); a strong model samples data and annotates an evaluation set (2); an agent analyzes cases and tunes the labeling prompt (3); the strong model uses that prompt to label training data (4); a compact classifier is distilled from these labels (5) and evaluated (6). The component can be deployed as a tuned prompt (Option A) or a distilled model
    (Option B). \emph{Right}: accuracy
    and F1 over three iterations (v1--v3) for classifiers of ad intent,
    computing subject, and clinical health subject.}
    \label{fig:automated_classifier}
\end{figure}

Earlier work on \emph{Data Darwinism} provides a
framework for the first question. It introduces a
ten-level taxonomy (L0--L9) that organizes data
processing operations, from data acquisition and
format normalization to filtering, model-driven
refinement, and synthesis~\citep{qin2026datadarwinismi}.
Following its use in daVinci-LLM~\citep{qin2026davincillm},
we use this framework to guide pipeline design.
For example, OCR belongs to format normalization
(L1), rule-based and model-based filtering correspond
to L2--L3, and refinement for cleaning or
repairing content falls under L4. This taxonomy
helps us define what each component should do before
choosing its model, prompt, or implementation.

DataEvolve explores a complementary question: how data-curation strategies
can improve through repeated evaluation and diagnostic
feedback~\citep{mi2026dataevolve}. Inspired by this approach, we develop
classifiers, filters, OCR models, and rewriters through a shared evaluation
workflow.Data Darwinism
guides the choice of processing operations, while
the evaluation loop helps determine whether a
particular implementation performs its intended
task well. Figure~\ref{fig:automated_classifier} illustrates this process for
classifiers, with three stages (six steps in the figure).

\emph{Step 1: Define the task and construct an
evaluation set.}
A researcher specifies what the classifier
should do. A strong model then annotates
representative samples to create a reference
evaluation set. This establishes quality
criteria and provides a common basis for comparing prompts, models, and processing rules.
Since the reference labels come from a strong model,
the evaluation measures agreement with that model
rather than with human judgment.

\emph{Step 2: Analyze errors and improve the component.}
Agents analyze hundreds to thousands of examples
in parallel, identify recurring error patterns,
and use these findings to revise prompts, compare
models, or adjust processing rules. Each revision
is evaluated against the same task definition and
evaluation set, allowing us to determine whether
the changes actually improve performance.

\emph{Step 3: Validate and deploy.}
The revised component is evaluated before deployment. For a classifier, the
result can be used as a tuned labeling prompt or distilled into a
smaller model using training data labeled by the strong model. The distilled
classifier is evaluated and the loop continues revisions improve
its performance. The evaluation catches regressions: in
Figure~\ref{fig:automated_classifier}, all three classifiers lost accuracy at
v2 before improving at v3. 

In under three months, we built six data pipelines, with
agents contributing 36 model-selection decisions as well as prompt
optimization and classifier construction.

\subsection{Web Acquisition: Site Discovery and Crawl-Coverage Loops}
\label{sec:webdata}

\begin{figure}[!htb]
    \centering
    \includegraphics[width=\linewidth]{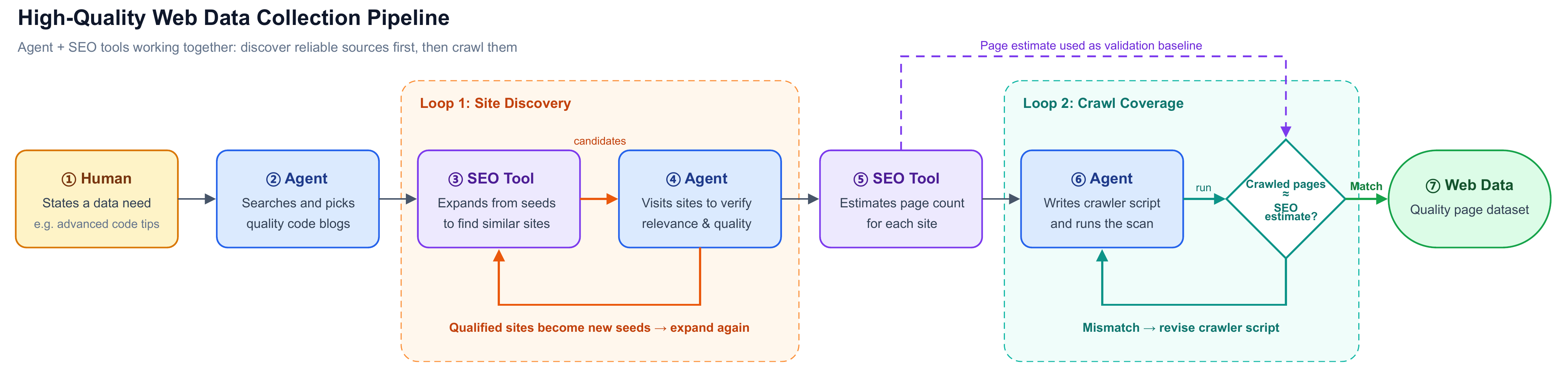}
    \caption{Structure-aware web data acquisition. A human requirement seeds an
    agent-driven pipeline with two loops. Loop~1 (site discovery) expands and verifies related
    sites through the web's link structure. Loop~2 (crawl coverage) crawls each confirmed site to
    a coverage target informed by its SEO page estimate. Links, search indexes, and
    robots directives are treated as structured signals rather than discarded.}
    \label{fig:webdata}
\end{figure}

The public web is a major source of pre-training data, but
systematically finding high-quality sources and collecting
their content as completely as possible remain challenging.
A common approach starts from seed documents, uses a
classifier to retrieve relevant pages, and identifies the
domains where these pages are concentrated~\citep{deepseekmath}.
However, two questions remain: how can we continuously discover
more valuable sites, and how can we tell whether we have
collected enough data from each site?

We address these questions with two connected agent-driven
loops: the \emph{site discovery loop}, which finds high-quality data sources, and the \emph{crawl-coverage loop},
which aims to collect their content as completely as possible.
Both use information from search-engine-optimization (SEO)
tools, which provide related-site information and estimates
of indexed page counts, as signals to guide discovery and
measure crawl coverage. Figure~\ref{fig:webdata} shows the pipeline.

\textbf{Loop1: Site Discovery Loop.}
The goal of site discovery is to expand a small set of high-quality seed sites into a broader collection
of relevant data sources. We use SEO tools to identify
candidate sites related to the seeds through the web's
link structure. An agent then visits each candidate to
check its relevance and quality, adding verified sites
to the seed set for the next round of discovery.
We measure the effectiveness of each round by its
\emph{verified yield}, the number of discovered
sites that pass verification. Each round's verified
sites become seeds for the next round, allowing discovery
to expand iteratively. The process stops when a round
produces no new qualified sites. This turns the
experience-based decision of where to look next into
a measurable and repeatable discovery process.

\textbf{Loop2: Crawl-Coverage Loop.}
Once valuable sites have been identified, the next
task is to collect their content as completely as
possible. For each confirmed site, an agent writes
and runs a crawler, then compares the number of
retrieved pages with the page-count estimate provided
by SEO tools.
We use the \emph{coverage ratio} as the evaluation
metric, defined as the number of retrieved pages
divided by the estimated total number of pages:

\[
\text{Coverage Ratio}
=
\frac{\text{Retrieved Pages}}{\text{Estimated Pages}}
\]

Although SEO page counts are estimates, they provide
a quantitative reference for assessing crawl coverage.
The agent compares the retrieved count with this
estimate, investigates potential gaps, revises the
crawler, and runs it again to measure the resulting
coverage. In this way, crawling becomes an iterative
process driven by measurable coverage rather than
an informal judgment of when a site is sufficiently
crawled.

\paragraph{Closing the loops.}

The site discovery and crawl-coverage loops address
two complementary challenges in web data acquisition:
where to find more valuable data sources and how to
collect their content as completely as possible.
Together with the classifier development loop, they
follow the same principle: turn experience-driven
tasks into workflows with explicit evaluation metrics,
so that agents can execute the work, assess the results,
and improve their approaches through repeated
iterations. Specifically, classifier development
uses task-specific evaluation scores, site discovery
uses the verified yield of each round, and web crawling
uses the coverage ratio.
The key is to give tasks that once relied on expert
judgment explicit objectives and evaluation procedures.
Agents can then use feedback to determine how well
a task has been completed and what to improve next.
This turns experience-based decisions into systematic
and repeatable processes, making pre-training more
scientific one component at a time.

%% file: 5_scaling_agentic_data.tex
\section{Scaling Agentic Data into Pre-Training}
\label{sec:agentic_data}

Agentic data, consisting of trajectories generated by agents
solving real-world tasks with verifiable outcomes, has proven
effective in post-training~\citep{pan2025swegym,yang2025swe}.
Real-world software development provides a rich source of
such tasks and offers a new opportunity for pre-training.
Our goal is to scale this data source to hundreds of billions
of tokens while preserving task diversity and data quality.

However, achieving this scale requires more than conventional
data-processing pipelines. Real-world software tasks are diverse and involve complex task construction, execution environments, testing, and integration across agent harnesses. We argue that \emph{extensive
use of agents is a necessary condition for scaling agentic
data into pre-training}. The main bottleneck is not
generating a handful of high-quality examples, but
systematically transforming diverse real-world software
development activities into executable, verifiable, and
scalable training data.

\begin{figure}[t]
    \centering
    \includegraphics[width=0.9\linewidth]{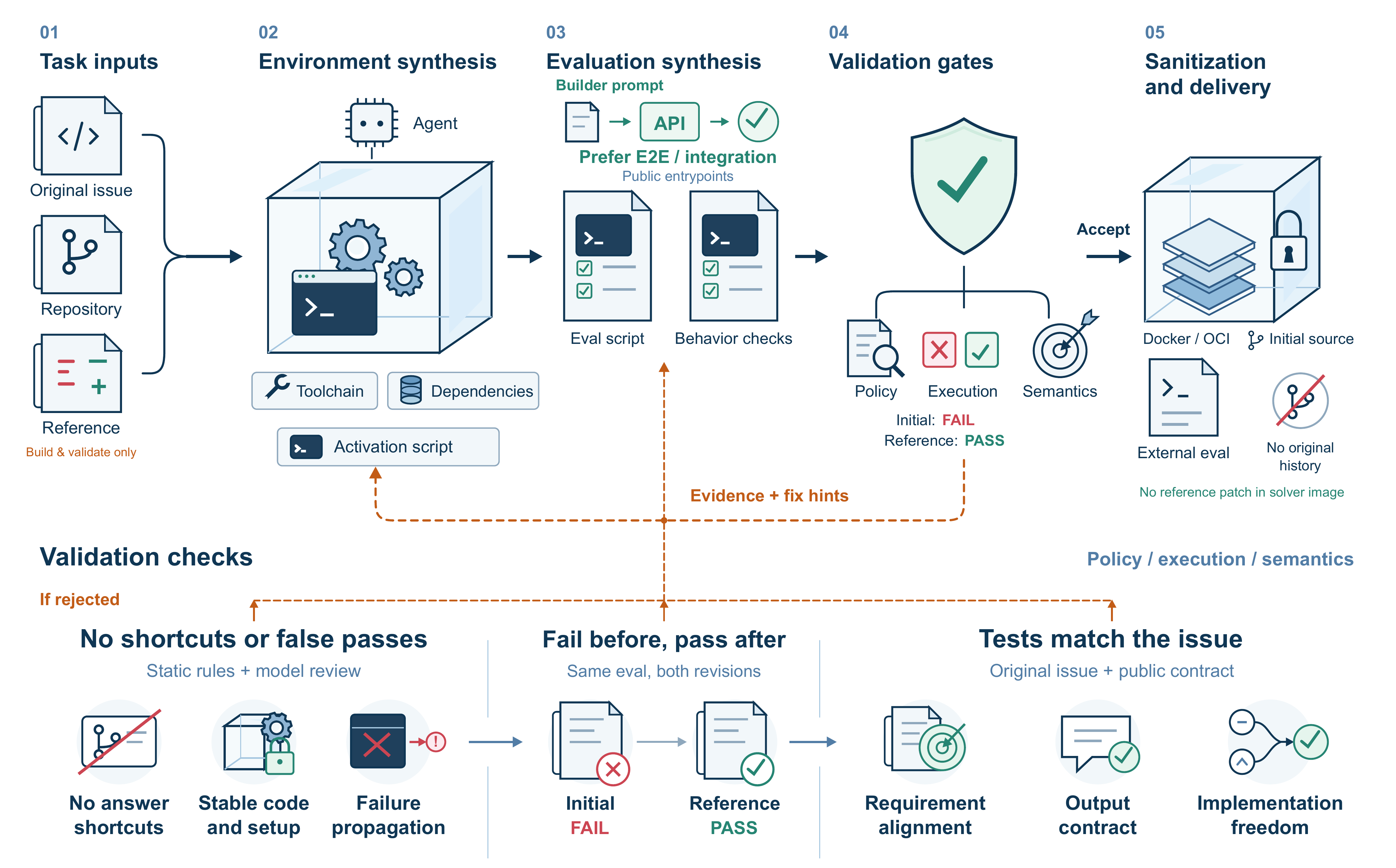}
    \caption{Overview of agentic-data pipeline. The key component is the Validation gates,  which admits a task only when it clears policy, execution, and semantic checks and the same evaluation fails on the initial code but passes on the reference, and sanitization into a Docker/OCI image that keeps the external eval but strips the reference patch and original history. Rejected tasks return with evidence and fix hints (dashed path). }
    \label{fig:agentic_pipeline}
\end{figure}

\subsection{Why Agentic Data Requires Agents}

\textbf{Complex task construction and validation.}
Real software development offers a rich source of tasks
for agentic data. Recent work has explored several 
directions: making development tasks executable through
automated environment and evaluation-script construction
~\citep{fu2026davincienv}, preserving development context
and execution feedback in mid-training trajectories
~\citep{zeng2026davinci}, and connecting related PRs into longer task chains
~\citep{jiang2026davinciagency}.

Scaling this data source to pre-training requires handling
the full diversity of real-world development activity.
This includes both PRs with clear task definitions and
well-maintained tests~\citep{jimenez2023swe,pan2025swegym,
badertdinov2025swe}, and those with incomplete descriptions,
implicit requirements, or multiple work items bundled
together.
For each PR, the pipeline must construct three mutually
consistent components: a reproducible execution
environment, an instruction specifying what the agent
is expected to do, and tests and evaluation scripts
that verify whether the task requirements are met.

These components must be developed together. For example,
a repository's CI may inject environment variables that
were omitted from the constructed environment, causing
the evaluation to fail for reasons unrelated to the code
change. In another case, a test may require a function
with a specific name, while the instruction describes
only the desired behavior, causing a valid implementation
to be rejected. A single PR may also bundle multiple
work items, leaving the instruction, environment, and
tests with inconsistent task boundaries.

Task construction therefore involves more than simply
extracting a task description from a PR. The environment,
instruction, and evaluation must agree on what the task
requires. Agents need to inspect repository context,
identify inconsistencies across these components, and
revise them accordingly. At scale, this process must
handle the long tail of repository-specific conventions
and development practices.

\begin{figure}[!htb]
    \centering
    \includegraphics[width=0.9\linewidth]{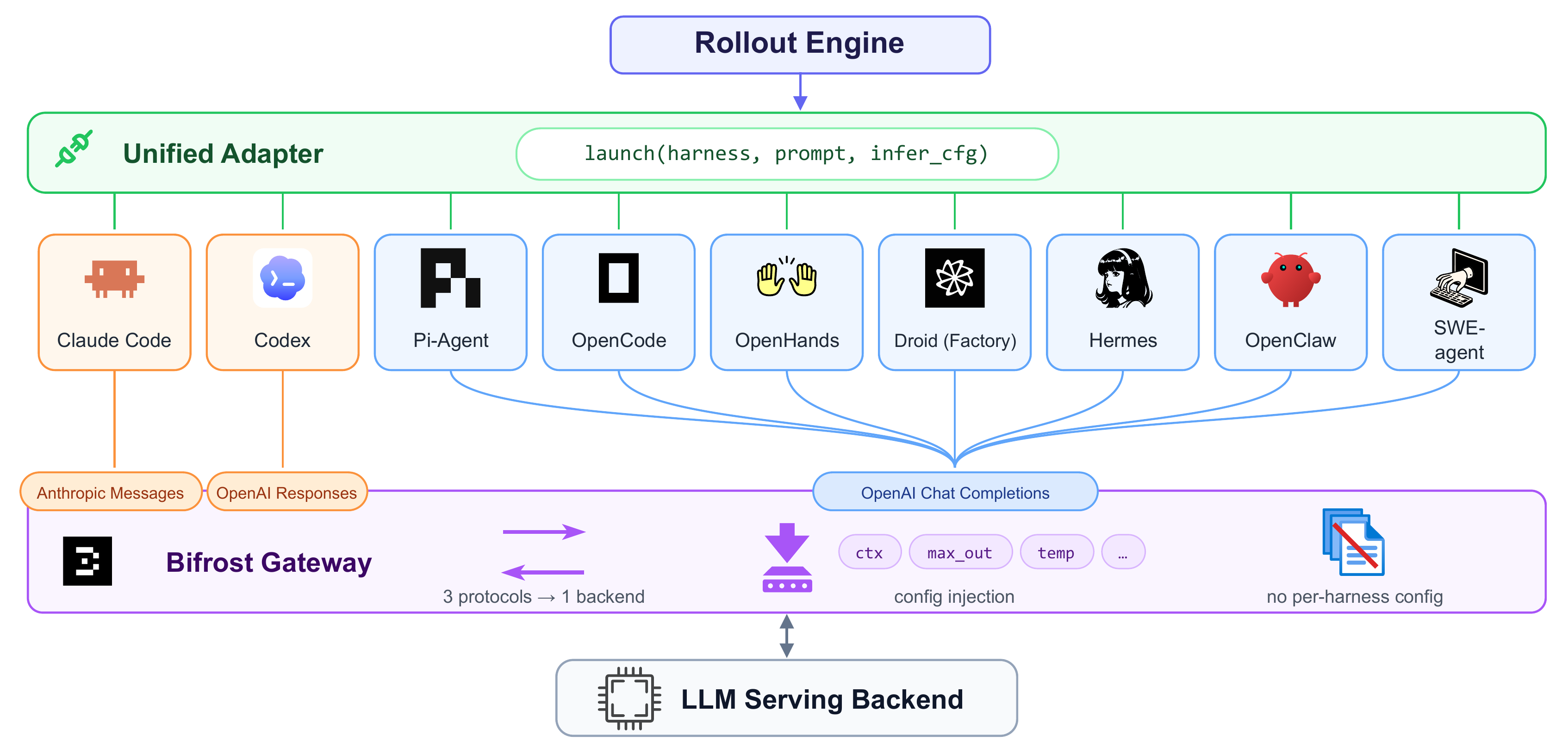}
    \caption{The multi-harness rollout layer. A single
\texttt{launch(harness, prompt, infer\_cfg)} call in the unified adapter drives
all nine integrated harnesses, which among them speak three
API protocols (Anthropic Messages~\citep{anthropic_messages_api}, OpenAI Responses~\citep{openai_responses_api}, and OpenAI Chat
Completions~\citep{openai_chat_api}). The Bifrost gateway~\citep{maxim2025bifrost} normalizes the three protocols onto one LLM serving backend and injects inference settings (context length, maximum output, temperature) centrally.}
    \label{fig:multi-harness}
\end{figure}

\textbf{Multi-harness integration and maintenance.}
Even after constructing a task, generating diverse
trajectories at scale requires supporting multiple
harnesses during execution. We integrated nine harnesses
into our agentic data pipeline: Claude Code
~\citep{anthropic2025claudecode}, Codex
~\citep{openai2025codex}, Pi-Agent~\citep{pi_agent},
OpenCode~\citep{opencode}, OpenHands
~\citep{wang2025openhands}, Droid
~\citep{factory2025droid}, Hermes
~\citep{nous2025hermesagent}, OpenClaw~\citep{openclaw},
and SWE-agent~\citep{yang2024swe}.
Each harness has its own interfaces, dependencies, and
failure modes. Even the same model can behave differently
across harnesses. In addition, harness updates and new
plugins introduce further configuration changes.

As a result, the combinations of models, harnesses, and
versions require ongoing integration and maintenance.
We use a unified adapter to launch rollouts across
different harnesses, with agents helping diagnose
execution failures, revise integrations, and validate
updated configurations. As these combinations grow
and change, manual maintenance becomes unsustainable.
Agents are therefore necessary for sustaining
large-scale trajectory generation.

\begin{table}[!htb]
\centering
\small
\caption{Agentic anti-hack patterns, grouped by category. Each is a concrete way
a model can pass a synthesized task without solving it. Agents scan trajectories
for these behaviors and flag candidates for review or re-synthesis.}
\renewcommand{\arraystretch}{1.2}
\begin{tabular}{@{}>{\raggedright\arraybackslash}p{0.26\linewidth}p{0.66\linewidth}@{}}
\toprule
\textbf{Category} & \textbf{Specific behavior} \\
\midrule
\multirow{2}{=}{Pull the reference fix from version control}
 & Fetch its own merged PR diff or the fixing commit patch \\
 & Recover the fix from unreachable Git objects (\texttt{git fsck}, \texttt{cat-file}) \\
\midrule
\multirow{2}{=}{Fetch upstream code over the network}
 & Download files from \texttt{raw.githubusercontent}, the GitHub API, or PyPI JSON \\
 & Clone, fetch, or add a remote pointing at the upstream repository \\
\midrule
\multirow{2}{=}{Install an already-fixed version}
 & \texttt{pip download} or \texttt{install -{}-target} with a pinned version range \\
 & Pull packages through conda, apt, or an external mirror (aliyun, tuna, ustc) \\
\midrule
\multirow{2}{=}{Find a solved copy already on disk}
 & Read installed, cached, or \texttt{site-packages} source copies \\
 & Scan from root with \texttt{find}, \texttt{grep}, or \texttt{rg}, or walk the conda tree \\
\midrule
\multirow{2}{=}{Tamper with the evaluation}
 & Write or rewrite test files \\
 & Deselect or skip the failing tests (\texttt{pytest -{}-deselect}) \\
\midrule
Reason then exfiltrate
 & Plan in chain-of-thought to leave the sandbox, then issue a network call \\
\bottomrule
\end{tabular}
\label{tab:antihack}
\end{table}

\textbf{Trajectory quality control and anti-hacking.}
Anti-hacking is critical to producing high-quality
agentic data. Models can exploit evaluation shortcuts
in many ways~\citep{zhong2025impossiblebench,
kahn2025swebenchleak}, and agents are essential for
detecting such behavior at scale.
In a data synthesis pipeline, agents must distinguish
between two types of problems: \emph{hacks}, where a
model bypasses the task rather than genuinely solving
it, and \emph{synthesis defects}, where the execution
environment, tests, or task instructions themselves
are flawed. This distinction is particularly difficult
because a model's cheating behavior can mask defects
in the task, making an invalid task appear solvable.
To address this, agents scan execution trajectories
for anomalous behaviors, such as accessing Git
repositories and logs, downloading related packages,
searching for code commits, or using shell commands
that bypass the actual task logic. Suspicious cases
are flagged for human review or re-synthesis
(Table~\ref{tab:antihack}).

Thus, producing agentic data at scale requires not
only making tasks executable, but also verifying
that models genuinely solve them and that the
resulting trajectories are trustworthy.

\subsection{Scaling Agentic Data for Pre-Training}

Through the task-construction, multi-harness execution,
and quality-control processes described above, we
established a pipeline for producing agentic data at scale.

\begin{figure}[!htb]
    \centering
    \includegraphics[width=0.9\linewidth]{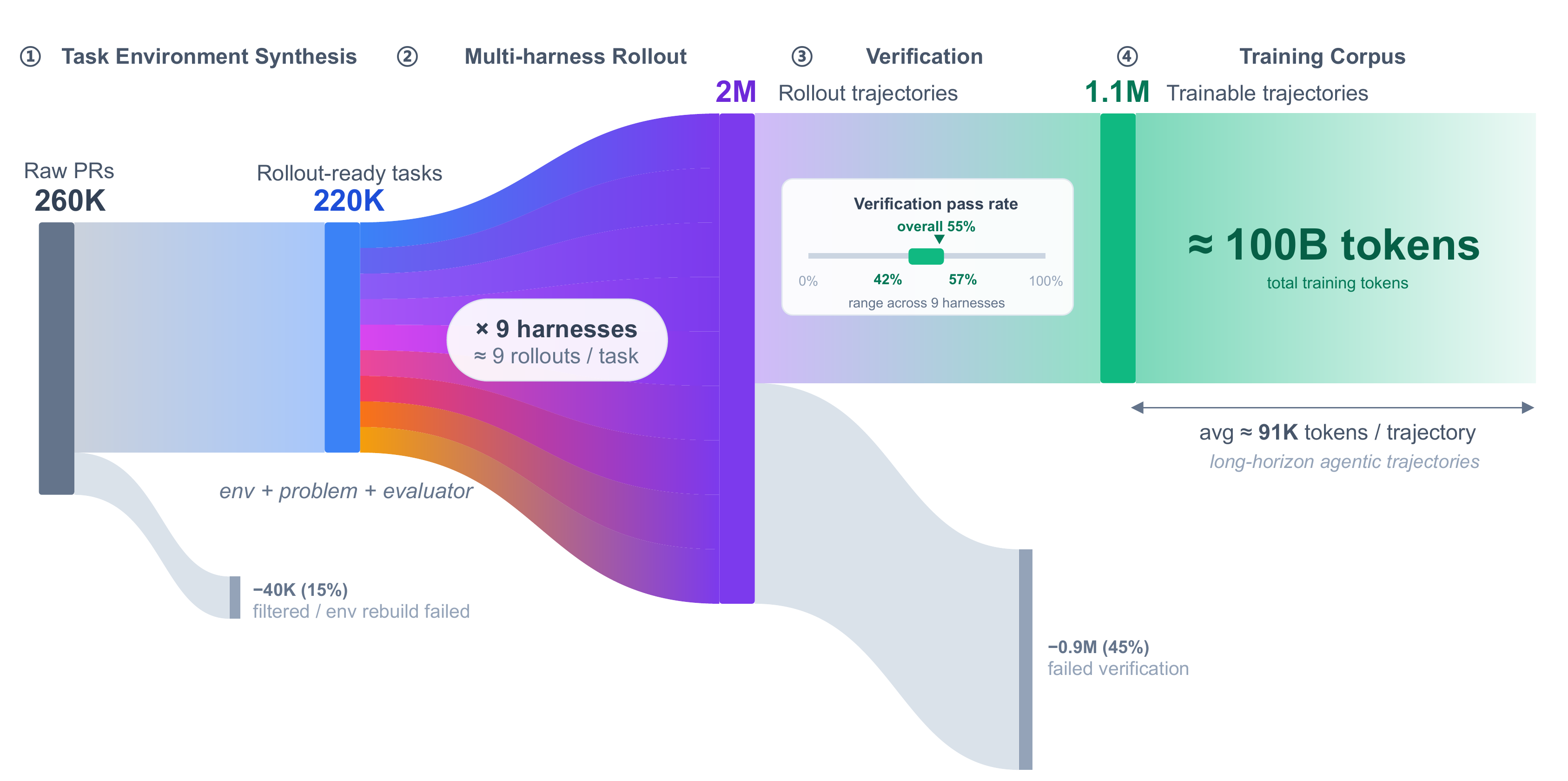}
    \caption{Funnel chart of agentic data pipeline. 260K raw PRs are filtered into 220K executable tasks. Rolling each task out under 9 harnesses gives about 2M trajectories. After verification, 1.1M trajectories remain, totaling more than 100B training tokens (about 91K tokens per trajectory on average). Tasks and trajectories are drawn on separate scales.%
    }
    \label{fig:agentic_funnel}
\end{figure}

More than 84\% of the raw pull requests we collected
were successfully converted into executable tasks.
For these PRs, we constructed execution environments,
extracted task requirements, and synthesized
corresponding tests (Figure~\ref{fig:agentic_funnel}).
We then rolled out the resulting 220K executable PRs
across nine harnesses. The trajectories were generated
using small and medium-sized open-weight models,
including DeepSeek-Flash and GPT-OSS.
After layered filtering, we obtained 1.1M agent
trajectories that met our pre-training quality
standards, totaling more than 100B tokens, with
approximately 91K tokens per trajectory on average.
Together with trajectories from other diverse sources,
the total volume of agentic data available for
pre-training reached \textbf{134B} tokens. Importantly, these trajectories are not merely an
auxiliary ingredient for post-training, but a
\emph{first-class data source for pre-training}.

Agent-driven data pipelines represent a shift in how
training data is produced. As models become more capable,
they can solve harder tasks and generate richer training
signals. At the same time, as harnesses and tools evolve,
agents can adapt the underlying integration layer.
Beyond adding more harnesses, the plugin ecosystems
within each harness are also growing rapidly, further
expanding the space of configurations that must be
supported.

We expect agent-driven data pipelines to become an
increasingly important source of high-quality, diverse,
and verifiable training data for future frontier
pre-training runs.

%% file: 6_training_recipe.tex
\section{Training Recipe of \modelname}

\begin{figure}
\centering
\includegraphics[width=\linewidth]{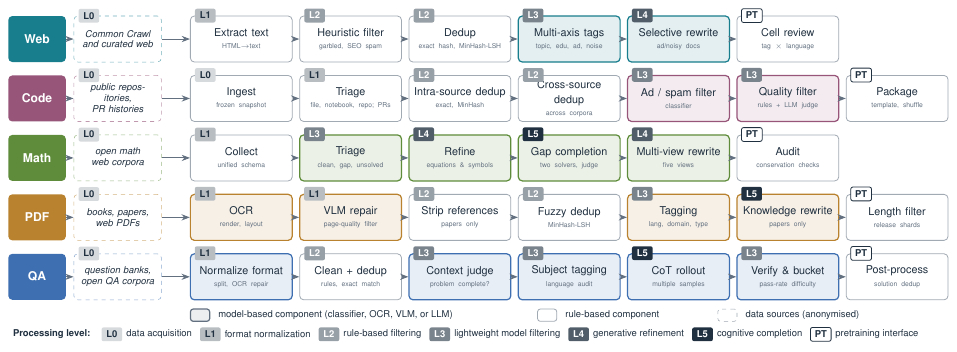}
\caption{Pretrain data pipelines except the agentic data pipeline. Filled
boxes mark model-based components such as classifiers, OCR, VLMs, and LLM
rewriters, which we build and tune with the workflow in
Figure~\ref{fig:automated_classifier}~\citep{qin2026datadarwinismi}.}
\label{fig:data-pipelines}
\label{fig:label}
\end{figure}

\begin{figure}
\centering
\includegraphics[width=\linewidth]{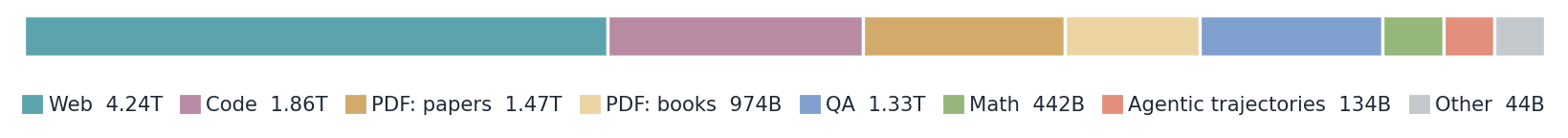}
\caption{Composition of the pre-training corpus, measured in
unique tokens that entered training (10.5T in total).}
\label{fig:data-distribution}
\end{figure}

\newsavebox{\recipetablebox}
\begin{figure}[!htb]
\centering
\sbox{\recipetablebox}{%
\scriptsize
\setlength{\tabcolsep}{2.5pt}%
\renewcommand{\arraystretch}{1.5}%
\begin{tabular*}{0.56\linewidth}{@{\extracolsep{\fill}}lccccc@{}}
\toprule
& \multicolumn{3}{c}{Stage1 (11.5T)}
& \multicolumn{2}{c}{Ctx.\ extension (300B)} \\
\cmidrule(lr){2-4}\cmidrule(l){5-6}
\textbf{Setup}
& \textbf{1-1}
& \textbf{1-2}
& \textbf{1-3}
& \textbf{Stage2}
& \textbf{Stage3} \\
\midrule
Tokens added & 7T & 1T & 3.5T & 100B & 200B \\
Seq.\ length & 4K & 4K & 4K & 32K & 256K \\
RoPE $\theta$ & $10^4$ & $5\times10^5$ & $5\times10^6$ & $5\times10^6$ & $5\times10^6$ \\
Learning rate & $5\times10^{-4}$ & $5\times10^{-4}$ & $5\times10^{-4}$ & $5\times10^{-4}$ & \begin{tabular}[c]{@{}c@{}}$5\times10^{-4}$\\$\to5\times10^{-5}$\end{tabular} \\
Warmup steps & 2,000 & No & No & 1,000 & 1,000 \\
LB loss coef. & $10^{-4}$ & $10^{-4}$ & $10^{-4}$ & Off & Off \\
Router z-loss & $10^{-3}$ & $10^{-3}$ & $10^{-3}$ & Off & Off \\
\bottomrule
\end{tabular*}}
\begin{subfigure}[t]{0.56\linewidth}
\vspace{0pt}
\centering
\usebox{\recipetablebox}
\caption{Training stages and key recipe changes.}
\label{tab:training-stages}
\end{subfigure}\hfill
\begin{subfigure}[t]{0.41\linewidth}
\vspace{0pt}
\centering
\includegraphics[height=\dimexpr\ht\recipetablebox+\dp\recipetablebox\relax]{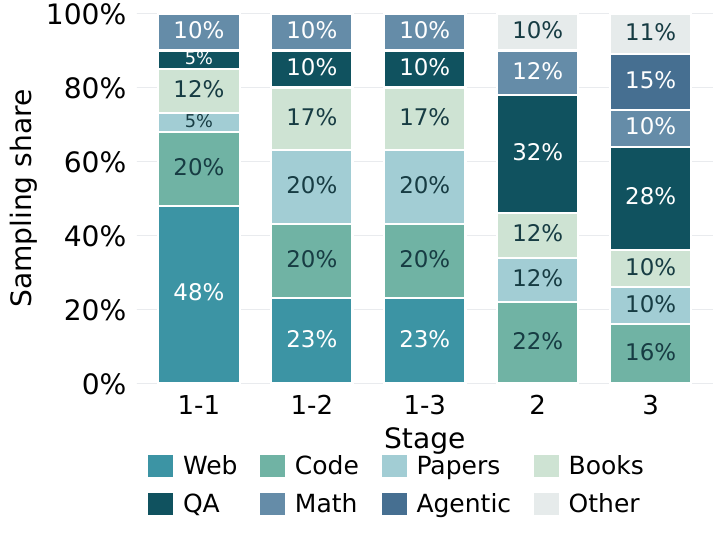}
\caption{Configured data sampling proportions.}
\label{fig:stage-data-mixture}
\end{subfigure}
\caption{Training recipe of \modelname{} across the five training stages.
\emph{(a)} Key recipe changes, including the progression from 4K to 32K and
256K sequence lengths and the RoPE-base updates during Stage1.
\emph{(b)} Configured data sampling proportions. Stage1-2 and Stage1-3 retain
the same category-level mixture despite updates to their data sources. Other
denotes capability-targeted data.}
\label{fig:training-recipe}
\end{figure}

We adopt the Qwen3-30B-A3B~\citep{yang2025qwen3} Mixture-of-Experts architecture, with approximately
30 billion total parameters and 3 billion active parameters per token.
Figure~\ref{tab:training-stages} summarizes the selected training recipe:
three pre-training sub-stages followed by two context-extension stages.
All stages use a global batch of 33,554,432 tokens and Muon~\citep{jordan2024muon,liu2025muon}, with AdamW~\citep{loshchilov2019adamw} for
the excluded parameter groups, including embeddings, the output head,
normalization parameters, biases, and routers.

Stage1 maintains a sequence length of 4,096 tokens. Stage1-1 trains from
scratch through step 210,000 (approximately 7T tokens); Stage1-2 continues
to step 240,000 (approximately 8T), and Stage1-3 supplies the checkpoint at
step 345,000 (approximately 11.5T
). At these transitions, the data mixture
is updated and the RoPE~\citep{su2024rope} base increases from $10^4$ to $5\times10^5$ and $5\times10^6$, following the base-frequency adjustment for longer contexts~\citep{xiong2024longcontext}. The learning rate remains $5\times10^{-4}$ after an 2,000-step warmup; Stage1-2 and Stage1-3 resume the optimizer and
training state without a warmup. Stage1 uses global-batch load balancing~\citep{fedus2022switch,qiu2025globalbatch} with coefficient
$10^{-4}$ and router z-loss~\citep{zoph2022stmoe} with coefficient $10^{-3}$.

Stage2 extends the sequence length to 32,768 tokens and is configured for
3,000 steps (approximately 100B tokens). Stage3 further extends it to 262,144 tokens
and is configured for 6,000 steps (approximately 200B tokens). Both retain the RoPE base
of $5\times10^6$, initialize from the preceding stage's weights with fresh
optimizer state, use a 1,000-step warmup, and disable both load-balancing
and router z-loss penalties. Stage2 keeps the learning rate at
$5\times10^{-4}$ after warmup, whereas Stage3 applies cosine decay to
$5\times10^{-5}$. Stage3 also enables length-aware packing. The three stages together correspond
to approximately %
11.8T tokens.

The training corpus draws from multiple data types (shown in Figure~\ref{fig:data-distribution}): web text, PDF documents,
question-answer pairs, code repositories, and agentic trajectories. Beyond the
agentic data pipeline detailed in Section~\ref{sec:agentic_data}, each major
category branches into sub-pipelines (shown in Figure~\ref{fig:data-pipelines}). \emph{Web data} is
cross-classified by advertisement presence, content type tags, domain category,
and language, creating a Cartesian product of fine-grained cells that serve as
the atomic units for review and mixture tuning. \emph{Math data} undergoes
model-driven rewriting to increase diversity and refine equation and symbols. \emph{PDF data}
passes through OCR extraction, vision-language model (VLM) refinement to
correct OCR errors, and multi-stage quality filters. \emph{QA data} pipelines
classify whether the problem's context is complete or not and rollout the CoT with monitoring pass rate to probe their
difficulty. \emph{Code data} includes static code files and pull-request records.

Figure~\ref{fig:stage-data-mixture} shows how the configured sampling mixture
changes across training stages. These proportions describe sampling weights,
whereas Figure~\ref{fig:data-distribution} reports unique corpus tokens.
Stage1-1 assigns 48\% to web text and 20\% to code. In Stage1-2 and
Stage1-3, the web share falls to 23\%, while papers, books, and QA increase
to 20\%, 17\%, and 10\%, respectively. These two sub-stages share the same
category-level proportions, although their underlying data sources differ.
Stage2 removes the dedicated web component, raises QA to 32\%, and assigns
10\% to capability-targeted data. Stage3 introduces a 15\% agentic component
and assigns 11\% to capability-targeted data, including instruction-following
and long-context examples. The capability-targeted category also includes
retrieved web, book, and paper material, so the absence of a dedicated web
component does not imply the absence of all web-derived content.

\subsection{Attention-Balanced Data Sampling}
Mixing long sequences into training data can cause throughput degradation far beyond what their proportion would suggest. In theory, the expected training time should be the proportion-weighted sum of the time spent on long and short samples. In practice, adding only 10\% long samples can halve the training speed. 

Using the context infrastructure, the agent traces this unexpected slowdown to computation bubbles caused by imbalanced attention workloads across data-parallel (DP) ranks. 
In synchronous data-parallel training, ranks with heavier attention workloads become stragglers, leaving other ranks idle at synchronization points. This imbalance is particularly severe for packed sequences, where attention cost depends on individual document lengths rather than total token count. We estimate each pack's cost as where $m_p$ counts document segments and $\ell_{p,j}$ is the length of segment $j$.
\begin{equation}
  \widehat{C}(p)=\sum_{j=1}^{m_p}\ell_{p,j}^{2},
\end{equation}

To mitigate this imbalance, the dataloader sort packs by estimated attention cost within a bounded window and group similarly costly packs into global microbatches. When each DP rank receives multiple packs, we further assign packs in descending cost order to the least-loaded rank with remaining capacity (Figure~\ref{fig:scheduling-concise}a).

The sorting window limits sample reordering. Larger windows improve load
balance (Figure~\ref{fig:scheduling-concise}b), but document length may be
associated with data source and difficulty. Aggressive sorting can therefore
bias the per-step mixture and substantially increase loss fluctuations early in long-context
training. We used a limited sorting window with a buffer of roughly twice the
distributed world size to balance computational efficiency and sample diversity.

\begin{figure}
  \centering
  \input{figures/scheduling_tradeoff}
  \caption{(a) Grouping packs by attention cost reduces imbalance; values are
  illustrative. (b) Imbalance in our 256K corpus across sorting windows,
  using 128 DP ranks and one pack per rank. The ratio is maximum to mean rank
  cost, averaged across microbatches; 1 denotes perfect balance.}
  \label{fig:scheduling-concise}
\end{figure}
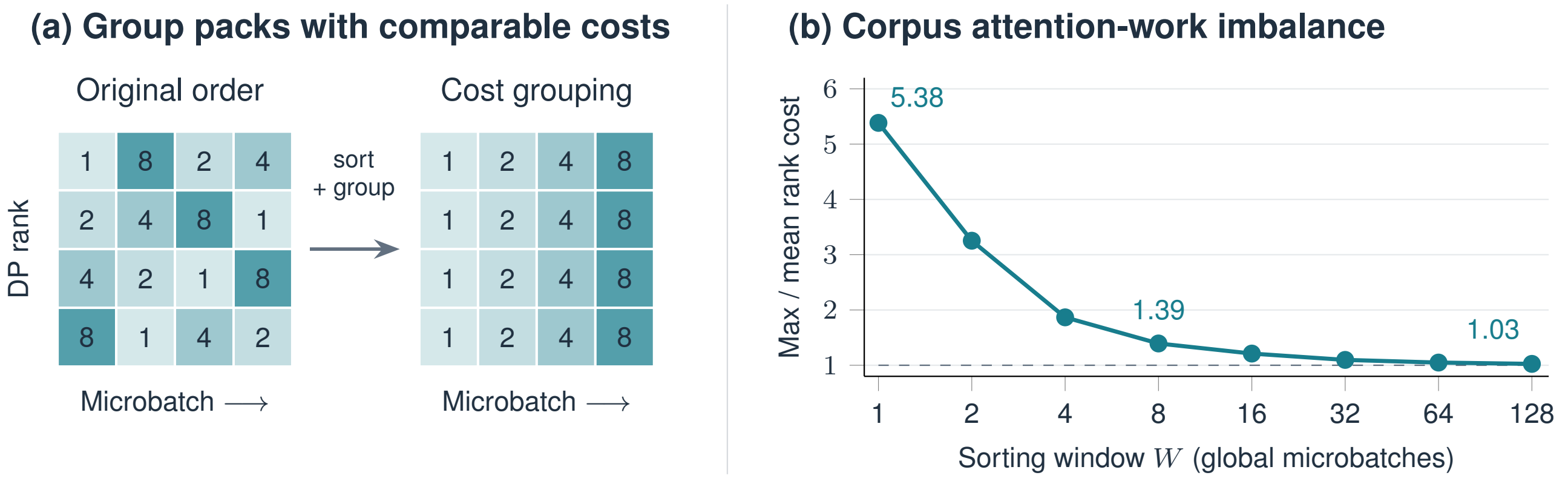
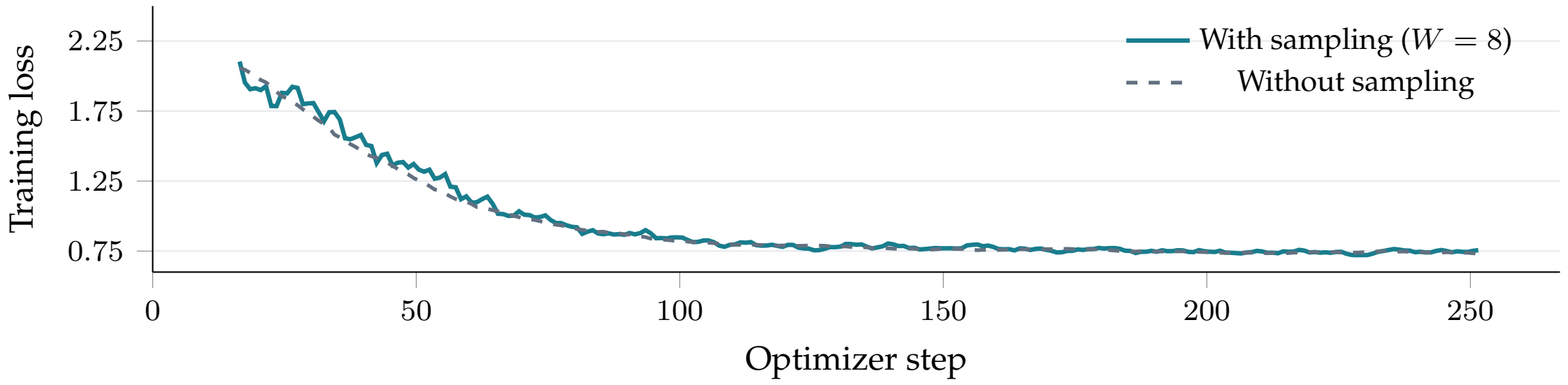
\begin{figure}
  \centering
  \input{figures/smoothed_loss}
  \caption{Training loss with and without attention-balanced data sampling
  (32-step moving average).}
  \label{fig:smoothed-loss-concise}
\end{figure}

To evaluate this strategy, we compared two 256K-context pretraining runs on
256 GPUs with matching configurations. We evaluate the proposed dataloader 
on 256 GPUs with context parallelism of 4 (CP=4), yielding 64 data-parallel (DP) replicas. 
With one pack per DP replica and a sorting window of eight global microbatches, our 
method achieves a $\SchedulingSpeedup\times$ throughput improvement (from 1,126 to 2,543 TGS). 
Smoothed training losses follow similar trajectories (Figure~\ref{fig:smoothed-loss-concise}).
Because this scheduling alters the global sample order, we
evaluated final model quality separately, and benchmark
differences with and without scheduling remained within run-to-run fluctuations.

\subsection{Optimizer Allocation}
We use Muon for expert and attention weight matrices, and AdamW for router,
embedding, output, normalization, and bias parameters. To examine this
allocation alongside the optimizer diagnosis in
Table~\ref{tab:diagnoses}, we compare four configurations from our second
pre-training pilot (Table~\ref{tab:optimizer_allocation}). Three use global
load balancing and vary the parameters assigned to Muon; a fourth uses the
same expert-and-attention allocation with local load balancing. All four
use the same configured data mixture, a sequence length of 4,096, a global
batch of 8,192 sequences (33.55M tokens), and a learning rate of
$5\times10^{-4}$ after 2,000 warmup steps. The load-balancing and router
z-loss coefficients are $10^{-4}$ and $10^{-3}$, respectively.

\begin{table}[t]
\centering
\small
\caption{Optimizer allocation in the second pre-training pilot. E denotes
expert matrices and E-A denotes expert and attention matrices. LBL denotes
the load-balancing scope. All Muon variants use RMS matching to AdamW~\citep{liu2025muon},
momentum 0.95, and five Newton--Schulz iterations; AdamW uses
$(\beta_1,\beta_2)=(0.9,0.95)$. Weight decay is 0.1. Router parameters use
AdamW in every setting.}
\setlength{\tabcolsep}{4pt}
\renewcommand{\arraystretch}{1.15}
\begin{tabular*}{\linewidth}{@{\extracolsep{\fill}}lccccc@{}}
\toprule
Setting & Experts & Attention & Router &
Embedding/output/norm/bias & LBL \\
\midrule
AdamW--Global & AdamW & AdamW & AdamW & AdamW & Global \\
Muon-E--Global & Muon & AdamW & AdamW & AdamW & Global \\
Muon-E-A--Global & Muon & Muon & AdamW & AdamW & Global \\
Muon-E-A--Local & Muon & Muon & AdamW & AdamW & Local \\
\bottomrule
\end{tabular*}
\label{tab:optimizer_allocation}
\end{table}

\begin{figure}[t]
\centering
\includegraphics[width=0.9\linewidth]{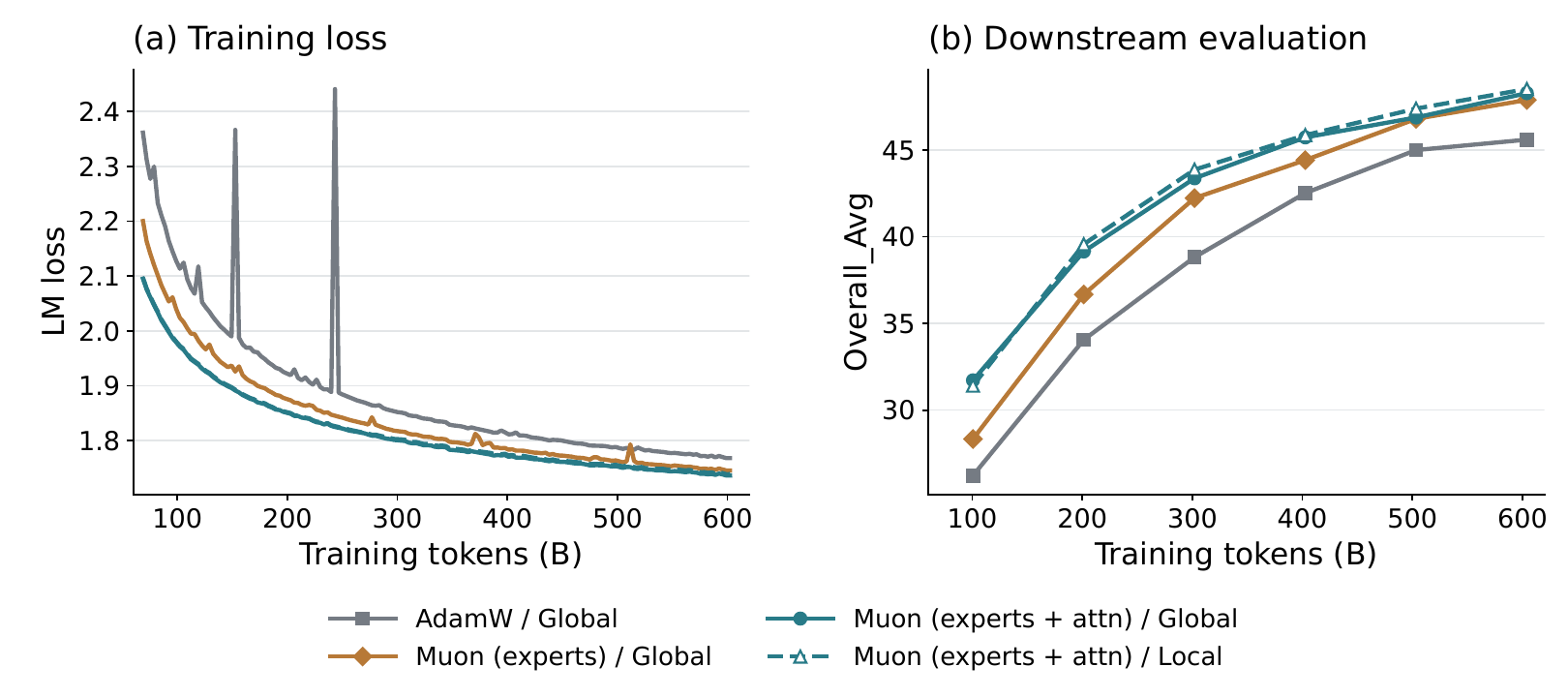}
\caption{Optimizer allocation and load-balancing comparisons for the four settings in Table~\ref{tab:optimizer_allocation}. 
\emph{Left}: training LM loss, excluding auxiliary losses, averaged in consecutive 100-step bins after the 2,000-step warmup. \emph{Right}: direct base-model Overall Avg scores at 100B-token intervals. The two expert-and-attention settings share a color, with a dashed line denoting local load balancing. Both global Muon allocations improve on all-AdamW in this interval, while the local/global score differences are small.}
\label{fig:optimizer_ablation}
\end{figure}

Figure~\ref{fig:optimizer_ablation} compares these runs over their evaluation interval, ending at step 18,000 (604B tokens). Under global load
balancing, both hybrid configurations have higher downstream scores than
all-AdamW at every evaluated checkpoint. At the endpoint,
\texttt{Overall\_Avg} is 45.60 for all-AdamW, 47.89 for Muon on experts,
and 48.28 for Muon on experts and attention. Their mean LM losses over the
last 100 steps are 1.768, 1.746, and 1.737, respectively. Extending Muon to
attention provides an improvement in these runs, although the
score gap narrows with training. The local and global expert-and-attention
variants remain close: their endpoint scores are 48.52 and 48.28.
These observations support the hybrid parameter allocation. Downstream
evaluation, however, does not distinguish local from global load balancing
over this interval. We nonetheless adopt global load balancing for main
pre-training because its internal metrics are more stable. Each setting is
represented by a single training run.

%% file: figures/scheduling_tradeoff.tex
\resizebox{\linewidth}{!}{%
\begin{tikzpicture}[font=\sffamily\small,>=Stealth,text=scheduleInk]
  \path[use as bounding box] (-0.2,-0.4) rectangle (16.2,4.5);
  \node[anchor=west,font=\sffamily\bfseries] at (0,4.2)
    {(a) Group packs with comparable costs};
  \node at (1.55,3.55) {Original order};
  \node at (5.3,3.55) {Cost grouping};
  \foreach \r/\a/\b/\c/\d in {0/1/8/2/4,1/2/4/8/1,2/4/2/1/8,3/8/1/4/2}{
    \foreach \col/\v in {0/\a,1/\b,2/\c,3/\d}{
      \pgfmathtruncatemacro{\shade}{10+8*\v}
      \filldraw[fill=scheduleTeal!\shade,draw=white,line width=0.7pt]
        ({0.4+\col*0.6},{2.55-\r*0.6}) rectangle ++(0.6,0.6);
      \node[font=\sffamily\footnotesize] at
        ({0.7+\col*0.6},{2.85-\r*0.6}) {\v};
    }
    \foreach \col/\v in {0/1,1/2,2/4,3/8}{
      \pgfmathtruncatemacro{\shade}{10+8*\v}
      \filldraw[fill=scheduleTeal!\shade,draw=white,line width=0.7pt]
        ({4.1+\col*0.6},{2.55-\r*0.6}) rectangle ++(0.6,0.6);
      \node[font=\sffamily\footnotesize] at
        ({4.4+\col*0.6},{2.85-\r*0.6}) {\v};
    }
  }
  \node[rotate=90,font=\sffamily\footnotesize] at (0,1.95) {DP rank};
  \draw[->,line width=1pt,scheduleGray] (2.98,1.95)--(3.9,1.95);
  \node[font=\sffamily\scriptsize,align=center] at (3.43,2.7)
    {sort\\+ group};
  \node[font=\sffamily\footnotesize] at (1.6,0.4) {Microbatch $\longrightarrow$};
  \node[font=\sffamily\footnotesize] at (5.3,0.4) {Microbatch $\longrightarrow$};
  \draw[scheduleGray!25] (7.25,-0.35)--(7.25,4.45);
  \node[anchor=west,font=\sffamily\bfseries] at (7.75,4.2)
    {(b) Corpus attention-work imbalance};
  \begin{axis}[
    at={(8.65cm,0.65cm)},anchor=south west,
    width=7cm,height=3.05cm,scale only axis,
    xmode=log,log basis x=2,xmin=0.9,xmax=145,
    ymin=0.8,ymax=6.2,
    xtick={1,2,4,8,16,32,64,128},
    xticklabels={1,2,4,8,16,32,64,128},
    ytick={1,2,3,4,5,6},minor tick num=0,
    xlabel={Sorting window $W$ (global microbatches)},
    ylabel={Max / mean rank cost},
    label style={font=\sffamily\footnotesize},
    tick label style={font=\sffamily\footnotesize},
    axis lines*=left,tick align=outside,
    ymajorgrids,grid style={scheduleGray!15},
  ]
    \addplot[scheduleGray,dashed,thin] coordinates {(1,1) (128,1)};
    \addplot[scheduleTeal,line width=1.2pt,mark=*,mark size=2pt]
      table[x=window,y=mean_max_over_mean,col sep=comma]
      {figures/data/corpus_window_imbalance.csv};
    \node[anchor=south west,font=\sffamily\footnotesize,text=scheduleTeal]
      at (axis cs:1,5.45) {5.38};
    \node[anchor=south,font=\sffamily\footnotesize,text=scheduleTeal]
      at (axis cs:8,1.6) {1.39};
    \node[anchor=south east,font=\sffamily\footnotesize,text=scheduleTeal]
      at (axis cs:128,1.25) {1.03};
  \end{axis}
\end{tikzpicture}%
}

%% file: figures/smoothed_loss.tex
\begin{tikzpicture}
  \begin{axis}[
    width=0.85\linewidth,height=4.1cm,
    xmin=0,xmax=267,ymin=0.6,ymax=2.5,
    xtick={0,50,100,150,200,250},ytick={0.75,1.25,1.75,2.25},
    xlabel={Optimizer step},ylabel={Training loss},
    tick label style={font=\footnotesize},label style={font=\small},
    axis lines*=left,tick align=outside,
    ymajorgrids,grid style={scheduleGray!15},
    legend style={font=\footnotesize,draw=none,fill=none,
      at={(0.98,0.98)},anchor=north east},
  ]
    \addplot[scheduleTeal,line width=1.2pt]
      table[x=step,y=loss,col sep=comma]{figures/data/enabled_loss_smoothed.csv};
    \addlegendentry{With sampling ($W=8$)}
    \addplot[scheduleGray,dashed,line width=1pt]
      table[x=step,y=loss,col sep=comma]{figures/data/disabled_loss_smoothed.csv};
    \addlegendentry{Without sampling}
  \end{axis}
\end{tikzpicture}

%% file: 7_evaluation.tex
\section{Evaluation}
\label{sec:evaluation}

We evaluate \modelname through complementary comparisons of base-model
performance, general capabilities after supervised fine-tuning (SFT),
and agentic capabilities against post-trained models. The base-model
comparison assesses the pre-trained checkpoint directly, while the
Base+SFT comparison examines general capabilities after adaptation. The agentic comparison further evaluates performance on SWE-bench
Verified, SWE-bench Pro, CyberGym, and KernelBench. These settings are presented
separately to distinguish direct base-model evaluation from downstream
performance after adaptation.

Given current resource and time constraints, we have focused our initial
evaluation and optimization on the settings reported in this section, with
particular emphasis on agentic tasks. We plan to broaden task coverage and
provide more comprehensive evaluation support in future work.

\subsection{Evaluation Settings}
\label{sec:eval-settings}

\subsubsection{Base Evaluation Settings}
\label{sec:eval-base-settings}

We evaluate base checkpoints on 13 benchmarks across five categories:
commonsense, with ARC-Easy and ARC-Challenge~\citep{clark2018arc};
reasoning, with MMLU-Pro~\citep{wang2024mmlupro},
MMLU-Redux~\citep{gema2025mmluredux}, BBH~\citep{suzgun2023bbh}, and
AGIEval-EN~\citep{zhong2024agieval};
code generation, with HumanEval~\citep{chen2021humaneval},
EvalPlus~\citep{evalplus}, and MultiPL-E~\citep{cassano2023multiple};
mathematics, with GSM-8K~\citep{cobbe2021gsm8k} and
MATH-500~\citep{lightman2024verify}; and
science, with GPQA-Main~\citep{rein2024gpqa} and
SuperGPQA~\citep{du2025supergpqa}. Table~\ref{tab:base-eval-settings}
summarizes the default prompting, scoring, and generation settings for
each benchmark. The model-specific protocol-selection procedure is
described at the end of this subsection.

\paragraph{Prompting and generation.}
All benchmarks use zero-shot prompting except BBH, which uses three
demonstrations, and AGIEval-EN, which uses five. For the non-code
benchmarks, we use greedy decoding with temperature 0 and generate one
candidate per problem. For HumanEval, EvalPlus, and
MultiPL-E, we sample four candidates per problem
with temperature 0.2 and top-$p$ 0.95. Each code-generation candidate is
limited to 2,048 generated tokens. The other generation limits range
from 1,024 to 16,384 tokens, as listed in the table.

\begin{table}[!htbp]
\centering
\small
\caption{Default base-evaluation settings (except Gemma-4-26B-A4B).
Rows follow the five benchmark categories introduced in the text.
$T$ denotes temperature, candidates are counted per problem, and
generation limits are in tokens. Greedy decoding uses $T=0$. MultiPL-E scores are weighted by the number
of problems in each language.}
\setlength{\tabcolsep}{3pt}
\renewcommand{\arraystretch}{1.12}
\begin{tabular*}{\linewidth}{@{\extracolsep{\fill}}lllrcrrrr@{}}
\toprule
Category & Benchmark & Metric & Shots & Decoding & $T$ & Top-$p$ & Candidates & Max. tokens \\
\midrule
\arrayrulecolor{black!30}
\multirow{2}{*}{Commonsense} & ARC-Easy & Exact match & 0 & \multicolumn{3}{c}{Greedy} & 1 & 16,384 \\
& ARC-Challenge & Exact match & 0 & \multicolumn{3}{c}{Greedy} & 1 & 16,384 \\
\midrule[0.2pt]
\multirow{4}{*}{Reasoning} & MMLU-Pro & Exact match & 0 & \multicolumn{3}{c}{Greedy} & 1 & 16,384 \\
& MMLU-Redux & Exact match & 0 & \multicolumn{3}{c}{Greedy} & 1 & 4,096 \\
& BBH & Exact match & 3 & \multicolumn{3}{c}{Greedy} & 1 & 1,024 \\
& AGIEval-EN & Exact match & 5 & \multicolumn{3}{c}{Greedy} & 1 & 2,048 \\
\midrule[0.2pt]
\multirow{3}{*}{Code} & HumanEval & Avg@4 & 0 & Sampling & 0.2 & 0.95 & 4 & 2,048 \\
& EvalPlus & Avg@4 & 0 & Sampling & 0.2 & 0.95 & 4 & 2,048 \\
& MultiPL-E & Avg@4 & 0 & Sampling & 0.2 & 0.95 & 4 & 2,048 \\
\midrule[0.2pt]
\multirow{2}{*}{Math} & GSM-8K & Exact match & 0 & \multicolumn{3}{c}{Greedy} & 1 & 2,048 \\
& MATH-500 & Math Verify & 0 & \multicolumn{3}{c}{Greedy} & 1 & 16,384 \\
\midrule[0.2pt]
\multirow{2}{*}{Science} & GPQA-Main & Exact match & 0 & \multicolumn{3}{c}{Greedy} & 1 & 16,384 \\
& SuperGPQA & Exact match & 0 & \multicolumn{3}{c}{Greedy} & 1 & 4,096 \\
\arrayrulecolor{black}
\bottomrule
\end{tabular*}
\label{tab:base-eval-settings}
\end{table}

\paragraph{Scoring and aggregation.}
We use exact match for the non-code benchmarks except MATH-500, whose
answers are scored with Math Verify. For HumanEval, EvalPlus, and
MultiPL-E, we report Avg@4: test-passing outcomes are averaged over the
four generated candidates for each problem and then across problems.
Our MultiPL-E evaluation covers ten languages: Python, C++, C\#, Go,
Java, JavaScript, PHP, Rust, Bash, and TypeScript. We aggregate their
scores with weights proportional to the number of evaluated problems
in each language.

\paragraph{Code execution.}
For HumanEval, EvalPlus, and MultiPL-E, we set the compilation and
execution timeouts to 15 seconds each. Runtime and compiler versions,
along with preparation, compilation, and execution commands, are
provided in Appendix~\ref{app:code-eval-environments}
(Table~\ref{tab:code-runtimes}).

\paragraph{Model-specific protocol selection for Gemma.}
For Gemma-4-26B-A4B, limited instruction-following ability makes the
results particularly sensitive to prompting and evaluation conventions.
We therefore evaluate multiple prompt and protocol configurations and
report the maximum score obtained for each benchmark.

\subsubsection{Post-training Evaluation}
\label{sec:eval-posttrain-settings}

\paragraph{General Capabilities.}
We evaluate general capabilities on MMLU~\citep{hendrycks2021mmlu},
MMLU-Pro~\citep{wang2024mmlupro}, GPQA Diamond~\citep{rein2024gpqa},
AIME 2026 and HMMT February 2026~\citep{balunovic2025matharena},
GSM8K~\citep{cobbe2021gsm8k}, MATH-500~\citep{lightman2024verify},
LiveCodeBench v5 and v6~\citep{jain2025livecodebench},
MBPP~\citep{austin2021mbpp}, HumanEval~\citep{chen2021humaneval},
IFBench~\citep{pyatkin2025ifbench}, IFEval~\citep{zhou2023ifeval},
AA-LCR v1.1~\citep{artificialanalysis2025aalcr},
SimpleQA Verified~\citep{haas2025simpleqaverified}, the public set of
AA-Omniscience~\citep{jackson2025aaomniscience},
CL-bench~\citep{dou2026clbench}, and
CL-bench Life~\citep{dou2026clbenchlife}.

We use a sampling temperature of 0.6 throughout. The maximum output
length is 262,144 tokens for AIME 2026, HMMT February 2026,
LiveCodeBench v5 and v6, CL-bench, and CL-bench Life; 131,072 tokens
for GPQA Diamond; 32,768 tokens for MMLU, MMLU-Pro, GSM8K, MATH-500,
MBPP, HumanEval, IFBench, and IFEval; and 4,096 tokens for SimpleQA
Verified and AA-Omniscience. For CL-bench and CL-bench Life, top-$p$
is set to 0.95. We sample 32 responses per problem for AIME 2026 and
HMMT February 2026; eight for LiveCodeBench v5 and v6, GPQA Diamond,
MBPP, and HumanEval; four for IFBench, IFEval, AA-LCR v1.1, CL-bench,
and CL-bench Life; and one for the remaining benchmarks.

We report accuracy for all benchmarks, averaged over the sampled
responses. Mathematical answers are graded using Math-Verify,
multiple-choice answers are compared against the reference choices,
and generated code is evaluated against the benchmark test cases.
AA-LCR v1.1 employs GPT-5.6 Luna (medium) as its judge, while SimpleQA Verified and AA-Omniscience rely on Qwen2.5-32B as the judge. For CL-bench and CL-bench
Life, we use DeepSeek-V4.1 with thinking enabled and rubric-level
majority voting: each response is judged independently three times,
and each rubric's verdict is determined by a majority vote. Only final
answer content is graded; responses containing only reasoning receive
zero credit on every rubric.

\paragraph{Agentic Capabilities.}
We evaluate agentic capabilities on SWE-bench Verified
\citep{openai2024sweverified}, SWE-bench Pro~\citep{deng2025swe},
CyberGym~\citep{wang2025cybergym}, and
KernelBench~\citep{ouyang2025kernelbench}.

For SWE-bench Verified and SWE-bench Pro, we use the SWE-agent
harness~\citep{yang2024swe}, with a sampling temperature of 0.7,
a repetition penalty of 1.1, and a maximum of 300 interaction steps.
For SWE-bench Verified, we additionally apply anti-hacking safeguards
against the shortcut behaviors summarized in Table~\ref{tab:antihack}.

For KernelBench, we use a sampling temperature of 0.6, top-$p$ of
0.95, and a repetition penalty of 1.1. Each model turn allows up to
65,536 output tokens, with a maximum of 20 interaction rounds.
Context compression is triggered when the context reaches 150K tokens.

For CyberGym, we use the Claude Code harness
\citep{anthropic2025claudecode}, with a sampling temperature of 1.0
and top-$p$ of 0.95. The context window is 262,144 tokens, with
automatic compression triggered at 90\% utilization. Each task
attempt has a two-hour time limit.

\subsection{Base Model Evaluation}
\label{sec:eval-base}

\begin{table*}[t]
\centering
\caption{Base-model evaluation. All entries are base checkpoints.
Scores are percentages; AVG is the unweighted mean of the 13 benchmark
scores. Bold indicates the highest score among the models shown.
$^{*}$Gemma-4-26B-A4B reports the maximum score across multiple prompts
and evaluation protocols for each benchmark, as described in
Section~\ref{sec:eval-base-settings}.}
\label{tab:eval-base}
\begingroup
\small
\setlength{\tabcolsep}{2pt}
\renewcommand{\arraystretch}{1.12}
\begin{tabular*}{\linewidth}{@{\extracolsep{\fill}}llccccccc@{}}
\toprule
Category & Benchmark &
\shortstack{\modelname\\Base} &
\shortstack{Qwen3.5\\35B-A3B\\Base} &
\shortstack{Qwen3\\30B-A3B\\Base} &
\shortstack{Kimi-Linear\\48B-A3B\\Base} &
\shortstack{Nemotron\\3-Nano\\30B-A3B\\Base} &
\shortstack{Nemotron\\3.5-Lightning\\30B-A3B\\Base} &
\shortstack{Gemma-4\\26B-A4B$^{*}$} \\
\midrule
\arrayrulecolor{black!30}
\multirow{2}{*}{Commonsense} & ARC-Easy & 97.69 & \textbf{98.27} & 90.40 & 93.43 & 93.86 & 98.06 & 96.46 \\
& ARC-Challenge & 92.75 & \textbf{97.01} & 83.70 & 85.75 & 84.98 & 94.54 & 89.93 \\
\midrule[0.2pt]
\multirow{4}{*}{Reasoning} & MMLU-Pro & 68.38 & 67.01 & 45.15 & 49.56 & 59.26 & \textbf{69.13} & 43.55 \\
& MMLU-Redux & 83.14 & \textbf{86.14} & 62.26 & 73.44 & 76.95 & 83.74 & 73.30 \\
& BBH & 81.23 & \textbf{86.35} & 80.33 & 76.07 & 78.50 & 77.59 & 72.62 \\
& AGIEval-EN & 58.98 & \textbf{66.56} & 62.59 & 60.49 & 58.59 & 58.93 & 49.30 \\
\midrule[0.2pt]
\multirow{3}{*}{Code} & HumanEval & \textbf{78.81} & 71.80 & 71.95 & 68.14 & 64.79 & 69.97 & 63.11 \\
& EvalPlus & 75.46 & \textbf{83.08} & 79.42 & 75.15 & 73.02 & 80.79 & 53.20 \\
& MultiPL-E & 57.74 & \textbf{66.48} & 61.04 & 59.63 & 56.04 & 62.16 & 18.18 \\
\midrule[0.2pt]
\multirow{2}{*}{Math} & GSM-8K & \textbf{92.80} & 87.72 & 82.79 & 88.86 & 90.14 & 91.36 & 68.46 \\
& MATH-500 & 73.20 & \textbf{83.00} & 76.60 & 62.40 & 76.00 & 74.00 & 36.20 \\
\midrule[0.2pt]
\multirow{2}{*}{Science} & GPQA-Main & 51.34 & \textbf{55.58} & 33.26 & 25.22 & 42.19 & 51.12 & 35.27 \\
& SuperGPQA & 39.16 & \textbf{47.68} & 30.60 & 18.63 & 36.06 & 38.57 & 24.90 \\
\arrayrulecolor{black}
\midrule
\multicolumn{2}{@{}l}{AVG} & \silvermedal 73.13 & \goldmedal 76.67 & 66.16 & 64.37 & 68.49 & \bronzemedal 73.07 & 55.73 \\
\bottomrule
\end{tabular*}
\endgroup
\end{table*}

We first compare \modelname with models for which base checkpoints
are available: Qwen3.5-35B-A3B~\citep{qwen2026qwen35},
Qwen3-30B-A3B~\citep{yang2025qwen3},
Kimi-Linear-48B-A3B~\citep{kimi2025kimilinear},
Nemotron-3-Nano-30B-A3B~\citep{nvidia2025nemotron3nano},
Nemotron-3.5-Lightning-30B-A3B~\citep{nvidia2026nemotron35lightning}, and Gemma-4-26B-A4B~\citep{google2026gemma4}.
Table~\ref{tab:eval-base} reports results on the 13 benchmarks spanning
commonsense, reasoning, code generation, mathematics, and science,
together with their average score.

\paragraph{Competitive base-model performance.}
\modelname achieves an average score of 73.13, the second-highest
among the seven base models, behind Qwen3.5 at 76.67 and close to
Nemotron-3.5-Lightning at 73.07. It exceeds Qwen3, Nemotron-3-Nano,
and Kimi-Linear by 6.97, 4.64, and 8.76 percentage points,
respectively. This result establishes a competitive
starting point before the SFT evaluation considered below. \modelname obtains the highest scores on GSM-8K and HumanEval,
reaching 92.80\% and 78.81\%, respectively. On MMLU-Pro, it scores 68.38\%, compared with 67.01\% for
Qwen3.5 and 69.13\% for Nemotron-3.5-Lightning. The base-model
results demonstrate a competitive overall profile and distinct
task-level strengths of our model.

\subsection{SFT Model Results}
\label{sec:eval-sft}

\subsubsection{General Capabilities}

\begin{table*}[!htb]
\centering
\caption{General capabilities of Base+SFT models(\%). All base models are
fine-tuned on the same 6B-token SFT data without reinforcement learning.
Scores are percentages; higher is better. AVG5 and AVG13 are the
unweighted means of the upper and lower blocks, respectively.
Bold indicates the highest score among the models shown, including ties.}
\label{tab:eval-general}
\begingroup
\small
\setlength{\tabcolsep}{4pt}
\renewcommand{\arraystretch}{1.1}
\begin{tabular*}{\linewidth}{@{\extracolsep{\fill}}lcccccc@{}}
\toprule
Benchmark &
\shortstack{\modelname\\Base+SFT} &
\shortstack{Qwen3.5\\35B-A3B\\Base+SFT} &
\shortstack{Nemotron\\3-Nano\\30B-A3B\\Base+SFT} &
\shortstack{Qwen3\\30B-A3B\\Base+SFT} &
\shortstack{Nemotron\\3.5-Lightning\\30B-A3B\\Base+SFT} &
\shortstack{Gemma-4\\26B-A4B\\Base+SFT} \\
\midrule
\multicolumn{7}{l}{\textit{Specialized-capability Evaluations}} \\
AA-LCR v1.1 & \textbf{61.00} & 52.00 & 32.00 & 19.75 & 18.50 & 18.25 \\
AA-OmniScience (public set) & \textbf{31.30} & 21.70 & 16.70 & 18.80 & 17.80 & 19.00 \\
SimpleQA-verified & \textbf{34.50} & 12.00 & 7.00 & 9.10 & 9.90 & 10.20 \\
CL-bench & \textbf{13.22} & 12.80 & 7.74 & 6.95 & 5.00 & 4.42 \\
CLbench-life & 3.46 & \textbf{4.20} & 3.95 & 2.22 & 1.48 & 1.73 \\
\midrule
AVG5 & \goldmedal \textbf{28.70}  & \silvermedal 20.54  & \bronzemedal 13.48 & 11.36 & 10.54 & 10.72 \\
\midrule
\multicolumn{7}{l}{\textit{Classic General Benchmarks}} \\
AIME 2026 & 86.77 & \textbf{87.60} & 78.65 & 82.29 & 81.25 & 45.73 \\
HMMT Feb 2026 & 68.37 & \textbf{71.02} & 60.04 & 65.62 & 64.02 & 31.06 \\
LiveCodeBench v5 & 69.53 & \textbf{77.37} & 72.00 & 65.73 & 70.83 & 48.07 \\
LiveCodeBench v6 & 66.41 & \textbf{71.95} & 68.19 & 62.02 & 64.22 & 41.41 \\
GPQA Diamond & 74.75 & \textbf{75.76} & 66.54 & 65.97 & 70.45 & 53.28 \\
MMLU-Pro & 80.00 & \textbf{82.13} & 76.58 & 77.40 & 79.22 & 72.62 \\
IFBench & 63.33 & \textbf{64.92} & 59.42 & 58.08 & 58.17 & 45.17 \\
IFEval Strict & 85.54 & \textbf{89.33} & 87.34 & 86.74 & 86.69 & 68.95 \\
MMLU & 84.62 & \textbf{88.75} & 83.34 & 84.01 & 86.60 & 82.87 \\
GSM8K & 95.75 & \textbf{95.98} & 94.62 & 95.60 & \textbf{95.98} & 93.78 \\
Math500 & \textbf{98.40} & \textbf{98.40} & 96.80 & 96.40 & 97.20 & 84.80 \\
MBPP & 88.83 & \textbf{90.30} & 87.44 & 86.25 & 89.75 & 72.55 \\
HumanEval & 95.12 & \textbf{95.35} & 93.67 & 91.39 & 94.05 & 84.98 \\
\midrule
AVG13 & \silvermedal 81.34 & \goldmedal83.76 & 78.82 & 78.27 & \bronzemedal79.88 & 63.48 \\
\bottomrule
\end{tabular*}%
\endgroup
\end{table*}

We then examine whether the strength of the base model carries over after
adaptation. Every model here is fine-tuned on the same SFT data, which is
deliberately lightweight: 6.33 B tokens and no reinforcement learning
(Section~\ref{sec:measure-methods}). Differences among the resulting models
therefore mainly reflect what each base model brings to adaptation.

Table~\ref{tab:eval-general} reports two groups of benchmarks.
The lower block contains the \emph{classic general benchmarks},
a suite of 13 widely used benchmarks covering the general
capabilities commonly evaluated in models of this class,
summarized by AVG13. The upper block contains five
\emph{specialized-capability evaluations}, covering three
specific areas: long-context reasoning (AA-LCR v1.1),
factual knowledge (AA-OmniScience on its public set and
SimpleQA-verified), and learning from context
(CL-bench and CL-bench Life). Their mean is reported as AVG5.

\paragraph{Performance on specialized-capability evaluations.}
\modelname achieves an AVG5 of 28.70, leading Qwen3.5
(20.54) by 8.16 points and ranking first on four of the
five evaluations. Its strongest results are on long-context
reasoning and factual knowledge, particularly AA-LCR v1.1
(61.00\%), AA-OmniScience (31.30\%), and SimpleQA-verified
(34.50\%). The pre-training corpus was decontaminated against
SimpleQA using 8-gram overlap matching, making test-set
leakage an unlikely explanation for this result. The exception
is CL-bench Life, where \modelname trails Qwen3.5 and
Nemotron-3-Nano.

\paragraph{Performance on classic general benchmarks.}
\modelname achieves an AVG13 of 81.34, ranking second
behind Qwen3.5-35B-A3B (83.76), consistent with its
ranking in base-model evaluation (Table~\ref{tab:eval-base}).
It performs particularly well on mathematics and code,
matching the highest Math500 score (98.40\%) and reaching
95.75\% on GSM8K and 95.12\% on HumanEval. Its main
weaknesses are in competitive programming and instruction
following, as reflected by its results on LiveCodeBench
v5 and IFEval Strict.

\subsubsection{Agentic Capabilities}
\label{sec:eval-agentic}
We further compare \modelname, after the same lightweight SFT, with
post-trained models released by their providers, which have generally
gone through a full post-training process, on SWE-bench Verified,
SWE-bench Pro, CyberGym, and KernelBench. In addition to general-purpose
models of a similar size, namely
Qwen3-30B-A3B-Thinking-2507~\citep{qwen2025qwen3thinking2507},
Qwen3.5-35B-A3B~\citep{qwen2026qwen35},
Qwen3.6-35B-A3B~\citep{qwen2026qwen36},
K2-Horizon-MoVA-36B-A4B~\citep{ifm2026k2horizon},
Gemma-4-26B-A4B-it~\citep{google2026gemma4}, and
Nemotron-3.5-Lightning-30B-A3B~\citep{nvidia2026nemotron35lightning},
the comparison includes Flint-40B-A2.5B~\citep{recursive2026flint}, a
coding-specialized MoE model.
Table~\ref{tab:eval-agentic} summarizes these results. Unlike the
Base+SFT comparison above, this setting compares our model after
6.33B tokens of tiny SFT with released models that have gone through
complete post-training pipelines, so we report it as a reference point
for post-training headroom rather than as a like-for-like comparison.

\begin{table*}[!htbp]
\centering
\caption{Agentic benchmark results (\%). Bold indicates the highest
score among the models shown; AVG is the unweighted mean of the four
benchmark scores. All baselines are provider-released post-trained
models; \modelname is evaluated after the lightweight SFT stage only.
}
\label{tab:eval-agentic}
\begingroup
\small
\setlength{\tabcolsep}{5pt}
\renewcommand{\arraystretch}{1.1}
\begin{tabular*}{\linewidth}{@{\extracolsep{\fill}}lccccc@{}}
\toprule
Model &
\shortstack{SWE-bench\\Verified} &
\shortstack{SWE-bench\\Pro} &
CyberGym & KernelBench & AVG \\
\midrule
Qwen3-30B-A3B-Thinking-2507 & 20.30 & 7.01 & 1.19 & 5.60 & 8.53 \\
Qwen3.5-35B-A3B & 69.10 & 39.23 & 13.47 & 17.20 & 34.75 \\
Qwen3.6-35B-A3B & 71.65 & 45.28 & 24.22 & 37.20 & 44.59 \\
K2-Horizon-MoVA-36B-A4B & 64.15 & 44.19 & \textbf{39.35} & 33.60 & 45.32 \\
Gemma-4-26B-A4B-it & 57.15 & 34.57 & 8.49 & 44.00 & 36.05 \\
Nemotron-3.5-Lightning-30B-A3B & 55.35 & 32.49 & 8.29 & 20.40 & 29.13 \\
Flint-40B-A2.5B & \textbf{72.80} & \textbf{53.01} & 33.91 & 41.20 & 50.23 \\
\midrule
\modelname Base+SFT& 68.95 & 48.29 & 38.90 & \textbf{77.60} & \textbf{58.43} \\
\bottomrule
\end{tabular*}%
\endgroup
\end{table*}

\modelname achieves the highest score on KernelBench
(77.60\%) and the highest average across the four
benchmarks (58.43). It also achieves competitive results
on SWE-bench Pro and CyberGym. Notably, \modelname is
evaluated after only a lightweight SFT stage using
6.33B tokens, without reinforcement learning, whereas
the compared models have generally undergone much more
extensive post-training. These results demonstrate
the strong agentic capabilities that \modelname
already exhibits after lightweight post-training.

\subsection{Learning aptitude}
\label{sec:eval-aptitude}

\begin{figure}[!htb]
\centering
\includegraphics[width=\linewidth]{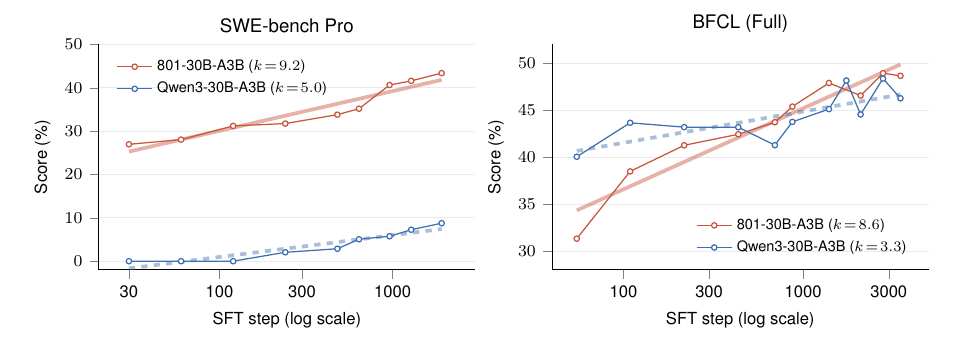}
\caption{Learning aptitude of \modelname and Qwen3-30B-A3B on SWE-bench Pro
(left) and BFCL Full (right). Hollow markers are scores at successive SFT
checkpoints; the pale line is the least-squares fit of
Eq.~\eqref{eq:learnability}, and $k$ in the legend is its slope, the gain
in points per tenfold increase in SFT steps.}
\label{fig:learnability-real}
\end{figure}

\paragraph{Lower start, faster learning on BFCL}
At the first checkpoint (step 55), \modelname scores 31.32 on BFCL Full,
below the 40.04 of Qwen3-30B-A3B. Its slope, however, is 8.62 points per
tenfold increase in SFT steps against 3.33, about $2.6\times$ steeper. From
step 697 onward it leads at every shared checkpoint and ends at 48.65
against 46.26 at step 3{,}485. This is the case that motivates
learning aptitude: a checkpoint that starts lower but learns faster overtakes
one that starts higher once fine-tuning runs long enough.

\paragraph{Higher start and faster learning on SWE-bench Pro}
On SWE-bench Pro, \modelname leads from the first checkpoint, 26.95
against 0 at step 30, and learns faster, with a slope of 9.15 against
5.03. Qwen3-30B-A3B stays at zero for the first three checkpoints (up to
step 120) and reaches 8.76 at step 1{,}922, where \modelname reaches 43.37. Here the intercept and the slope point the same way, so the gap widens rather than closes as fine-tuning continues.

%% file: 8_conclusion.tex
\section{Conclusion}

We attempted to answer a question that current RSI successes leave open: what
can agents do for frontier pre-training, where the
$\textit{propose}\rightarrow\textit{run}\rightarrow\textit{measure}
\rightarrow\textit{update}$ loop cannot close because the run is not
repeatable and the objective is not a score? Our answer, drawn from building
\modelname{} in three months, is to organize improvement around three types
of loops. Debugging loops comprise
the infra debugging loop and the data debugging loop, which connect unusual
internal metrics and output patterns to their causes. Five cross-stack
diagnoses, resolved in about one month, came out of these investigations.
The headroom loop is instantiated as the recipe loop,
using tiny-SFT, learning aptitude, and max-over-protocols evaluation to guide
recipe revisions and monitor training progress. Experience-to-metric loops
comprise the classifier development loop, the site discovery loop, and the
crawl-coverage loop, turning experience-based judgment into feedback
that agents can use to improve data workflows. All of them share a context
infrastructure that integrates evidence across teams at scale and preserves
negative results and their reasons. 

The resulting model is evidence that this regime works at scale. \modelname{}
is state-of-the-art among open base models at the 30B MoE scale in both direct
base evaluation and matched-SFT comparison, and after a 6B-token SFT stage it
approaches a fully post-trained frontier of its size class on agentic tasks.
But we hold the model to be the byproduct and the account to be the
contribution. Our experience suggests that RSI need not stop at frontier pre-training.
It takes a different form: loops eliminate anomalies, headroom
measurements guide training, and quantitative feedback improves data
workflows. Shared context allows evidence and lessons to accumulate
across teams and iterations. There may be no second run, but the one
we have can still learn from its own progress.

%% file: 9_contributors.tex
\newpage
\phantomsection
\addcontentsline{toc}{section}{Author Contributions}
\noindent{\large\textsc{Author Contributions}}

\vspace{0.8em}
\noindent \textbf{The authors are listed in alphabetical order by their last names.}

\vspace{1.5em}
\noindent\textbf{Contributors:}\quad
Xiaojie Cai, Jiayi Chen, Junwei Chen, Yu Chen, Sikai Cheng, Jinquan Dai, Siyuan Feng, Bin Fu, Dayuan Fu, Guosheng Gao, Qipeng Guo, Chenwei He, Zhulin Hu, Yaxing Huang, Shuomeng Jia, Mohan Jiang, Keyang Lei, Hongliang Li, Xuefeng Li, Zhaocheng Li, Jifan Lin, Pengfei Liu, Xinghan Liu, Yixiu Liu, Pengrui Lu, Siyuan Lu, Kai Lv, Tiantian Mi, Yang Nan, Yu Qiao, Yiwei Qin, Xipeng Qiu, Junhao Shi, Runzhou Shi, Jiadi Su, Bohao Tang, Hanning Wang, Keqi Wang, Pan Wang, Ran Wang, Shaojing Wang, Yishuo Wang, Zhixin Wang, Shenyu Wu, Muhang Xie, Tianze Xu, Hang Yan, Lvmanshan Ye, Hongche Yu, Junjie Yu, Lantian Zhang, Ruihang Zhang, Bin Zhao, Chenyang Zhou, Zijin Zhou, Wenhong Zhu, Zemu Zhu, Yumin Zhuang, Haoyang Zou

\vspace{1.5em}
\noindent\textbf{Affiliations:}\quad
Shanghai Innovation Institute, Xinxu AI

%% file: 10_appendix.tex
\appendix

\section{Layer-wise MaxVio Monitoring}
\label{app:layerwise-maxvio}
\begin{figure*}[h]
\centering
\includegraphics[width=0.9\textwidth]{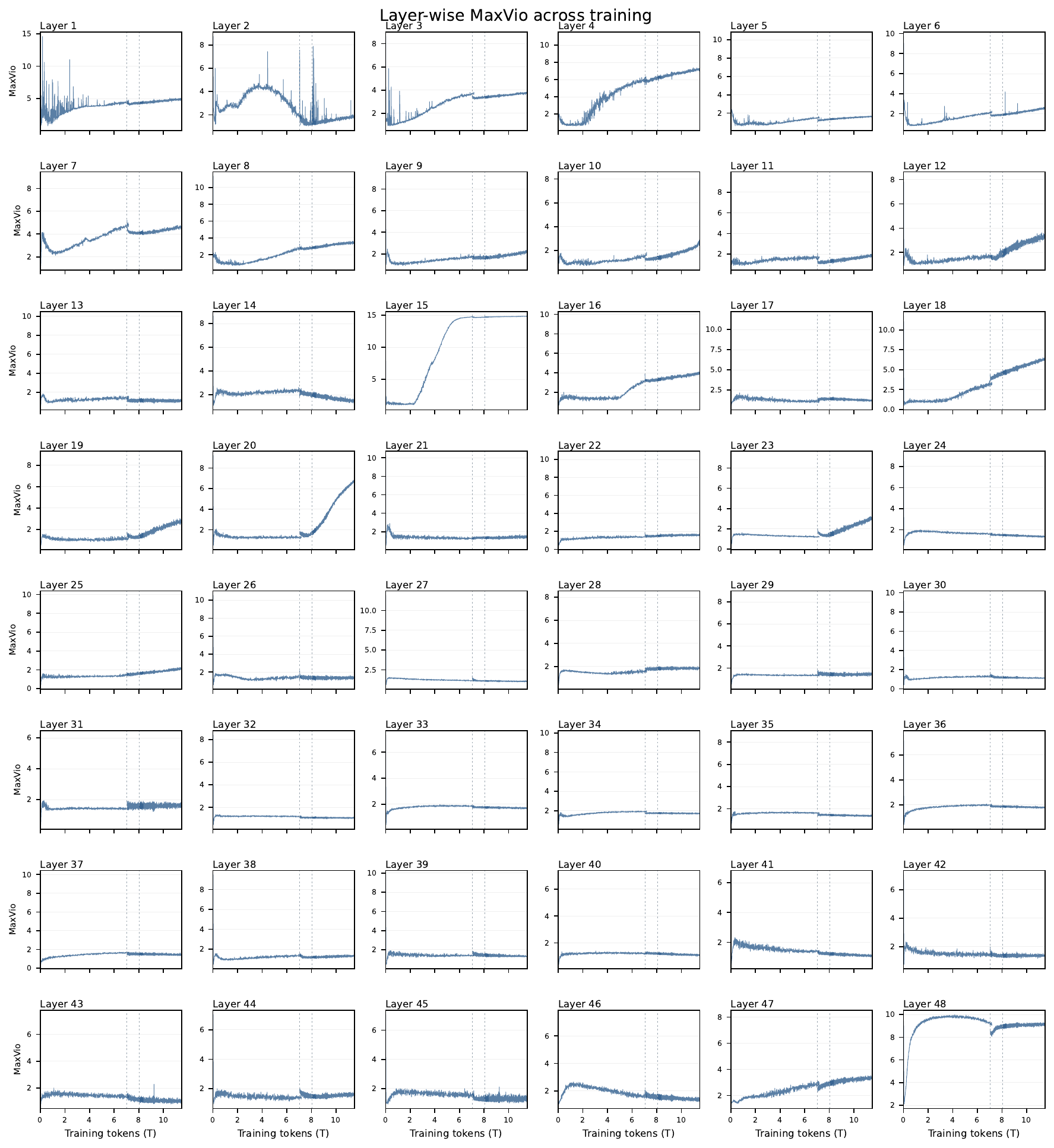}
\caption{Layer-wise MaxVio trajectories for all 48 MoE layers.}
\label{fig:layerwise-maxvio}
\end{figure*}

Figure~\ref{fig:layerwise-maxvio} reports the MaxVio trajectories for all 48
Mixture-of-Experts layers over training. Most layers remain within a relatively
stable range, while a small number develop substantially larger violations; in
particular, layer 15 exhibits the pronounced rise discussed in
Figure~\ref{fig:maxvio-saturation}. The vertical dashed lines indicate the
stage boundaries, and the horizontal axis is training tokens in trillions.

\clearpage
\section{Benchmarks for Pre-training Monitoring}
\label{app:pretraining-benchmarks}

Table~\ref{tab:pretraining-benchmarks} summarizes the 22 evaluation items
used to monitor pre-training, grouped as in the training-progress plots.
Overall\_Avg is the unweighted mean of their 22 reported percentage scores.

\begin{table}[!htbp]
\centering
\small
\caption{The 22-benchmark pre-training monitoring suite, with brief
introductions to the evaluated tasks.}
\label{tab:pretraining-benchmarks}
\setlength{\tabcolsep}{6pt}
\renewcommand{\arraystretch}{1.12}
\begin{tabularx}{\linewidth}{@{}>{\raggedright\arraybackslash}p{0.38\linewidth}>{\raggedright\arraybackslash}X@{}}
\toprule
Benchmark & Description \\
\midrule
\multicolumn{2}{@{}l}{\textit{General commonsense (7)}} \\
ARC-Easy~\citep{clark2018arc} & School-level science questions, easy split. \\
ARC-Challenge~\citep{clark2018arc} & More challenging science questions requiring reasoning. \\
HellaSwag~\citep{zellers2019hellaswag} & Selecting plausible continuations of everyday events. \\
OpenBookQA~\citep{mihaylov2018openbookqa} & Combining basic science facts with commonsense. \\
PIQA~\citep{bisk2019piqa} & Physical commonsense and practical problem solving. \\
WinoGrande~\citep{sakaguchi2019winogrande} & Resolving ambiguous references using commonsense. \\
TriviaQA~\citep{joshi2017triviaqa} & Factual short-answer questions without evidence context. \\
\arrayrulecolor{black!30}
\midrule[0.2pt]
\multicolumn{2}{@{}l}{\textit{General reasoning (6)}} \\
MMLU~\citep{hendrycks2021mmlu} & Multiple-choice knowledge questions across 57 subjects. \\
MMLU-Pro~\citep{wang2024mmlupro} & Harder, reasoning-oriented questions across 14 subjects. \\
MMLU-Redux~\citep{gema2025mmluredux} & Human-reviewed MMLU questions. \\
BBH~\citep{suzgun2023bbh} & Diverse challenging reasoning tasks; 27 configured subtasks. \\
AGIEval-EN~\citep{zhong2024agieval} & Ten English exam subsets covering logic, reading, and math. \\
DROP~\citep{dua2019drop} & Reading comprehension with discrete reasoning. \\
\midrule[0.2pt]
\multicolumn{2}{@{}l}{\textit{Code (3)}} \\
HumanEval~\citep{chen2021humaneval} & Implementing Python functions from signatures and natural-language specifications to assess programming and problem-solving ability. \\
EvalPlus~\citep{evalplus} & Code-generation evaluation with augmented tests, emphasizing functional correctness and edge-case handling. \\
MBPP~\citep{austin2021mbpp} & Writing short Python programs from natural-language requirements. \\
\midrule[0.2pt]
\multicolumn{2}{@{}l}{\textit{Math (2)}} \\
GSM-8K~\citep{cobbe2021gsm8k} & Grade-school word problems requiring multi-step arithmetic. \\
MATH-500~\citep{lightman2024verify} & Competition-style mathematics. \\
\midrule[0.2pt]
\multicolumn{2}{@{}l}{\textit{Science (4)}} \\
GPQA-Main~\citep{rein2024gpqa} & Graduate-level biology, physics, and chemistry questions. \\
SuperGPQA~\citep{du2025supergpqa} & Graduate-level knowledge and reasoning across diverse disciplines, including science, engineering, medicine, law, economics, and the humanities. \\
MMLU-STEM & The 19 MMLU subjects: abstract algebra, anatomy, astronomy, college biology, college chemistry, college computer science, college mathematics, college physics, computer security, conceptual physics, electrical engineering, elementary mathematics, high school biology, high school chemistry, high school computer science, high school mathematics, high school physics, high school statistics, and machine learning. \\
MMLU-Pro-STEM & Six MMLU-Pro domains: biology, chemistry, computer science, engineering, mathematics, and physics. \\
\arrayrulecolor{black}
\bottomrule
\end{tabularx}
\end{table}

\clearpage
\section{Evaluation Protocols for Base Models}
\label{app:base-eval-protocols}

Figure~\ref{fig:base-evaluation-protocols} evaluates the same checkpoints
under several protocols. A protocol is determined by four choices:
perplexity-based (PPL) or generation-based scoring, the number of in-context
examples (\textit{k}-shot), the required final-answer format, and the
answer-matching method. Table~\ref{tab:base-eval-protocols} defines each
protocol label used in the figure.

\begin{table}[!htbp]
\centering
\small
\caption{Evaluation protocols in
Figure~\ref{fig:base-evaluation-protocols}.}
\label{tab:base-eval-protocols}
\setlength{\tabcolsep}{4pt}
\renewcommand{\arraystretch}{1.12}
\begin{tabularx}{\linewidth}{@{}>{\raggedright\arraybackslash}p{0.2\linewidth}>{\raggedright\arraybackslash}X>{\raggedright\arraybackslash}p{0.22\linewidth}>{\raggedright\arraybackslash}p{0.2\linewidth}@{}}
\toprule
Protocol & In-context examples & Final-answer format & Scoring \\
\midrule
\multicolumn{4}{@{}l}{\textit{Non-code tasks}} \\
5-shot & 5 solved examples: final answers only for ARC, GPQA-Main,
MMLU-Redux, and AGIEval-EN; worked solutions for MMLU-Pro and SuperGPQA
& ``The answer is (X).'', shown by the examples & Extracted letter, exact match \\
4-shot Minerva & 4 worked solutions in Minerva format (MATH-500)
& Follows the examples & Math equivalence \\
8-shot CoT & 8 worked solutions (GSM-8K) & ``The answer is N.''
& Strict match, exact match \\
3-shot CoT & 3 official worked solutions per subtask (BBH)
& ``the answer is \ldots'' & Extracted answer, exact match \\
0-shot answer-is & None & Instructed to end with ``the answer is X''
& Extracted answer, exact match$^{\dagger}$ \\
0-shot boxed & None & Instructed to put the answer in
\texttt{\textbackslash boxed\{\}} & Extracted answer, exact match$^{\dagger}$ \\
PPL acc\_norm & None & No generation & Highest length-normalized
option log-likelihood \\
\arrayrulecolor{black!30}
\midrule[0.2pt]
\multicolumn{4}{@{}l}{\textit{Code tasks}} \\
3-shot completion & 3 & Continue the bare function signature and
docstring & Unit tests, pass@1 \\
\textit{k}-shot instruction + signature & 3 or 0 & Instruction prompt;
signature pre-filled in the answer & Unit tests, pass@1 \\
\textit{k}-shot instruction, no signature & 3 or 0 & Instruction prompt;
the model writes the full function & Unit tests, pass@1 \\
\arrayrulecolor{black}
\bottomrule
\end{tabularx}

\vspace{2pt}
\begin{minipage}{\linewidth}
\footnotesize $^{\dagger}$Math equivalence for MATH-500.
\end{minipage}
\end{table}

\clearpage
\section{Code Evaluation Environments}
\label{app:code-eval-environments}

Table~\ref{tab:code-runtimes} provides the code execution environments for
the base-model evaluations in Section~\ref{sec:eval-base-settings}.
HumanEval and EvalPlus use the Python environment, while MultiPL-E uses
all ten language environments.

\begin{table}[!htbp]
\centering
\small
\caption{Code execution environments. Preparation or compilation commands
precede execution.
Java uses JVM build \texttt{25.0.1-internal-adhoc.conda.src}.
\texttt{JAVATUPLES\_JAR} and \texttt{TSC\_JS} identify the JavaTuples
dependency and the TypeScript compiler entry point, respectively.}
\setlength{\tabcolsep}{6pt}
\renewcommand{\arraystretch}{1.1}
\begin{tabularx}{\linewidth}{@{}>{\raggedright\arraybackslash}p{0.26\linewidth}>{\raggedright\arraybackslash}X@{}}
\toprule
Language / runtime & Preparation and execution commands \\
\midrule
\arrayrulecolor{black!30}
\textbf{Python}\newline CPython 3.10.19 &
\texttt{exec(check\_program, exec\_globals)} \\
\midrule[0.2pt]
\textbf{C++}\newline GCC 13.3.0\newline libstdc++ 15.2.0 &
\texttt{g++ -std=c++17 program.cpp -o program}\newline
\texttt{./program} \\
\midrule[0.2pt]
\textbf{C\#}\newline Mono JIT / mcs 6.12.0.199 &
\texttt{mcs -define:DEBUG -r:System.Numerics.dll \textbackslash}\newline
\hspace*{1em}\texttt{-out:Program.exe Program.cs}\newline
\texttt{MONO\_TRACE\_LISTENER=Console.Error mono Program.exe} \\
\midrule[0.2pt]
\textbf{Go}\newline Go 1.23.4 (linux/amd64) &
\texttt{go mod init solution}\newline
\texttt{go test -p=8} \\
\midrule[0.2pt]
\textbf{Java}\newline OpenJDK / javac\newline 25.0.1-internal &
\texttt{javac -cp ".:\$JAVATUPLES\_JAR" Problem.java}\newline
\texttt{java -ea -cp ".:\$JAVATUPLES\_JAR" Problem} \\
\midrule[0.2pt]
\textbf{JavaScript}\newline Node.js 20.12.2 &
\texttt{node program.js} \\
\midrule[0.2pt]
\textbf{PHP}\newline PHP CLI 8.5.4\newline Zend Engine 4.5.4 &
\texttt{php program.php} \\
\midrule[0.2pt]
\textbf{Rust}\newline rustc 1.96.0; LLVM 22.1.2 &
\texttt{rustc --edition 2021 program.rs -o program}\newline
\texttt{./program} \\
\midrule[0.2pt]
\textbf{Bash}\newline GNU Bash 5.1.16 &
\texttt{bash program.sh} \\
\midrule[0.2pt]
\textbf{TypeScript}\newline TypeScript 5.4.5\newline Node.js 20.12.2 &
\texttt{node "\$TSC\_JS" program.ts --target es2020 \textbackslash}\newline
\hspace*{1em}\texttt{--module commonjs --lib es2020,dom --skipLibCheck}\newline
\texttt{node program.js} \\
\arrayrulecolor{black}
\bottomrule
\end{tabularx}
\label{tab:code-runtimes}
\end{table}

%% file: main.bib
@article{fu2026davinci,
  title={daVinci-kernel: Co-Evolving Skill Selection, Summarization, and Utilization via {RL} for {GPU} Kernel Optimization},
  author={Fu, Dayuan and Jiang, Mohan and Wang, Tongyu and Yang, Dian and Hu, Jiarui and Liu, Liming and Hou, Jinlong and Liu, Pengfei},
  journal={arXiv preprint arXiv:2606.16497},
  year={2026}
}

@misc{ahmad2025opencodereasoningiisimpletesttime,
      title={OpenCodeReasoning-II: A Simple Test Time Scaling Approach via Self-Critique}, 
      author={Wasi Uddin Ahmad and Somshubra Majumdar and others},
      year={2025},
      eprint={2507.09075},
      archivePrefix={arXiv},
      primaryClass={cs.CL},
      url={https://arxiv.org/abs/2507.09075},
}

@misc{xu2026logicsstemempoweringllmreasoning,
      title={Logics-STEM: Empowering {LLM} Reasoning via Failure-Driven Post-Training and Document Knowledge Enhancement}, 
      author={Mingyu Xu and Cheng Fang and Keyue Jiang and Yuqian Zheng and Yanghua Xiao and Baojian Zhou and Qifang Zhao and Suhang Zheng and Xiuwen Zhu and Jiyang Tang and Yongchi Zhao and Yijia Luo and Zhiqi Bai and Yuchi Xu and Wenbo Su and Wei Wang and Bing Zhao and Lin Qu and Xiaoxiao Xu},
      year={2026},
      eprint={2601.01562},
      archivePrefix={arXiv},
      primaryClass={cs.AI},
      url={https://arxiv.org/abs/2601.01562}, 
}

@article{zeng2026davinci,
  title={daVinci-Dev: Agent-native Mid-training for Software Engineering},
  author={Zeng, Ji and Fu, Dayuan and Mi, Tiantian and Zhuang, Yumin and Huang, Yaxing and Li, Xuefeng and Ye, Lyumanshan and Xie, Muhang and Hua, Qishuo and Huang, Zhen and others},
  journal={arXiv preprint arXiv:2601.18418},
  year={2026}
}

@misc{anthropic2025claudecode,
  title        = {{Claude Code}},
  author       = {{Anthropic}},
  year         = {2025},
  howpublished = {\url{https://github.com/anthropics/claude-code}},
  note         = {Accessed 2026-10-14},
  url          = {https://github.com/anthropics/claude-code}
}

@misc{anthropic_messages_api,
  title        = {Messages {API} Reference},
  author       = {{Anthropic}},
  year         = {2026},
  howpublished = {\url{https://docs.claude.com/en/api/messages}},
  note         = {Accessed 2026-10-14},
  url          = {https://docs.claude.com/en/api/messages}
}

@misc{artificialanalysis2025aalcr,
  title        = {Artificial Analysis Long Context Reasoning Benchmark ({AA-LCR})},
  author       = {{Artificial Analysis Team}},
  year         = {2025},
  publisher    = {Artificial Analysis, Inc.},
  howpublished = {Hugging Face dataset \url{https://huggingface.co/datasets/ArtificialAnalysis/AA-LCR}},
  note         = {Version 1.1 released September 2026; methodology at \url{https://artificialanalysis.ai/methodology/intelligence-benchmarking}},
  url          = {https://huggingface.co/datasets/ArtificialAnalysis/AA-LCR}
}

@misc{austin2021mbpp,
  title         = {Program Synthesis with Large Language Models},
  author        = {Austin, Jacob and Odena, Augustus and Nye, Maxwell and Bosma, Maarten and Michalewski, Henryk and Dohan, David and Jiang, Ellen and Cai, Carrie and Terry, Michael and Le, Quoc and others},
  year          = {2021},
  eprint        = {2108.07732},
  archivePrefix = {arXiv},
  primaryClass  = {cs.PL},
  url           = {https://arxiv.org/abs/2108.07732}
}

@inproceedings{badertdinov2025swe,
  title={{SWE-rebench}: An Automated Pipeline for Task Collection and Decontaminated Evaluation of Software Engineering Agents},
  author={Badertdinov, Ibragim and Golubev, Alexander and Nekrashevich, Maksim and Shevtsov, Anton and Karasik, Simon and Andriushchenko, Andrei and Trofimova, Maria and Litvintseva, Daria and Yangel, Boris},
  booktitle={Advances in Neural Information Processing Systems (NeurIPS), Datasets and Benchmarks Track},
  year={2025},
  url={https://arxiv.org/abs/2505.20411},
  note={arXiv:2505.20411}
}

@inproceedings{balunovic2025matharena,
  title     = {{MathArena}: Evaluating {LLMs} on Uncontaminated Math Competitions},
  author    = {Balunovi{\'c}, Mislav and Dekoninck, Jasper and Petrov, Ivo and Jovanovi{\'c}, Nikola and Vechev, Martin},
  booktitle = {Advances in Neural Information Processing Systems (NeurIPS), Datasets and Benchmarks Track},
  year      = {2025},
  eprint    = {2505.23281},
  archivePrefix = {arXiv},
  url       = {https://arxiv.org/abs/2505.23281}
}

@article{besiroglu2024chinchilla,
  title={Chinchilla Scaling: A replication attempt},
  author={Besiroglu, Tamay and Erdil, Ege and Barnett, Matthew and You, Josh},
  journal={arXiv preprint arXiv:2404.10102},
  year={2024},
  url={https://arxiv.org/abs/2404.10102}
}

@misc{cao2026neohorse,
  title         = {{NeoHorse-1}: Towards Recursive Self-Improvement via Agentic Post-Training with Routing Harness},
  author        = {{NeoHorse Team} and Cao, Guoliang and Dai, Guohao and Guo, Tianyu and Han, Kai and Hu, Hailin and Jiang, Zihan and Kuang, Xiang and Li, Boxun and Li, Yulong and others},
  year          = {2026},
  eprint        = {2609.08183},
  archivePrefix = {arXiv},
  primaryClass  = {cs.CL},
  url           = {https://arxiv.org/abs/2609.08183}
}

@article{cassano2023multiple,
  title = {{MultiPL-E}: A Scalable and Polyglot Approach to Benchmarking Neural Code Generation},
  author = {Cassano, Federico and Gouwar, John and Nguyen, Daniel and Nguyen, Sydney and Phipps-Costin, Luna and Pinckney, Donald and Yee, Ming-Ho and Zi, Yangtian and Anderson, Carolyn Jane and Feldman, Molly Q. and others},
  journal = {IEEE Transactions on Software Engineering},
  volume = {49},
  number = {7},
  pages = {3675--3691},
  year = {2023},
  doi = {10.1109/TSE.2023.3267446},
  url = {https://doi.org/10.1109/TSE.2023.3267446},
}

@article{chen2021humaneval,
  title = {Evaluating Large Language Models Trained on Code},
  author = {Chen, Mark and Tworek, Jerry and Jun, Heewoo and Yuan, Qiming and Pinto, Henrique Ponde de Oliveira and Kaplan, Jared and Edwards, Harri and Burda, Yuri and Joseph, Nicholas and Brockman, Greg and others},
  journal = {arXiv preprint arXiv:2107.03374},
  year = {2021},
  url = {https://arxiv.org/abs/2107.03374},
}

@misc{chen2026rsisurvey,
  title         = {Recursive Self-Improvement in {AI}: From Bounded Self-Refinement to Autonomous Research Loops},
  author        = {Chen, Mingguang and Wang, Licheng and Qu, Bo},
  year          = {2026},
  eprint        = {2607.07663},
  archivePrefix = {arXiv},
  primaryClass  = {cs.AI},
  url           = {https://arxiv.org/abs/2607.07663}
}

@article{clark2018arc,
  title = {Think you have Solved Question Answering? Try {ARC}, the {AI2} Reasoning Challenge},
  author = {Clark, Peter and Cowhey, Isaac and Etzioni, Oren and Khot, Tushar and Sabharwal, Ashish and Schoenick, Carissa and Tafjord, Oyvind},
  journal = {arXiv preprint arXiv:1803.05457},
  year = {2018},
  url = {https://arxiv.org/abs/1803.05457},
}

@article{cobbe2021gsm8k,
  title = {Training Verifiers to Solve Math Word Problems},
  author = {Cobbe, Karl and Kosaraju, Vineet and Bavarian, Mohammad and Chen, Mark and Jun, Heewoo and Kaiser, Lukasz and Plappert, Matthias and Tworek, Jerry and Hilton, Jacob and Nakano, Reiichiro and others},
  journal = {arXiv preprint arXiv:2110.14168},
  year = {2021},
  url = {https://arxiv.org/abs/2110.14168},
}

@article{deepseekmath,
  title={{DeepSeekMath}: Pushing the Limits of Mathematical Reasoning in Open Language Models},
  author={Shao, Zhihong and Wang, Peiyi and Zhu, Qihao and Xu, Runxin and Song, Junxiao and Bi, Xiao and Zhang, Haowei and Zhang, Mingchuan and Li, Y. K. and Wu, Y. and others},
  journal={arXiv preprint arXiv:2402.03300},
  year={2024},
  url={https://arxiv.org/abs/2402.03300}
}

@inproceedings{deng2025swe,
  title={{SWE-Bench Pro}: Can {AI} Agents Solve Long-Horizon Software Engineering Tasks?},
  author={Deng, Xiang and Da, Jeff and Pan, Edwin and He, Yannis Yiming and Ide, Charles and Garg, Kanak and Lauffer, Niklas and Park, Andrew and Pasari, Nitin and Rane, Chetan and others},
  booktitle={Proceedings of the 43rd International Conference on Machine Learning (ICML)},
  year={2026},
  url={https://arxiv.org/abs/2509.16941},
  note={arXiv:2509.16941}
}

@misc{dou2026clbench,
  title         = {{CL-bench}: A Benchmark for Context Learning},
  author        = {Dou, Shihan and Zhang, Ming and Yin, Zhangyue and Huang, Chenhao and Shen, Yujiong and Wang, Junzhe and Chen, Jiayi and Ni, Yuchen and Ye, Junjie and Zhang, Cheng and others},
  year          = {2026},
  eprint        = {2602.03587},
  archivePrefix = {arXiv},
  primaryClass  = {cs.CL},
  url           = {https://arxiv.org/abs/2602.03587}
}

@misc{dou2026clbenchlife,
  title         = {{CL-bench Life}: Can Language Models Learn from Real-Life Context?},
  author        = {Dou, Shihan and Shen, Yujiong and Huang, Chenhao and Ye, Junjie and Chen, Jiayi and Wang, Junzhe and He, Qianyu and Liu, Shichun and Lv, Changze and Lin, Jiahang and others},
  year          = {2026},
  eprint        = {2604.27043},
  archivePrefix = {arXiv},
  primaryClass  = {cs.CL},
  url           = {https://arxiv.org/abs/2604.27043}
}

@inproceedings{du2025supergpqa,
  title = {{SuperGPQA}: Scaling {LLM} Evaluation across 285 Graduate Disciplines},
  author = {{M-A-P Team} and Du, Xinrun and Yao, Yifan and Ma, Kaijing and Wang, Bingli and Zheng, Tianyu and Zhu, King and Liu, Minghao and Liang, Yiming and Jin, Xiaolong and others},
  booktitle = {The Thirty-ninth Annual Conference on Neural Information Processing Systems (NeurIPS)},
  year = {2025},
  note = {arXiv:2502.14739},
  url = {https://arxiv.org/abs/2502.14739},
}

@inproceedings{evalplus,
  title = {Is Your Code Generated by {ChatGPT} Really Correct? Rigorous Evaluation of Large Language Models for Code Generation},
  author = {Liu, Jiawei and Xia, Chunqiu Steven and Wang, Yuyao and Zhang, Lingming},
  booktitle = {Thirty-seventh Conference on Neural Information Processing Systems},
  year = {2023},
  url = {https://openreview.net/forum?id=1qvx610Cu7},
}

@misc{factory2025droid,
  title        = {Droid: {Factory}'s Agent-Native Development Platform},
  author       = {{Factory AI}},
  year         = {2025},
  howpublished = {\url{https://github.com/factory-ai/factory}},
  note         = {Accessed 2026-10-14},
  url          = {https://github.com/factory-ai/factory}
}

@article{fedus2022switch,
  title={{Switch Transformers}: Scaling to Trillion Parameter Models with Simple and Efficient Sparsity},
  author={Fedus, William and Zoph, Barret and Shazeer, Noam},
  journal={Journal of Machine Learning Research},
  volume={23},
  number={120},
  pages={1--39},
  year={2022},
  url={https://jmlr.org/papers/v23/21-0998.html}
}

@inproceedings{gema2025mmluredux,
  title = {Are We Done with {MMLU}?},
  author = {Gema, Aryo Pradipta and Leang, Joshua Ong Jun and Hong, Giwon and Devoto, Alessio and Mancino, Alberto Carlo Maria and Saxena, Rohit and He, Xuanli and Zhao, Yu and Du, Xiaotang and Ghasemi Madani, Mohammad Reza and others},
  booktitle = {Proceedings of the 2025 Conference of the Nations of the Americas Chapter of the Association for Computational Linguistics: Human Language Technologies (Volume 1: Long Papers)},
  pages = {5069--5096},
  year = {2025},
  publisher = {Association for Computational Linguistics},
  doi = {10.18653/v1/2025.naacl-long.262},
  url = {https://aclanthology.org/2025.naacl-long.262/},
}

@misc{google2026gemma4,
  title         = {Gemma 4 Technical Report},
  author        = {{Gemma Team} and Sherif El Abd and Vaibhav Aggarwal and Robin Algayres and Alek Andreev and Olivier Bachem and Ian Ballantyne and Cormac Brick and Victor C{\u{a}}rbune and Michelle Casbon and others},
  year          = {2026},
  eprint        = {2607.02770},
  archivePrefix = {arXiv},
  primaryClass  = {cs.CL},
  url           = {https://arxiv.org/abs/2607.02770}
}

@misc{haas2025simpleqaverified,
  title         = {{SimpleQA Verified}: A Reliable Factuality Benchmark to Measure Parametric Knowledge},
  author        = {Haas, Lukas and Yona, Gal and D'Antonio, Giovanni and Goldshtein, Sasha and Das, Dipanjan},
  year          = {2025},
  eprint        = {2509.07968},
  archivePrefix = {arXiv},
  primaryClass  = {cs.CL},
  url           = {https://arxiv.org/abs/2509.07968}
}

@inproceedings{hendrycks2021mmlu,
  title     = {Measuring Massive Multitask Language Understanding},
  author    = {Hendrycks, Dan and Burns, Collin and Basart, Steven and Zou, Andy and Mazeika, Mantas and Song, Dawn and Steinhardt, Jacob},
  booktitle = {International Conference on Learning Representations (ICLR)},
  year      = {2021},
  eprint    = {2009.03300},
  archivePrefix = {arXiv},
  url       = {https://openreview.net/forum?id=d7KBjmI3GmQ}
}

@misc{ifm2026k2horizon,
  title        = {Introducing {K2 Horizon}: Frontier Performance, Radically Open},
  author       = {{IFM Team}},
  year         = {2026},
  howpublished = {Official blog, \url{https://ifm.ai/blog/k2/}},
  note         = {Model card: \url{https://huggingface.co/IFM/K2-Horizon-MoVA-36B-A4B}},
  url          = {https://ifm.ai/blog/k2/}
}

@misc{jackson2025aaomniscience,
  title         = {{AA-Omniscience}: Evaluating Cross-Domain Knowledge Reliability in Large Language Models},
  author        = {Jackson, Declan and Keating, William and Cameron, George and Hill-Smith, Micah},
  year          = {2025},
  eprint        = {2511.13029},
  archivePrefix = {arXiv},
  primaryClass  = {cs.CL},
  note          = {Public set: \url{https://huggingface.co/datasets/ArtificialAnalysis/AA-Omniscience-Public}},
  url           = {https://arxiv.org/abs/2511.13029}
}

@inproceedings{jain2025livecodebench,
  title     = {{LiveCodeBench}: Holistic and Contamination Free Evaluation of Large Language Models for Code},
  author    = {Jain, Naman and Han, King and Gu, Alex and Li, Wen-Ding and Yan, Fanjia and Zhang, Tianjun and Wang, Sida and Solar-Lezama, Armando and Sen, Koushik and Stoica, Ion},
  booktitle = {The Thirteenth International Conference on Learning Representations (ICLR)},
  year      = {2025},
  eprint    = {2403.07974},
  archivePrefix = {arXiv},
  url       = {https://openreview.net/forum?id=chfJJYC3iL}
}

@inproceedings{jimenez2023swe,
  title={{SWE}-bench: Can Language Models Resolve Real-World {GitHub} Issues?},
  author={Jimenez, Carlos E. and Yang, John and Wettig, Alexander and Yao, Shunyu and Pei, Kexin and Press, Ofir and Narasimhan, Karthik},
  booktitle={The Twelfth International Conference on Learning Representations (ICLR)},
  year={2024},
  url={https://openreview.net/forum?id=VTF8yNQM66},
  note={arXiv:2310.06770}
}

@misc{jordan2024muon,
  title={Muon: An optimizer for hidden layers in neural networks},
  author={Jordan, Keller and Jin, Yuchen and Boza, Vlado and You, Jiacheng and Cesista, Franz and Newhouse, Laker and Bernstein, Jeremy},
  year={2024},
  howpublished={Blog post, \url{https://kellerjordan.github.io/posts/muon/}},
  note={December 8, 2024; reference implementation at \url{https://github.com/KellerJordan/Muon}},
  url={https://kellerjordan.github.io/posts/muon/}
}

@misc{kahn2025swebenchleak,
  title={Repo State Loopholes During Agentic Evaluation},
  author={{jacobkahn}},
  note={Opened 2025-09-03 by GitHub user jacobkahn; marked fixed for SWE-bench Verified on 2026-02-03},
  howpublished={GitHub issue \#465, SWE-bench/SWE-bench},
  year={2025},
  month=sep,
  url={https://github.com/SWE-bench/SWE-bench/issues/465}
}

@misc{kernelsurvey2026,
  title         = {Towards Automated Kernel Generation in the Era of {LLMs}},
  author        = {Yu, Yang and Zang, Peiyu and Tsai, Chi Hsu and Wu, Haiming and Shen, Yixin and Zhang, Jialing and Wang, Haoyu and Xiao, Zhiyou and Shi, Jingze and Luo, Yuyu and others},
  year          = {2026},
  eprint        = {2601.15727},
  archivePrefix = {arXiv},
  primaryClass  = {cs.LG},
  url           = {https://arxiv.org/abs/2601.15727}
}

@misc{kimi2025kimilinear,
  title         = {{Kimi Linear}: An Expressive, Efficient Attention Architecture},
  author        = {{Kimi Team} and Yu Zhang and Zongyu Lin and Xingcheng Yao and Jiaxi Hu and Fanqing Meng and Chengyin Liu and Xin Men and Songlin Yang and Zhiyuan Li and others},
  year          = {2025},
  eprint        = {2510.26692},
  archivePrefix = {arXiv},
  primaryClass  = {cs.CL},
  url           = {https://arxiv.org/abs/2510.26692}
}

@inproceedings{lightman2024verify,
  title = {Let's Verify Step by Step},
  author = {Lightman, Hunter and Kosaraju, Vineet and Burda, Yura and Edwards, Harri and Baker, Bowen and Lee, Teddy and Leike, Jan and Schulman, John and Sutskever, Ilya and Cobbe, Karl},
  booktitle = {The Twelfth International Conference on Learning Representations (ICLR)},
  year = {2024},
  url = {https://openreview.net/forum?id=v8L0pN6EOi},
}

@article{liu2025muon,
  title={Muon is Scalable for {LLM} Training},
  author={Liu, Jingyuan and Su, Jianlin and Yao, Xingcheng and Jiang, Zhejun and Lai, Guokun and Du, Yulun and Qin, Yidao and Xu, Weixin and Lu, Enzhe and Yan, Junjie and others},
  journal={arXiv preprint arXiv:2502.16982},
  year={2025},
  url={https://arxiv.org/abs/2502.16982}
}

@inproceedings{loshchilov2019adamw,
  title={Decoupled Weight Decay Regularization},
  author={Loshchilov, Ilya and Hutter, Frank},
  booktitle={International Conference on Learning Representations (ICLR)},
  year={2019},
  url={https://openreview.net/forum?id=Bkg6RiCqY7}
}

@inproceedings{madaan2023selfrefine,
  title={Self-Refine: Iterative Refinement with Self-Feedback},
  author={Madaan, Aman and Tandon, Niket and Gupta, Prakhar and Hallinan, Skyler and Gao, Luyu and Wiegreffe, Sarah and Alon, Uri and Dziri, Nouha and Prabhumoye, Shrimai and Yang, Yiming and others},
  booktitle={Advances in Neural Information Processing Systems (NeurIPS)},
  year={2023},
  url={https://arxiv.org/abs/2303.17651}
}

@misc{maxim2025bifrost,
  title        = {Bifrost: {AI} Gateway},
  author       = {{Maxim AI}},
  year         = {2025},
  howpublished = {\url{https://github.com/maximhq/bifrost}},
  note         = {Accessed 2026-10-14},
  url          = {https://github.com/maximhq/bifrost}
}

@misc{nous2025hermesagent,
  title        = {{Hermes Agent}: The Self-Improving {AI} Agent},
  author       = {{Nous Research}},
  year         = {2025},
  howpublished = {\url{https://github.com/NousResearch/hermes-agent}},
  note         = {Accessed 2026-10-14},
  url          = {https://github.com/NousResearch/hermes-agent}
}

@misc{novikov2025alphaevolve,
  title         = {{AlphaEvolve}: A coding agent for scientific and algorithmic discovery},
  author        = {Novikov, Alexander and V{\~u}, Ng{\^a}n and Eisenberger, Marvin and Dupont, Emilien and Huang, Po-Sen and Wagner, Adam Zsolt and Shirobokov, Sergey and Kozlovskii, Borislav and Ruiz, Francisco J. R. and Mehrabian, Abbas and others},
  year          = {2025},
  eprint        = {2506.13131},
  archivePrefix = {arXiv},
  primaryClass  = {cs.AI},
  url           = {https://arxiv.org/abs/2506.13131}
}

@misc{nvidia2025nemotron3nano,
  title         = {Nemotron 3 {Nano}: Open, Efficient Mixture-of-Experts Hybrid {Mamba-Transformer} Model for Agentic Reasoning},
  author        = {{NVIDIA} and Aaron Blakeman and Aaron Grattafiori and Aarti Basant and Abhibha Gupta and Abhinav Khattar and Adi Renduchintala and Aditya Vavre and Akanksha Shukla and Akhiad Bercovich and others},
  year          = {2025},
  eprint        = {2512.20848},
  archivePrefix = {arXiv},
  primaryClass  = {cs.CL},
  url           = {https://arxiv.org/abs/2512.20848}
}

@misc{nvidia2026nemotron35lightning,
  title         = {Nemotron 3.5 {Super} and {Lightning}: Open, Efficient Mixture-of-Experts Hybrid {Mamba-Transformer} Models for Agentic {AI}},
  author        = {{NVIDIA} and Aastha Jhunjhunwala and Aayushi Ashokkumar Srivastava and Abhibha Gupta and Abhinav Garg and Abhinav Khattar and Achsah Wildish and Adam Rajfer and Adi Renduchintala and Adil Asif and Adriana Flores Miranda and others},
  year          = {2026},
  eprint        = {2610.38095},
  archivePrefix = {arXiv},
  primaryClass  = {cs.CL},
  note          = {Model card: \url{https://huggingface.co/nvidia/NVIDIA-Nemotron-3.5-Lightning-30B-A3B-Base-BF16}},
  url           = {https://arxiv.org/abs/2610.38095}
}

@misc{openai2024sweverified,
  title={Introducing {SWE-bench Verified}},
  author={Chowdhury, Neil and Aung, James and Shern, Chan Jun and Jaffe, Oliver and Sherburn, Dane and Starace, Giulio and Mays, Evan and Dias, Rachel and Aljubeh, Marwan and Glaese, Mia and others},
  howpublished={OpenAI Blog},
  year={2024},
  month=aug,
  url={https://openai.com/index/introducing-swe-bench-verified/},
  note={Updated February 24, 2025}
}

@misc{openai2025codex,
  title        = {Codex {CLI}},
  author       = {{OpenAI}},
  year         = {2025},
  howpublished = {\url{https://github.com/openai/codex}},
  note         = {Accessed 2026-10-14},
  url          = {https://github.com/openai/codex}
}

@misc{openai_chat_api,
  title        = {Chat Completions {API} Reference},
  author       = {{OpenAI}},
  year         = {2026},
  howpublished = {\url{https://platform.openai.com/docs/api-reference/chat}},
  note         = {Accessed 2026-10-14},
  url          = {https://platform.openai.com/docs/api-reference/chat}
}

@misc{openai_responses_api,
  title        = {Responses {API} Reference},
  author       = {{OpenAI}},
  year         = {2026},
  howpublished = {\url{https://platform.openai.com/docs/api-reference/responses}},
  note         = {Accessed 2026-10-14},
  url          = {https://platform.openai.com/docs/api-reference/responses}
}

@misc{openclaw,
  title        = {{OpenClaw}: Your Assistant, on Your Devices, in Your Chats},
  author       = {{OpenClaw contributors}},
  year         = {2026},
  howpublished = {\url{https://github.com/openclaw/openclaw}},
  note         = {Accessed 2026-10-14},
  url          = {https://github.com/openclaw/openclaw}
}

@misc{opencode,
  title        = {{OpenCode}: The Open Source {AI} Coding Agent},
  author       = {{Anomaly}},
  year         = {2025},
  howpublished = {\url{https://github.com/anomalyco/opencode}},
  note         = {Accessed 2026-10-14},
  url          = {https://github.com/anomalyco/opencode}
}

@inproceedings{ouyang2025kernelbench,
  title={{KernelBench}: Can {LLMs} Write Efficient {GPU} Kernels?},
  author={Ouyang, Anne and Guo, Simon and Arora, Simran and Zhang, Alex L. and Hu, William and R{\'e}, Christopher and Mirhoseini, Azalia},
  booktitle={Proceedings of the 42nd International Conference on Machine Learning (ICML)},
  series={Proceedings of Machine Learning Research},
  volume={267},
  pages={47356--47415},
  publisher={PMLR},
  year={2025},
  url={https://proceedings.mlr.press/v267/ouyang25a.html},
  note={arXiv:2502.10517}
}

@inproceedings{pan2025swegym,
  title={Training Software Engineering Agents and Verifiers with {SWE-Gym}},
  author={Pan, Jiayi and Wang, Xingyao and Neubig, Graham and Jaitly, Navdeep and Ji, Heng and Suhr, Alane and Zhang, Yizhe},
  booktitle={Proceedings of the 42nd International Conference on Machine Learning (ICML)},
  series={Proceedings of Machine Learning Research},
  volume={267},
  pages={47717--47737},
  publisher={PMLR},
  year={2025},
  url={https://proceedings.mlr.press/v267/pan25g.html},
  note={arXiv:2412.21139}
}

@misc{pi_agent,
  title        = {Pi: A Minimal, Extensible Agent Harness},
  author       = {{Earendil Works}},
  year         = {2025},
  howpublished = {\url{https://github.com/earendil-works/pi}},
  note         = {Accessed 2026-10-14},
  url          = {https://github.com/earendil-works/pi}
}

@inproceedings{pyatkin2025ifbench,
  title     = {Generalizing Verifiable Instruction Following},
  author    = {Pyatkin, Valentina and Malik, Saumya and Graf, Victoria and Ivison, Hamish and Huang, Shengyi and Dasigi, Pradeep and Lambert, Nathan and Hajishirzi, Hannaneh},
  booktitle = {Advances in Neural Information Processing Systems (NeurIPS), Datasets and Benchmarks Track},
  year      = {2025},
  eprint    = {2507.02833},
  archivePrefix = {arXiv},
  url       = {https://arxiv.org/abs/2507.02833}
}

@inproceedings{qiu2025globalbatch,
  title={Demons in the Detail: On Implementing Load Balancing Loss for Training Specialized Mixture-of-Expert Models},
  author={Qiu, Zihan and Huang, Zeyu and Zheng, Bo and Wen, Kaiyue and Wang, Zekun and Men, Rui and Titov, Ivan and Liu, Dayiheng and Zhou, Jingren and Lin, Junyang},
  booktitle={Proceedings of the 63rd Annual Meeting of the Association for Computational Linguistics (Volume 1: Long Papers)},
  pages={5005--5018},
  year={2025},
  address={Vienna, Austria},
  publisher={Association for Computational Linguistics},
  doi={10.18653/v1/2025.acl-long.249},
  url={https://aclanthology.org/2025.acl-long.249/}
}

@misc{qwen2025qwen3thinking2507,
  title        = {{Qwen3-30B-A3B-Thinking-2507}},
  author       = {{Qwen Team}},
  year         = {2025},
  month        = jul,
  howpublished = {Hugging Face model card, \url{https://huggingface.co/Qwen/Qwen3-30B-A3B-Thinking-2507}},
  url          = {https://huggingface.co/Qwen/Qwen3-30B-A3B-Thinking-2507}
}

@misc{qwen2026qwen35,
  title        = {{Qwen3.5}: Towards Native Multimodal Agents},
  author       = {{Qwen Team}},
  year         = {2026},
  month        = feb,
  howpublished = {Official blog, \url{https://qwen.ai/blog?id=qwen3.5}},
  note         = {Model cards: \url{https://huggingface.co/Qwen/Qwen3.5-35B-A3B}, \url{https://huggingface.co/Qwen/Qwen3.5-35B-A3B-Base}},
  url          = {https://qwen.ai/blog?id=qwen3.5}
}

@misc{qwen2026qwen36,
  title        = {{Qwen3.6-35B-A3B}: Agentic Coding Power, Now Open to All},
  author       = {{Qwen Team}},
  year         = {2026},
  month        = apr,
  howpublished = {Official blog, \url{https://qwen.ai/blog?id=qwen3.6-35b-a3b}},
  note         = {Model card: \url{https://huggingface.co/Qwen/Qwen3.6-35B-A3B}},
  url          = {https://qwen.ai/blog?id=qwen3.6-35b-a3b}
}

@misc{recursive2026flint,
  title        = {Introducing {Flint}},
  author       = {{Recursive}},
  year         = {2026},
  month        = oct,
  howpublished = {Official blog, \url{https://recursive.com/updates/introducing-recursive-flint}},
  note         = {Model card: \url{https://huggingface.co/RecursiveSI/flint}},
  url          = {https://recursive.com/updates/introducing-recursive-flint}
}

@inproceedings{rein2024gpqa,
  title = {{GPQA}: A Graduate-Level {Google}-Proof {Q\&A} Benchmark},
  author = {Rein, David and Hou, Betty Li and Stickland, Asa Cooper and Petty, Jackson and Pang, Richard Yuanzhe and Dirani, Julien and Michael, Julian and Bowman, Samuel R.},
  booktitle = {First Conference on Language Modeling (COLM)},
  year = {2024},
  url = {https://openreview.net/forum?id=Ti67584b98},
}

@inproceedings{shazeer2017moe,
  title={Outrageously Large Neural Networks: The Sparsely-Gated Mixture-of-Experts Layer},
  author={Shazeer, Noam and Mirhoseini, Azalia and Maziarz, Krzysztof and Davis, Andy and Le, Quoc and Hinton, Geoffrey and Dean, Jeff},
  booktitle={International Conference on Learning Representations (ICLR)},
  year={2017},
  url={https://openreview.net/forum?id=B1ckMDqlg}
}

@article{shumailov2024collapse,
  title={{AI} models collapse when trained on recursively generated data},
  author={Shumailov, Ilia and Shumaylov, Zakhar and Zhao, Yiren and Papernot, Nicolas and Anderson, Ross and Gal, Yarin},
  journal={Nature},
  volume={631},
  pages={755--759},
  year={2024},
  doi={10.1038/s41586-024-07566-y},
  url={https://doi.org/10.1038/s41586-024-07566-y}
}

@article{su2024rope,
  title={{RoFormer}: Enhanced {Transformer} with Rotary Position Embedding},
  author={Su, Jianlin and Ahmed, Murtadha and Lu, Yu and Pan, Shengfeng and Bo, Wen and Liu, Yunfeng},
  journal={Neurocomputing},
  volume={568},
  pages={127063},
  year={2024},
  doi={10.1016/j.neucom.2023.127063},
  url={https://arxiv.org/abs/2104.09864}
}

@inproceedings{suzgun2023bbh,
  title = {Challenging {BIG-Bench} Tasks and Whether Chain-of-Thought Can Solve Them},
  author = {Suzgun, Mirac and Scales, Nathan and Sch{\"a}rli, Nathanael and Gehrmann, Sebastian and Tay, Yi and Chung, Hyung Won and Chowdhery, Aakanksha and Le, Quoc V. and Chi, Ed H. and Zhou, Denny and Wei, Jason},
  booktitle = {Findings of the Association for Computational Linguistics: ACL 2023},
  pages = {13003--13051},
  year = {2023},
  publisher = {Association for Computational Linguistics},
  url = {https://aclanthology.org/2023.findings-acl.824/},
}

@article{wang2024auxfree,
  title={Auxiliary-Loss-Free Load Balancing Strategy for Mixture-of-Experts},
  author={Wang, Lean and Gao, Huazuo and Zhao, Chenggang and Sun, Xu and Dai, Damai},
  journal={arXiv preprint arXiv:2408.15664},
  year={2024},
  url={https://arxiv.org/abs/2408.15664}
}

@inproceedings{wang2024mmlupro,
  title = {{MMLU-Pro}: A More Robust and Challenging Multi-Task Language Understanding Benchmark},
  author = {Wang, Yubo and Ma, Xueguang and Zhang, Ge and Ni, Yuansheng and Chandra, Abhranil and Guo, Shiguang and Ren, Weiming and Arulraj, Aaran and He, Xuan and Jiang, Ziyan and others},
  booktitle = {Advances in Neural Information Processing Systems 37 (Datasets and Benchmarks Track)},
  year = {2024},
  url = {https://papers.nips.cc/paper_files/paper/2024/hash/ad236edc564f3e3156e1b2feafb99a24-Abstract-Datasets_and_Benchmarks_Track.html},
}

@inproceedings{wang2025cybergym,
  title={{CyberGym}: Evaluating {AI} Agents' Real-World Cybersecurity Capabilities at Scale},
  author={Wang, Zhun and Shi, Tianneng and He, Jingxuan and Cai, Matthew and Zhang, Jialin and Song, Dawn},
  booktitle={The Fourteenth International Conference on Learning Representations (ICLR)},
  year={2026},
  url={https://openreview.net/forum?id=2YvbLQEdYt},
  note={arXiv:2506.02548}
}

@inproceedings{wang2025openhands,
  title     = {{OpenHands}: An Open Platform for {AI} Software Developers as Generalist Agents},
  author    = {Wang, Xingyao and Li, Boxuan and Song, Yufan and Xu, Frank F. and Tang, Xiangru and Zhuge, Mingchen and Pan, Jiayi and Song, Yueqi and Li, Bowen and Singh, Jaskirat and others},
  booktitle = {The Thirteenth International Conference on Learning Representations (ICLR)},
  year      = {2025},
  note      = {arXiv:2407.16741. Code: \url{https://github.com/OpenHands/OpenHands}},
  url       = {https://arxiv.org/abs/2407.16741}
}

@misc{weng2026scaling,
  title        = {Scaling Laws, Carefully},
  author       = {Weng, Lilian},
  year         = {2026},
  month        = jun,
  howpublished = {Lil'Log blog post, \url{https://lilianweng.github.io/posts/2026-06-24-scaling-laws/}},
  note         = {Accessed 2026-10-14},
  url          = {https://lilianweng.github.io/posts/2026-06-24-scaling-laws/}
}

@misc{xia2026rrsi,
  title         = {{RRSI}: Regularized Recursive Self-Improvement of Agent Harnesses},
  author        = {Xia, Peng and Han, Rujun and Wang, Zifeng and Chen, Yanfei and Zhuang, Yufan and Lee, Yoonho and Huang, Chengsong and Yu, Han and CuiZhu, Zhongying and Ming, Yifei and others},
  year          = {2026},
  eprint        = {2609.24972},
  archivePrefix = {arXiv},
  primaryClass  = {cs.LG},
  url           = {https://arxiv.org/abs/2609.24972}
}

@inproceedings{xiong2024longcontext,
  title={Effective Long-Context Scaling of Foundation Models},
  author={Xiong, Wenhan and Liu, Jingyu and Molybog, Igor and Zhang, Hejia and Bhargava, Prajjwal and Hou, Rui and Martin, Louis and Rungta, Rashi and Sankararaman, Karthik Abinav and Oguz, Barlas and others},
  booktitle={Proceedings of the 2024 Conference of the North American Chapter of the Association for Computational Linguistics: Human Language Technologies (Volume 1: Long Papers)},
  pages={4643--4663},
  year={2024},
  address={Mexico City, Mexico},
  publisher={Association for Computational Linguistics},
  doi={10.18653/v1/2024.naacl-long.260},
  url={https://aclanthology.org/2024.naacl-long.260/}
}

@inproceedings{yang2024swe,
  title     = {{SWE}-agent: Agent-Computer Interfaces Enable Automated Software Engineering},
  author    = {Yang, John and Jimenez, Carlos E. and Wettig, Alexander and Lieret, Kilian and Yao, Shunyu and Narasimhan, Karthik and Press, Ofir},
  booktitle = {Advances in Neural Information Processing Systems},
  volume    = {37},
  pages     = {50528--50652},
  year      = {2024},
  note      = {arXiv:2405.15793. Code: \url{https://github.com/SWE-agent/SWE-agent}},
  url       = {https://arxiv.org/abs/2405.15793}
}

@article{yang2025qwen3,
  title={Qwen3 Technical Report},
  author={Yang, An and Li, Anfeng and Yang, Baosong and Zhang, Beichen and Hui, Binyuan and Zheng, Bo and Yu, Bowen and Gao, Chang and Huang, Chengen and Lv, Chenxu and others},
  journal={arXiv preprint arXiv:2505.09388},
  year={2025},
  url={https://arxiv.org/abs/2505.09388}
}

@inproceedings{yang2025swe,
  title={{SWE-smith}: Scaling Data for Software Engineering Agents},
  author={Yang, John and Lieret, Kilian and Jimenez, Carlos E. and Wettig, Alexander and Khandpur, Kabir and Zhang, Yanzhe and Hui, Binyuan and Press, Ofir and Schmidt, Ludwig and Yang, Diyi},
  booktitle={Advances in Neural Information Processing Systems (NeurIPS), Datasets and Benchmarks Track},
  year={2025},
  url={https://arxiv.org/abs/2504.21798},
  note={arXiv:2504.21798}
}

@inproceedings{yue2025rlpassk,
  title={Does Reinforcement Learning Really Incentivize Reasoning Capacity in {LLMs} Beyond the Base Model?},
  author={Yue, Yang and Chen, Zhiqi and Lu, Rui and Zhao, Andrew and Wang, Zhaokai and Yue, Yang and Song, Shiji and Huang, Gao},
  booktitle={Advances in Neural Information Processing Systems (NeurIPS)},
  year={2025},
  url={https://arxiv.org/abs/2504.13837}
}

@inproceedings{zhao2025absolutezero,
  title={Absolute Zero: Reinforced Self-play Reasoning with Zero Data},
  author={Zhao, Andrew and Wu, Yiran and Yue, Yang and Wu, Tong and Xu, Quentin and Yue, Yang and Lin, Matthieu and Wang, Shenzhi and Wu, Qingyun and Zheng, Zilong and Huang, Gao},
  booktitle={Advances in Neural Information Processing Systems (NeurIPS)},
  year={2025},
  url={https://arxiv.org/abs/2505.03335}
}

@inproceedings{zhong2024agieval,
  title = {{AGIEval}: A Human-Centric Benchmark for Evaluating Foundation Models},
  author = {Zhong, Wanjun and Cui, Ruixiang and Guo, Yiduo and Liang, Yaobo and Lu, Shuai and Wang, Yanlin and Saied, Amin and Chen, Weizhu and Duan, Nan},
  booktitle = {Findings of the Association for Computational Linguistics: NAACL 2024},
  pages = {2299--2314},
  year = {2024},
  publisher = {Association for Computational Linguistics},
  doi = {10.18653/v1/2024.findings-naacl.149},
  url = {https://aclanthology.org/2024.findings-naacl.149/},
}

@article{zhong2025impossiblebench,
  title={{ImpossibleBench}: Measuring {LLMs}' Propensity of Exploiting Test Cases},
  author={Zhong, Ziqian and Raghunathan, Aditi and Carlini, Nicholas},
  journal={arXiv preprint arXiv:2510.20270},
  year={2025},
  url={https://arxiv.org/abs/2510.20270}
}

@misc{zhou2023ifeval,
  title         = {Instruction-Following Evaluation for Large Language Models},
  author        = {Zhou, Jeffrey and Lu, Tianjian and Mishra, Swaroop and Brahma, Siddhartha and Basu, Sujoy and Luan, Yi and Zhou, Denny and Hou, Le},
  year          = {2023},
  eprint        = {2311.07911},
  archivePrefix = {arXiv},
  primaryClass  = {cs.CL},
  url           = {https://arxiv.org/abs/2311.07911}
}

@article{zoph2022stmoe,
  title={{ST-MoE}: Designing Stable and Transferable Sparse Expert Models},
  author={Zoph, Barret and Bello, Irwan and Kumar, Sameer and Du, Nan and Huang, Yanping and Dean, Jeff and Shazeer, Noam and Fedus, William},
  journal={arXiv preprint arXiv:2202.08906},
  year={2022},
  url={https://arxiv.org/abs/2202.08906}
}

@misc{xu2026asiarch,
  author = {Weixian Xu and Yixiu Liu and Yang Nan and Lyumanshan Ye and Xiangkun Hu and Zhen Qin and Pengfei Liu},
  title = {{Neural Architecture Discovery via Autonomous Evolution}},
  year = {2026},
  eprint = {2507.18074},
  archiveprefix = {arXiv},
  primaryclass = {cs.AI},
  url = {https://arxiv.org/abs/2507.18074}
}

@article{xu2026asievolve,
  author = {Weixian Xu and Tiantian Mi and Yixiu Liu and Yang Nan and Zhimeng Zhou and Lyumanshan Ye and Lin Zhang and Yu Qiao and Pengfei Liu},
  title = {{ASI-Evolve: AI Accelerates AI}},
  journal = {CoRR},
  volume = {abs/2603.29640},
  year = {2026},
  doi = {10.48550/arXiv.2603.29640},
  eprint = {2603.29640},
  archiveprefix = {arXiv},
  url = {https://arxiv.org/abs/2603.29640}
}

@article{mi2026dataevolve,
  author = {Tiantian Mi and Dongming Shan and Zhen Huang and Yiwei Qin and Muhang Xie and Yuxuan Qiao and Yixiu Liu and Chenyang Zhou and Pengfei Liu},
  title = {{Data Darwinism Part II: DataEvolve -- AI can Autonomously Evolve Pretraining Data Curation}},
  journal = {CoRR},
  volume = {abs/2603.14420},
  year = {2026},
  doi = {10.48550/arXiv.2603.14420},
  eprint = {2603.14420},
  archiveprefix = {arXiv},
  url = {https://arxiv.org/abs/2603.14420}
}

@article{qin2026datadarwinismi,
  author = {Yiwei Qin and Zhen Huang and Tiantian Mi and Weiye Si and Chenyang Zhou and Qipeng Guo and Siyuan Feng and Pengfei Liu},
  title = {{Data Darwinism Part I: Unlocking the Value of Scientific Data for Pre-training}},
  journal = {CoRR},
  volume = {abs/2602.07824},
  year = {2026},
  doi = {10.48550/arXiv.2602.07824},
  eprint = {2602.07824},
  archiveprefix = {arXiv},
  url = {https://arxiv.org/abs/2602.07824}
}

@article{qin2026davincillm,
  author = {Yiwei Qin and Yixiu Liu and Tiantian Mi and Muhang Xie and Zhen Huang and Weiye Si and Pengrui Lu and Siyuan Feng and Xia Wu and Liming Liu and Ye Luo and Jinlong Hou and Qipeng Guo and Yu Qiao and Pengfei Liu},
  title = {{daVinci-LLM: Towards the Science of Pretraining}},
  journal = {CoRR},
  volume = {abs/2603.27164},
  year = {2026},
  doi = {10.48550/arXiv.2603.27164},
  eprint = {2603.27164},
  archiveprefix = {arXiv},
  url = {https://arxiv.org/abs/2603.27164}
}

@article{fu2026davincienv,
  author = {Dayuan Fu and Shenyu Wu and Yunze Wu and Zerui Peng and Yaxing Huang and Jie Sun and Ji Zeng and Mohan Jiang and Lin Zhang and Yukun Li and Jiarui Hu and Liming Liu and Jinlong Hou and Pengfei Liu},
  title = {{daVinci-Env: Open SWE Environment Synthesis at Scale}},
  journal = {CoRR},
  volume = {abs/2603.13023},
  year = {2026},
  doi = {10.48550/arXiv.2603.13023},
  eprint = {2603.13023},
  archiveprefix = {arXiv},
  url = {https://arxiv.org/abs/2603.13023}
}

@article{jiang2026davinciagency,
  author = {Mohan Jiang and Dayuan Fu and Junhao Shi and Ji Zeng and Weiye Si and Keyu Li and Xuefeng Li and Yang Xiao and Wenjie Li and Dequan Wang and Pengfei Liu},
  title = {{daVinci-Agency: Unlocking Long-Horizon Agency Data-Efficiently}},
  journal = {CoRR},
  volume = {abs/2602.02619},
  year = {2026},
  doi = {10.48550/arXiv.2602.02619},
  eprint = {2602.02619},
  archiveprefix = {arXiv},
  url = {https://arxiv.org/abs/2602.02619}
}

@inproceedings{zellers2019hellaswag,
  title = {{HellaSwag}: Can a Machine Really Finish Your Sentence?},
  author = {Zellers, Rowan and Holtzman, Ari and Bisk, Yonatan and Farhadi, Ali and Choi, Yejin},
  booktitle = {Proceedings of the 57th Annual Meeting of the Association for Computational Linguistics},
  year = {2019},
  pages = {4791--4800},
  doi = {10.18653/v1/P19-1472},
  url = {https://aclanthology.org/P19-1472/}
}

@inproceedings{mihaylov2018openbookqa,
  title = {Can a Suit of Armor Conduct Electricity? A New Dataset for Open Book Question Answering},
  author = {Mihaylov, Todor and Clark, Peter and Khot, Tushar and Sabharwal, Ashish},
  booktitle = {Proceedings of the 2018 Conference on Empirical Methods in Natural Language Processing},
  year = {2018},
  pages = {2381--2391},
  doi = {10.18653/v1/D18-1260},
  url = {https://aclanthology.org/D18-1260/}
}

@article{bisk2019piqa,
  title = {{PIQA}: Reasoning about Physical Commonsense in Natural Language},
  author = {Bisk, Yonatan and Zellers, Rowan and Le Bras, Ronan and Gao, Jianfeng and Choi, Yejin},
  journal = {arXiv preprint arXiv:1911.11641},
  year = {2019},
  url = {https://arxiv.org/abs/1911.11641}
}

@article{sakaguchi2019winogrande,
  title = {{WinoGrande}: An Adversarial {Winograd} Schema Challenge at Scale},
  author = {Sakaguchi, Keisuke and Le Bras, Ronan and Bhagavatula, Chandra and Choi, Yejin},
  journal = {arXiv preprint arXiv:1907.10641},
  year = {2019},
  url = {https://arxiv.org/abs/1907.10641}
}

@inproceedings{dua2019drop,
  title = {{DROP}: A Reading Comprehension Benchmark Requiring Discrete Reasoning Over Paragraphs},
  author = {Dua, Dheeru and Wang, Yizhong and Dasigi, Pradeep and Stanovsky, Gabriel and Singh, Sameer and Gardner, Matt},
  booktitle = {Proceedings of the 2019 Conference of the North American Chapter of the Association for Computational Linguistics: Human Language Technologies, Volume 1 (Long and Short Papers)},
  year = {2019},
  pages = {2368--2378},
  doi = {10.18653/v1/N19-1246},
  url = {https://aclanthology.org/N19-1246/}
}

@inproceedings{joshi2017triviaqa,
  title = {{TriviaQA}: A Large Scale Distantly Supervised Challenge Dataset for Reading Comprehension},
  author = {Joshi, Mandar and Choi, Eunsol and Weld, Daniel and Zettlemoyer, Luke},
  booktitle = {Proceedings of the 55th Annual Meeting of the Association for Computational Linguistics (Volume 1: Long Papers)},
  year = {2017},
  pages = {1601--1611},
  doi = {10.18653/v1/P17-1147},
  url = {https://aclanthology.org/P17-1147/}
}

@article{posttrainbench2026,
  title   = {PostTrainBench: Can {LLM} Agents Automate {LLM} Post-Training?},
  author  = {Rank, Ben and Bhatnagar, Hardik and Prabhu, Ameya and Eisenberg, Shira and Nguyen, Karina and Bethge, Matthias and Andriushchenko, Maksym},
  journal = {arXiv preprint arXiv:2603.08640},
  year    = {2026}
}

@article{rsiarena2026,
  title   = {{RSI Arena}: Towards Recursive Self-Improvement with Human Feedback},
  author  = {Huang, Yue and Wang, Yanbo and Li, Lichi and Chen, Yuanyuan and Haupt, Andreas and Luo, Xiaonan and Liu, Zheyuan and Choi, Jihye and others},
  journal = {arXiv preprint arXiv:2610.33624},
  year    = {2026}
}

@article{autodata2026,
  title   = {AutoData: Agentic Search for Pre-training Data Selection},
  author  = {Meng, Yan and Srikanth, Dhruv and Zhao, Bingchen and Jiang, Zhengyao and Wu, Yuxiang},
  journal = {arXiv preprint arXiv:2609.19754},
  year    = {2026}
}

@article{autollmresearch2026,
  title   = {AutoLLMResearch: Training Research Agents for Automating {LLM} Experiment Configuration -- Learning from Cheap, Optimizing Expensive},
  author  = {Guo, Taicheng and Chawla, Nitesh V. and Wiest, Olaf and Zhang, Xiangliang},
  journal = {arXiv preprint arXiv:2605.11518},
  year    = {2026}
}

@misc{karpathy2026autoresearch,
  title        = {autoresearch: {AI} agents running research on single-{GPU} nanochat training automatically},
  author       = {Karpathy, Andrej},
  howpublished = {\url{https://github.com/karpathy/autoresearch}},
  year         = {2026}
}
